\documentclass[10pt]{article}
\usepackage[preprint]{tmlr}
\usepackage[hyperfootnotes=false]{hyperref}
\hypersetup{colorlinks=true, linkcolor=blue!60!black, citecolor=blue!60!black, urlcolor=blue!60!black}
\usepackage{url}
\usepackage{booktabs}
\usepackage{graphicx}
\usepackage{xcolor}
\usepackage{amsmath,amssymb}
\usepackage{tabularx}
\usepackage{longtable}
\usepackage{placeins}
\usepackage{caption}
\graphicspath{{figures/}}

\newcommand{\coderevision}[3]{\href{https://github.com/#1/blob/#2/#3}{\texttt{#2}}}

\title{What Does Graph Self-Supervision Add\\to Node Classification?}

\author{\name Simone Copetti \email simone.copetti@mail.polimi.it \\
      \addr NECSTLab, DEIB, Politecnico di Milano
      \AND
      \name Leonardo De Grandis \email leonardo.degrandis@polimi.it \\
      \addr NECSTLab, DEIB, Politecnico di Milano
      \AND
      \name Marco D. Santambrogio \email marco.santambrogio@polimi.it \\
      \addr NECSTLab, DEIB, Politecnico di Milano}

\newcommand{\codestatement}{The experiment code, configurations and per-seed results used in the
  analyses are available at
  \url{https://github.com/Simocop7/graphssl/tree/paper-release-2026-11-29}.}

\begin{document}
\maketitle

\begin{abstract}
A graph encoder can improve after self-supervised training without outperforming a classifier
on fixed graph features. More subtly, its measured improvement can change without updating a
single encoder weight. We investigate what these observations reveal about the added value of
pretraining, crossing eight objectives with three architectures for frozen transductive node
classification. Under a shared training recipe, the average pretrained graph convolutional
network (GCN) improves on its initialization on six benchmarks. Yet when each pipeline receives
a validation-selected probe, its mean difference from fixed two-step propagation is negative on
all six, with seed intervals below zero on five. This average does not describe a method chosen
for use: retrospective selection on search-stage validation scores identifies pipelines that
outperform a broader fixed-feature family on two graphs, evaluated with training seeds held out
from selection. We then examine how the training gain itself is measured. Recomputing BatchNorm
statistics on the unlabeled graph, with weights fixed, reduces a short probe's mean excess gain
over a regularized probe from 6.4 to 0.6 percentage points across nine dataset--architecture
settings. In a three-seed control with Coauthor CS GCNs, matching embedding norms, which keeps
every affine linear decision boundary attainable, reduces the short probe's measured gain from
9.6 to 1.2 points, while probes that standardize their inputs report essentially unchanged
gains. Objective rankings are more stable across linear probes than across architectures.
Evaluating pretraining therefore requires both an architecture-matched starting point and a
competitive alternative without encoder training, together with an explicit account of how
their scores are measured.

\end{abstract}

{\let\thefootnote\relax\footnotetext{The authors used Claude (Anthropic) to assist with code and writing,
and Codex (OpenAI) to revise the manuscript. The authors take full
responsibility for the content.}}

\section{Introduction}
\label{sec:intro}

On Cora, a graph convolutional network (GCN) improves from 73.2\% accuracy at initialization
to 80.1\% after self-supervised pretraining, averaged over eight objectives. Yet mixing each
paper's word features with those of its citation neighbors twice, without learning an
encoder, gives 82.0\%. All three representations receive linear classifiers selected by the
same validation rule. The same trained models therefore have a positive 6.85-point training
gain and a negative $-1.90$-point difference from fixed propagation. What, then, has
pretraining added?

The comparisons answer different questions. The initialized encoder establishes whether the
training procedure improved its starting point. Fixed graph features establish what a
simpler predictive pipeline can achieve from the same inputs. Neither comparison substitutes
for the other. Nor does an average over objectives answer the practical question of which
method to deploy: selecting an objective and its configuration can produce a stronger
pipeline. We investigate the evidence required to distinguish these conclusions, rather than
use any one score as a general verdict on graph self-supervised learning (SSL).

A second question is less apparent. Even when the initialization is the relevant reference,
is its score being measured adequately? Training changes BatchNorm statistics as well as
weights, and a finite-budget classifier can respond differently to the scales of pretrained
and untrained embeddings. A measured gain could consequently change while every encoder
weight remains fixed. Distinguishing such a change from an improvement in the predictive
features requires interventions on the evaluation of both endpoints.

The ingredients are established. Simplified graph convolution uses a classifier on fixed
propagated features \citep{wu2019sgc,nt2019revisiting}; DGI, BGRL and FastGCL report competitive
random encoders \citep{velickovic2019dgi,thakoor2022bgrl,wang2022fastgcl}. PyGCL studies the
components of contrastive learning \citep{zhu2021pygcl}, and \citet{ma2024gcleval} show the sensitivity of graph SSL
rankings to hyperparameters and downstream tasks. Prior work also studies
probe normalization and inference-time BatchNorm statistics
\citep{lee2023normalization,wu2021batchnorm}. These findings motivate a further empirical
question: how do the references and their evaluation jointly affect the evidence attributed
to pretraining? We answer it on one grid that crosses objectives with architectures, by
comparing alternative pipelines and by intervening on the extraction and probing of the same
saved models. These interventions show that one evaluation choice can move the two endpoints
of a training gain unequally, and by how much. The reference baselines themselves, and the
arithmetic relating them, are not new.

The investigation proceeds from reference choice to reference measurement, then to the scope
of the resulting conclusions. We first establish the disagreement illustrated by Cora and
ask whether tuning, scarce training labels or different graphs change it
(Section~\ref{sec:graph}). A validation-selected SSL pipeline does outperform fixed propagation
on three graphs, two of them against a broader fixed-feature family, which changes what the
objective average can be taken to mean. We then hold weights
fixed and test how normalization statistics and embedding scale change the apparent training
gain (Section~\ref{sec:probe}). Finally, we distinguish the stability of that gain from the
stability of objective rankings across probes, architectures and prediction tasks
(Section~\ref{sec:robust}). Together, these comparisons establish what specific empirical
contrasts can support. They do not provide a causal decomposition of representation quality
or a bound on the performance of graph SSL.

\section{Experimental framework}
\label{sec:setup}

\paragraph{Four representations, two comparisons.}
The encoder maps each node to a vector, its representation or embedding. We freeze the
encoder while fitting the classifier. The setting is transductive: the features and edges
of test nodes are available during pretraining, and their labels are held out for evaluation. The classifier can receive the raw feature matrix $X$, or
the fixed representation $P=\hat A^2X$, where $\hat A$ is the symmetrically normalized
adjacency with self-loops: $\hat A=\tilde D^{-1/2}(A+I)\tilde D^{-1/2}$, where $\tilde D$ contains the row sums of $A+I$ \citep{wu2019sgc}. Only the classifier learns weights in these two
pipelines: $X$ measures performance accessible from the attributes directly, and $P$ adds
a fixed use of the graph. Alternatively it receives the output of an encoder at initialization ($U$) or
after SSL training from that initialization ($T$). The graphs are undirected or symmetrized before propagation. The two-step propagation matches the
operator and receptive-field depth of the GCN. For the other two encoder architectures it serves as an
alternative predictive pipeline. Width, nonlinearities and feature transformations can all
differ between $P$ and $U$.

Write $a(R)$ for the test score in percent (accuracy, or ROC-AUC on Minesweeper) of representation $R$ on a given dataset and
split under a specified probe-selection rule. The \emph{training gain} $G=a(T)-a(U)$ and the
\emph{propagation-relative advantage} $D=a(T)-a(P)$ are differences in percentage points.
The two differ by the score gap between the references,
\[
 G = D + \big[a(P)-a(U)\big].
\]
A large gap between the references leaves room for a positive $G$ even when $D$ is negative.
The identity is bookkeeping for three predictive pipelines, not a new estimator or an
additive decomposition of representation quality. It makes no assumption about how
much information an embedding contains or which component caused a particular prediction.

\paragraph{Two evaluation rules.}
Our default probe centers and scales each embedding coordinate using training-node statistics,
then fits multinomial logistic regression with the L-BFGS optimizer. An L2 penalty
discourages large classifier weights. We select its strength by accuracy on a separate set
of labeled validation nodes. For binary Minesweeper, probe fitting uses cross-entropy and
validation selection uses ROC-AUC.
For the primary propagation comparison, the same validation rule is applied separately to
\emph{each} fixed representation and each encoder checkpoint. It selects between the default
probe and row-normalized one-vs-rest logistic regression fitted with liblinear, each with its
own validation-selected penalty. Ties favor the default. The distinction matters on sparse
word features, particularly on CiteSeer (Appendix~\ref{app:e6}). On ogbn-arxiv only the
default regularized probe is available for this comparison. Encoders use their stored
BatchNorm statistics unless specified otherwise.

\paragraph{Experimental grid.}
We train each objective on three encoder architectures, called backbones: a GCN \citep{kipf2017gcn}, a graph isomorphism network (GIN)
\citep{xu2019gin} and a local-attention graph Transformer \citep{shi2021unimp}. All have two
layers of width 256, batch normalization and no dropout. Each objective includes the heads
and augmentations required by its loss. We use DGI, node-level GraphCL (a GRACE-style
contrastive objective), BGRL, AFGRL,
VICReg, Barlow Twins, GraphDINO and GraphMAE. The six main datasets use 300 full-batch AdamW updates
at learning rate $5\cdot10^{-4}$ \citep{loshchilov2019adamw} and common PyTorch Geometric
implementations \citep{fey2019pyg}. Sharing a recipe lets us change the objective while
holding the encoder and update count fixed. Equal updates do not imply equal computational
cost or convergence. Its suitability for each objective is an empirical
question, which Section~\ref{sec:stress} examines with a joint search over each objective's
hyperparameters. Appendix~\ref{app:protocol} gives the implementations and the sensitivity experiments.

Cora, CiteSeer and PubMed use their public splits with 20 labeled training nodes per class
\citep{yang2016planetoid}. Amazon Computers, Amazon Photo and Coauthor CS use one fixed
10/10/80 train/validation/test split \citep{shchur2018pitfalls}. These six datasets motivated
a follow-up on WikiCS, evaluated on public split 0, and one on three heterophilous graphs,
Roman-empire, Amazon-ratings and Minesweeper, evaluated on split 0 of their ten fixed splits
\citep{platonov2023heterophily}. Minesweeper has two classes and is scored by ROC-AUC. On
ogbn-arxiv we use 50 epochs with neighbor sampling and score the same saved models on subject
and year prediction.
Table~\ref{tab:prereg} records the rules fixed before particular follow-ups, including earlier
pilot results and prior exposure to the datasets.

\paragraph{Averaging and uncertainty.}
A seed specifies a run's random choices, including the initial encoder weights.
We average accuracy over seeds for each objective, then weight all objectives equally, even when
budget checks use only three seeds for two costly objectives on Coauthor CS. The main grids and
WikiCS have ten matched training seeds per objective, ogbn-arxiv three. Multiple probe
initializations on one checkpoint are averaged within that checkpoint. Random label splits
are separate sensitivity analyses, whose probe rules and label counts can differ from the
primary evaluation. Appendix~\ref{app:attribution} reports their observed ranges.
The joint search retrains each selected configuration
with five seeds. For the primary contrasts we bootstrap whole seed indices,
keeping all eight objectives and their shared untrained encoder together. The fixed features
are scored once. Intervals are 2.5th and 97.5th percentiles of 10{,}000 draws, conditional on
the observed split, dataset and protocol. We also report the original WikiCS interval, which
resamples seeds separately within each objective, to preserve its decision rule. The three
later tests of Section~\ref{sec:stress} use the joint resampling with 2{,}000 draws fixed in
their rules. For the joint search, intervals condition on the selected configurations and do not repeat
the search. For the label-budget test they resample paired seed--draw indices, not model
seeds and label draws independently. Neither the objectives nor the datasets are sampled
populations. We report pointwise intervals without a multiple-comparison correction; the
majority rules are descriptive decision criteria, not simultaneous significance guarantees.

\section{What does pretraining improve upon?}
\label{sec:graph}

\subsection{Six graphs under the shared recipe}

\begin{figure}[htb]
  \centering
  \includegraphics[width=\linewidth]{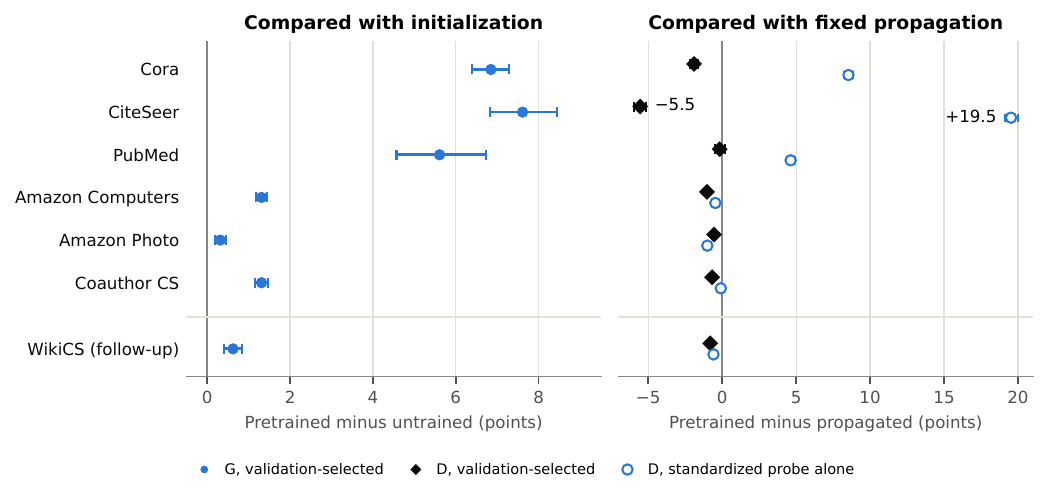}
  \caption{Two comparisons of the same pretrained GCNs. Each marker averages eight objective
  means on a fixed split. Positive values favor pretraining. Whiskers are 95\% percentile
  intervals from resampling ten whole seeds across objectives and their shared untrained
  reference. The first six graphs motivated the prespecified evaluation on WikiCS. Filled markers use the validation-selected probe for each representation. Open circles
  score all representations with the default standardized probe. The panels have different horizontal scales.}
  \label{fig:propagation}
\end{figure}

On each of the six main graphs, the average pretrained GCN improves over its initialization.
With validation-selected probes the mean is below fixed propagation on all six graphs.
The joint-seed interval is below zero on five and spans zero on PubMed
(Figure~\ref{fig:propagation}). The difference between the comparisons is already visible
before training. On Cora, for example, propagated features score 8.75 points above the untrained GCN.
Training gains 6.85 points on average, leaving a propagation-relative advantage of $-1.90$.
On Coauthor CS, both references already exceed 91\%, and there is much less room to improve.

\begin{table}[htb]
  \centering\small
  \caption{Absolute scores behind the two comparisons. Test accuracy (\%), with the same
  validation-selection rule applied separately to each representation. The GCN columns
  use the shared recipe and ten seeds; the pretrained column averages eight objectives.
  These initialized encoders are not references for differently configured tuned models.}
  \label{tab:four-references}
  \begin{tabular}{@{}lrrrr@{}}
\toprule
Dataset & Features $X$ & Fixed $P$ & Initial $U$ & Pretrained $T$ \\
\midrule
Cora & 60.2 & 82.0 & 73.2 & 80.1 \\
CiteSeer & 62.4 & 72.3 & 59.1 & 66.8 \\
PubMed & 72.9 & 79.7 & 73.9 & 79.5 \\
Computers & 79.9 & 89.5 & 87.2 & 88.5 \\
Photo & 86.8 & 92.8 & 91.9 & 92.2 \\
Coauthor CS & 92.1 & 93.1 & 91.1 & 92.4 \\
\bottomrule
\end{tabular}

\end{table}

\paragraph{Where the average conceals an advantage.}
Five of the 48 GCN objective--dataset means exceed propagation (Table~\ref{tab:absolute} and
Figure~\ref{fig:individual-propagation}). Three are on PubMed, where the best observed
objective leads by 1.6 points. The eight-objective mean there is $-0.17$ points relative to
propagation, with a joint-seed interval of $[-0.48,+0.17]$. Excluding GraphDINO, our adaptation
of a vision objective, reverses the mean to $+0.24$ points. The original within-objective
bootstrap gives $[+0.04,+0.43]$ for that seven-objective comparison. A retrospective bootstrap
keeping the seven objectives together gives $[-0.05,+0.54]$. The reversal in the point
estimate survives either calculation, while the evidence for a positive difference depends on
how shared seeds are handled. PubMed is also the one graph where tuning each objective tips
the comparison (Section~\ref{sec:stress}).

\paragraph{Why the probe can reverse this comparison.}
With the default standardized probe alone, the pretrained mean exceeds propagation on Cora,
CiteSeer and PubMed (open circles in Figure~\ref{fig:propagation}). The reversal comes mainly
from the propagated word features. They score substantially lower under coordinate
standardization and multinomial logistic regression than under row normalization and
one-vs-rest liblinear, while the pretrained means change little. The two probes differ in
both preprocessing and fitting, so this comparison cannot isolate standardization. On the
Amazon and Coauthor datasets and on dense WikiCS, both probes give the same sign
(Table~\ref{tab:probe-sensitivity}).

\subsection{When does the advantage change?}
\label{sec:stress}

The shared recipe may understate a useful method's performance, the labeled set may favor
a simple classifier, and averaging may suit the selected graphs unusually well. These are
different explanations for the initial observation. Before running four follow-ups, we wrote down what would overturn the
comparison (Table~\ref{tab:prereg}): a mean pretrained advantage over the fixed-feature
reference of at least one point, with a 95\% interval above zero. The required number of
graphs differs by test. These are decision thresholds for revising a claim, not tests of
equivalence or bounds on the best method. One
follow-up adds WikiCS; the others change the recipe, the labels and the graph family.
We also revisit the practical choice of an objective using the completed search records.
That additional selection analysis is retrospective, even though it uses validation scores only.

\begin{figure}[!htb]
  \centering
  \includegraphics[width=\linewidth]{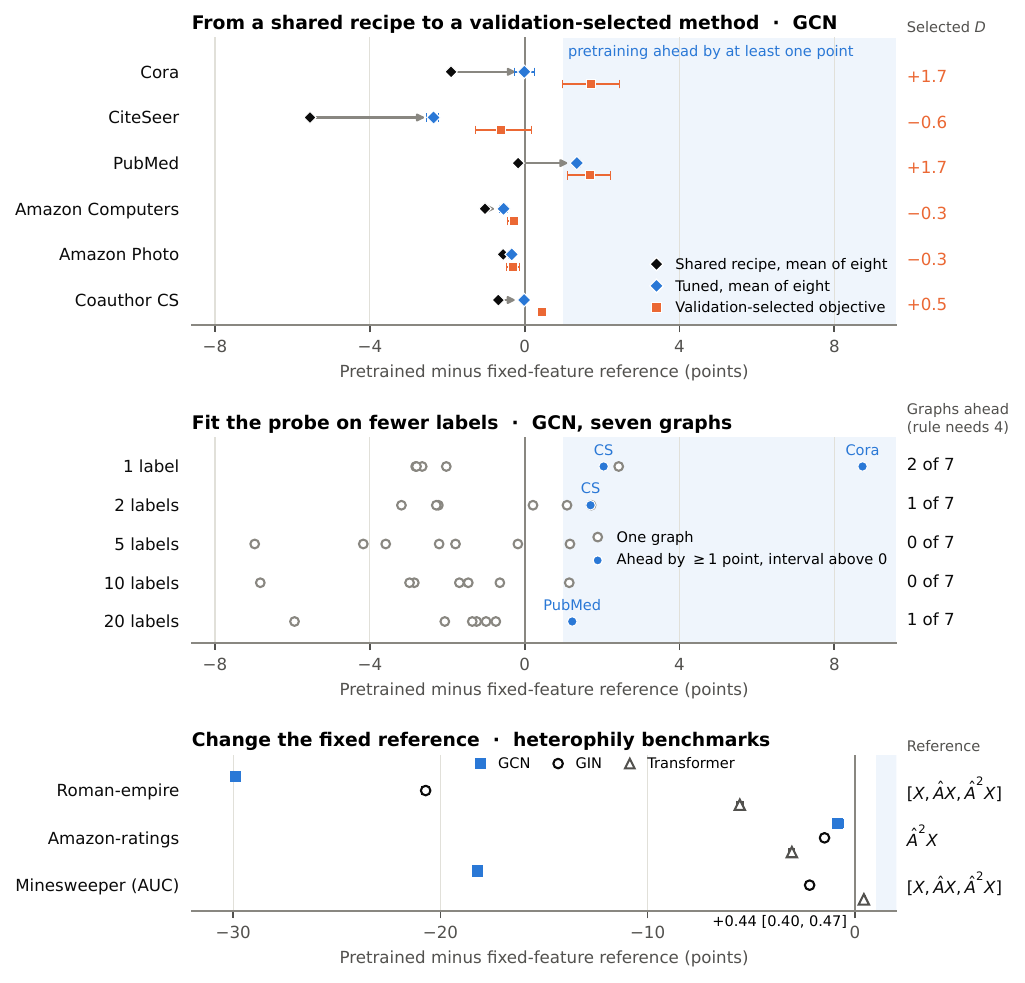}
  \caption{Where does pretraining exceed fixed features? Every marker is a pretrained score
  minus its reference. Top uses $\hat A^2X$: the shared-recipe mean, tuned objective mean and one objective
  selected using search-stage validation scores only. Both tuned estimates use final seeds
  0--4; the selected-objective analysis is retrospective. Whiskers condition on the completed
  search. The shaded one-point margin belongs to the original objective-average rules.
  Middle: probes refit with
  1 to 20 training labels per class, one marker per graph (the six main graphs and WikiCS);
  the rule needed four graphs at one budget. Bottom: three heterophilous graphs, each
  backbone's average over eight objectives against the best of $X$, $\hat A^2X$ and
  $[X,\hat AX,\hat A^2X]$ on validation (right column); the bottom panel has its own horizontal scale. Minesweeper uses ROC-AUC
  points. Error bars in the top and bottom panels are joint-seed intervals; the middle panel
  marks whether each seed--draw interval meets the decision rule.}
  \label{fig:stress}
\end{figure}

\paragraph{Tuning each objective.}
The shared recipe holds the encoder and budget fixed, so no objective gets its own best
settings. For each objective we drew 16 configurations once from a joint space: width,
depth, normalization, dropout, learning rate, weight decay, number of updates, augmentation
strength and the objective's own hyperparameters. The shared recipe was added as a
seventeenth. Validation accuracy alone selected one configuration per objective and graph,
which was then retrained with five seeds not used in configuration selection (Appendix~\ref{app:e11}). The tuned average is higher than the original shared-recipe average on every graph (Figure~\ref{fig:stress}, top). The remaining gap is
$0.01$ points on Cora and Coauthor CS, with intervals spanning zero; the average is positive only on
PubMed, by $1.35$ points $[1.26, 1.44]$, with seven of eight objectives above propagation.
CiteSeer remains 2.4 points below. The search tends to lift the weakest objectives most: on
Cora GraphDINO gains 5.4 points over the shared recipe, while GraphCL, already the best, gains
1.6.
Individual tuned objectives exceed propagation on five of the six graphs; the maximum
observed advantage is 1.7 points on Cora and 3.0 on PubMed. These maxima describe the test
results and cannot be treated as outcomes of method selection (Table~\ref{tab:e11}). Comparisons
with the original grid also change the device on three graphs; same-device, one-seed anchors
are reported separately in Appendix~\ref{app:e11}.

\paragraph{Selecting one method for use.}
The practitioner need not deploy the objective average. We retrospectively select one
objective per graph using the mean validation score of its chosen configuration on search
seeds 100--102. The selection never uses final seeds 0--4, whose test scores evaluate the
chosen pipeline with the same two-probe rule as the tuned averages. The search itself uses
the default probe.
GraphCL is selected on Cora and PubMed, exceeding propagation by
$1.72\ [0.98,2.46]$ and $1.68\ [1.10,2.22]$ points. GraphMAE is selected on Coauthor CS,
with a smaller advantage of $0.46\ [0.38,0.53]$ points. The selected methods trail propagation
on the remaining three graphs (Figure~\ref{fig:stress}, top;
Appendix~\ref{app:attribution}). Thus the objective-average result understates the performance
of a validation-selected method on some graphs. Its positive outcomes provide evidence for
specific learned pipelines, conditional on the search, split and probe; they do not establish
a general pretraining benefit. The tuned architectures also differ from the shared ones,
so their training gains require matching their own initializations.
For five graphs we replay the full model constructor with matching PyTorch and PyG versions
and rescore both endpoints together; the archived shared-recipe initial states provide a
check on this reconstruction. The selected CiteSeer model gains 21.5 points over its
replayed initialization while remaining 0.62 points below propagation. PubMed is excluded
from this comparison because its original initialization environment differs
(Appendix~\ref{app:attribution}).

\paragraph{Fewer labels for the probe.}
Self-supervision is often motivated by scarce labels. We refit the probes on the same frozen
encoders with 1, 2, 5, 10 and 20 training labels per class on all seven graphs, using nested
draws and taking all available training nodes when a class has fewer. Each model seed is
paired with one label draw. The probe is chosen separately for each representation, now
also allowing standardization using all nodes, and the standard validation set is retained.
With one label per class Cora favors pretraining by 8.7 points and Coauthor CS by 2.0
(Figure~\ref{fig:stress}, middle). No budget reaches the four graphs the rule required:
the prespecified majority criterion is not met at any tested budget. This does not negate
the substantial Cora advantage or the smaller positive differences on other graphs. Appendix~\ref{app:e9} gives every difference, and the counts when the
validation set is reduced to the same budget. The labels used to select the probe are a
separate resource.

\paragraph{Another homophilous graph.}
After examining the six graphs, we fixed the propagation depth, probe-selection rule and
decision criterion for WikiCS \citep{mernyei2020wikics}. A mean GCN advantage of at least one
point with an interval above zero would restrict the observation to the original datasets.
The estimate instead favors propagation ($D=-0.82$, original interval $[-0.88,-0.75]$;
Figure~\ref{fig:propagation}). Even GraphCL, the closest objective, falls slightly below
propagation. Prior exposure included published WikiCS results and short pilots
(Table~\ref{tab:prereg}). Outside the rule, the 100-step probe scores propagated features at
74.4\% and reverses the comparison ($D=+4.3$).

\paragraph{A stronger reference for heterophily benchmarks.}
On all seven graphs so far, neighboring nodes tend to share a label, which favors averaging
over neighbors. That averaging can also remove predictive information. To test this limitation, we trained the
full grid of objectives and backbones on Roman-empire, Amazon-ratings and Minesweeper
\citep{platonov2023heterophily}. These datasets have low adjusted homophily; Minesweeper
still has raw edge homophily 0.68 because its classes are imbalanced. On Roman-empire two-step propagation reaches only 44.8\%,
against 64.5\% for the raw features. The rule therefore compares each backbone with the validation-selected member
of three fixed-feature pipelines: $X$, $\hat A^2X$, and the side-by-side features
$[X,\hat AX,\hat A^2X]$, on which a linear probe can weight each propagation step separately.
The side-by-side features reach 76.0\% on Roman-empire and 89.4 ROC-AUC points on
Minesweeper. No backbone's average objective leads this reference by the prespecified one-point margin
on any graph (Figure~\ref{fig:stress}, bottom). On Roman-empire and Minesweeper the Transformer
comes closest, 5.6 points behind and $0.44$ points ahead ($[0.40,0.47]$), while the GCN, which also averages over
neighbors, trails by 30 and 18 points. On Amazon-ratings two-step propagation is itself the
best fixed pipeline, and all three backbones trail it by 0.8 to 3.0 points. Against $\hat A^2X$ alone, only Roman-empire
favors the pretrained GCN, by 1.4 points, short of the two graphs the rule required. The
Minesweeper advantage is positive despite missing the decision threshold. Its untrained
Transformer reaches 90.3 ROC-AUC points, above the pretrained mean of 89.8. Thus even this
advantage over fixed features cannot be attributed to pretraining. The appropriate reference
changes with the graph, and both comparisons remain necessary.

\paragraph{Using the same fixed-feature family across graphs.}
The heterophily result raises a further question: does restricting the other graphs to two
propagation steps favor SSL? In a retrospective sensitivity analysis, we select among
$X$, $\hat AX$, $\hat A^2X$ and $[X,\hat AX,\hat A^2X]$ on every main graph, WikiCS and the
three heterophily benchmarks. Selection uses validation only, including the probe choice.
Two-step propagation remains selected on the citation graphs, Computers and WikiCS.
Concatenation raises the reference to 94.1\% on Photo and 94.4\% on Coauthor CS. Thus the
validation-selected SSL methods retain their advantages on Cora and PubMed, while the
Coauthor CS advantage over two-step propagation disappears: the selected model is now
0.89 points below the fixed family. The three heterophily choices are unchanged.
This analysis strengthens the alternative pipeline without altering any original decision
rule (Appendix~\ref{app:attribution}).

\paragraph{Width and evaluation details.}
Under validation selection on Cora a projection of the propagated features onto 256 principal
components retains 81.5\% accuracy, close to the original 82.0\% and above the mean pretrained
GCN. The width-matched control is less competitive on Amazon Computers (88.0\%) and WikiCS
(78.7\%), where the mean pretrained GCN exceeds it (Table~\ref{tab:propagation}). Thus the
full-width reference matters. Recomputing normalization statistics and row-normalizing the
input features leave the average below propagation on the graphs tested. Two narrower forms
of tuning, selecting the optimizer and learning rate or a checkpoint through 3{,}000 steps on
validation, also leave the mean below propagation (Table~\ref{tab:checks}).

Each follow-up applied its own prespecified evaluation rule; the label-budget test also
expanded the probe choices. The earlier probe comparison alone moved the fixed-feature reference in Figure~\ref{fig:propagation}.
We next examine what the evaluation does to the other reference, the untrained encoder.

\section{How reliably can we measure the training gain?}
\label{sec:probe}

\begin{figure}[htb]
  \centering
  \includegraphics[width=\linewidth]{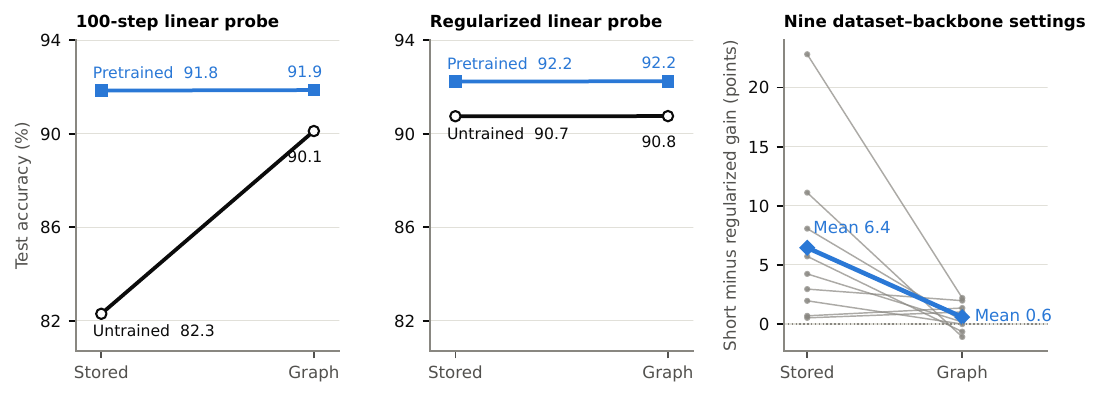}
  \caption{Which score moves when the encoder's normalization statistics are recomputed?
  The first two panels follow the same Coauthor CS GCNs through one unlabeled graph pass, keeping
  their weights fixed. The upper line averages eight pretrained objectives. The lower line
  follows their shared untrained reference. Each holds the classifier protocol fixed.
  Right: short-probe gain minus regularized-probe gain for the nine Amazon/Coauthor
  dataset--backbone settings; thin lines show settings and the thick line their mean.
  The right panel has its own scale and reports a difference of gains.}
  \label{fig:probe-scatter}
\end{figure}

For a fixed probe, the training gain is the vertical gap between pretrained and untrained
accuracy in Figure~\ref{fig:probe-scatter}. On Coauthor CS a linear classifier fitted for
100 Adam optimization steps reports a mean GCN training gain of 9.5 points. The
default regularized probe reports 1.5. Both receive exactly the same frozen embeddings and
labeled nodes. Most of the discrepancy comes from one term in the subtraction. Switching from
Adam to the regularized probe changes mean pretrained accuracy from 91.8\% to 92.2\%, while
untrained accuracy rises from 82.3\% to 90.7\%.

\paragraph{Intervening on the encoder statistics.}
Batch normalization (BatchNorm) rescales intermediate activations. To evaluate a frozen
encoder, each layer uses stored means and variances. At initialization these layers use zero running means and unit running variances in
evaluation mode \citep{ioffe2015bn}. After training, the stored statistics reflect repeated
passes through the encoder. On Coauthor CS the untrained GCN therefore presents the probe
with much smaller feature scales than its pretrained counterparts.
On a copy of the model, we reset BatchNorm's running statistics and counters, use cumulative
averaging for one unlabeled full-graph pass, then extract embeddings in evaluation mode.
Only BatchNorm enters training mode; dropout stays disabled. The stored variance is the
unbiased variance of each layer's inputs during that pass. The weights stay fixed, but
the model state and embeddings change. This uses unlabeled data, available at inference in
the transductive design (implementation audit in Appendix~\ref{app:attribution}).
With this intervention the short probe scores the untrained GCN at 90.1\% and the pretrained
GCNs at 91.9\% on average. The gain falls to 1.8 points. The pretrained scores changed very
little across the intervention, whereas the short probe gained almost eight points on the
untrained encoder alone.

\paragraph{Changing the classifier alone.}
On Coauthor CS two controls use the original embeddings extracted with stored statistics. Standardizing coordinates before the same 100 Adam
steps raises untrained accuracy to 89.2\% and reduces the gain to 2.3 points. Giving Adam
5{,}000 steps on unscaled embeddings, with checkpoint selection on validation, yields a
1.0-point gain. With standardization and the longer fit the gain is 1.8 points
(Table~\ref{tab:probe-ablation}). Both feature scaling and the finite fit budget matter here.
The longer protocol also changes the stopping rule.

\paragraph{A scale change that preserves linear decision boundaries.}
Recalibration changes the embeddings themselves, not only their scale. To isolate scale, a
retrospective control multiplies each saved Coauthor CS GCN embedding by one positive number.
For three encoder seeds and all eight objectives, the number matches the representation's
mean node norm to the mean over that seed's pretrained representations, without using labels.
The untrained embeddings grow 24 to 26 times. Any affine decision boundary available before
rescaling remains available after it: dividing the classifier weights by the same number
reproduces it. A fit from a fixed initialization, with a fixed number of updates or a fixed
weight penalty, is not invariant in this way. Table~\ref{tab:scale-main} reports the result.
With the same five classifier initializations and 100 Adam updates, the training gain falls
from 9.60 to 1.24 points, and only the untrained score moves. Probes that standardize
coordinates on the training nodes remove a global scale before fitting, so they act as
controls: they see the same inputs at either scale and report essentially the same gain. In
this setting, changing global scale alone removes most of the short probe's measured gain.
This does not account for every effect of recalibration, which also changes intermediate
activations. Appendix~\ref{app:attribution} adds fixed multipliers of 0.1 and 10.

\begin{table}[htb]
  \centering\small
  \caption{Positive rescaling of Coauthor CS GCN embeddings, without changing the encoder.
  $U$ and $T$ are test accuracies (\%) of the untrained encoder and the mean of eight
  pretrained objectives; $G=T-U$ in points. Three encoder seeds; each 100-step fit averages
  five classifier initializations. Norm matching multiplies each representation by one
  positive number chosen without labels. The ten-seed analysis of
  Figure~\ref{fig:probe-scatter} gives a 9.5-point short-probe gain on the same graph.}
  \label{tab:scale-main}
  \begin{tabular}{@{}lrrrrrr@{}}
\toprule
 & \multicolumn{3}{c}{Original scale} & \multicolumn{3}{c}{Norm matched} \\
\cmidrule(lr){2-4}\cmidrule(l){5-7}
Probe & $U$ & $T$ & $G$ & $U$ & $T$ & $G$ \\
\midrule
100-step Adam & 82.25 & 91.85 & $+9.60$ & 90.60 & 91.84 & $+1.24$ \\
Standardized, 100-step Adam & 89.24 & 91.50 & $+2.26$ & 89.24 & 91.50 & $+2.26$ \\
Regularized (default) & 90.87 & 92.23 & $+1.36$ & 90.88 & 92.23 & $+1.35$ \\
\bottomrule
\end{tabular}

\end{table}

\paragraph{Does the effect extend beyond Coauthor CS?}
For each dataset and backbone, we subtract the gain measured by the regularized probe from
the gain measured by the short probe. A positive difference means the short probe reports
more improvement from the same pretraining. Across the nine combinations of the three backbones with Amazon Computers, Amazon Photo and
Coauthor CS, the short probe's mean excess gain falls from 6.4 to 0.6 points
(Figure~\ref{fig:probe-scatter}, right; full grid in Appendix~\ref{app:e7}). This meets the prespecified criterion of a reduction by more
than half. Recomputing statistics for the untrained encoder alone, a descriptive check, leaves
0.4 points. Together these results locate most of the excess in the interaction of the short
probe with the untrained encoder's initial statistics. Other probes and architectures can
respond differently. On Cora, for example, the default-probe GIN gain drops from 30.1 to
21.6 points with graph statistics (Table~\ref{tab:e7}). On the heterophilous Roman-empire
graph the statistics even decide the sign of the regularized-probe gain. The mean GIN gain is
$-13.7$ points with stored statistics and $+10.3$ with graph statistics. The trained mean
barely moves (55.1\% and 56.0\%), while the untrained GIN falls from 68.8\% to 45.6\%
(Table~\ref{tab:e10}).

\paragraph{A different failure on CiteSeer.}
Under stored statistics, short-probe and regularized GCN gains are 1.5 and 8.6 points,
respectively. Here the short probe fits the training labels closely (mean final training loss
0.03 for pretrained GCNs). Across fits with zero or negligible L2 penalty their test accuracy is 55.7--60.6\%,
compared with 66.3--66.7\% under the regularized fits. Even converged fits with a negligible
penalty leave a gap, and validation selects one of the two strongest penalties for all 80
pretrained GCN checkpoints. This pattern implicates regularization as well as optimization.
After recalibration the two gains are 5.8 and 8.5 points. Across the three citation benchmarks
the short probe then gives a smaller gain in five of nine cells, below the six required by
the prespecified rule. Label counts, features and graphs all vary between the two groups.

The gain is a property of an encoder, an extraction rule and a downstream evaluation together.
Changing the untrained reference can shift every gain for a backbone by the same amount.
We therefore examine separately how often the trained objectives retain their order.

\section{When do comparisons between objectives transfer?}
\label{sec:robust}

\subsection{Changing the probe, backbone or training recipe}

\begin{figure}[htb]
  \centering
  \includegraphics[width=\linewidth]{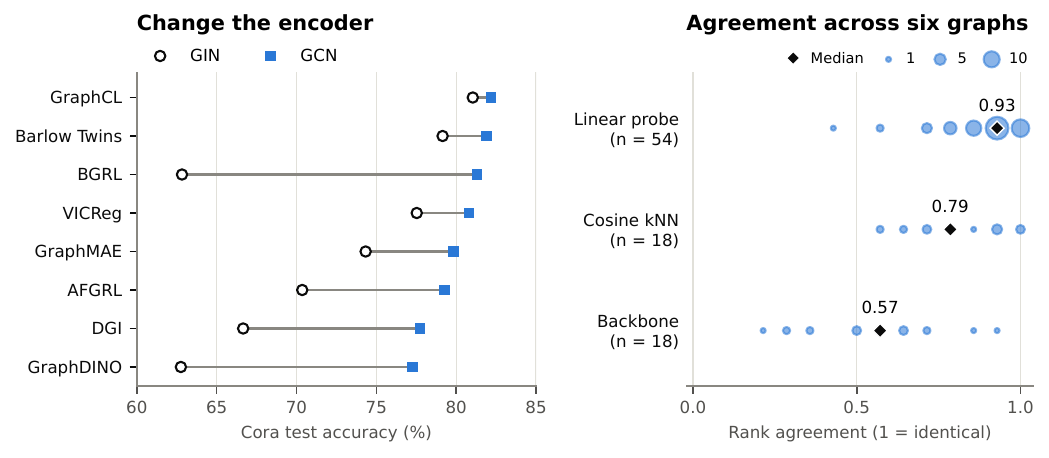}
  \caption{How well does an objective choice transfer? Left: each row follows one Cora
  objective between encoders under the same regularized probe, ordered by GCN accuracy.
  Right: each comparison ranks the eight objectives using all ten seeds. The first two rows
  change the classifier on the same embeddings. The last changes the backbone and holds the
  classifier fixed. Blue dot area counts comparisons at each agreement value. Diamonds mark
  the medians.}
  \label{fig:rank-transfer}
\end{figure}

The previous section changes how much training appears to help. If only the shared
untrained score moves, every objective's gain shifts equally and their ordering is unchanged.
Relative superiority therefore needs its own evidence: does picking an objective on one
encoder help on another? On Cora, the BGRL
objective has a higher mean test accuracy than AFGRL with a GCN. Their order reverses with a graph isomorphism
network (GIN). GraphCL stays near the top with either architecture.
Figure~\ref{fig:rank-transfer} shows the absolute scores for all eight objectives.
The size of the separation matters: on Amazon Photo the GCN and Transformer rankings
differ considerably, although the eight GCN means span less than one accuracy point.

To compare whole rankings, we use Kendall's $\tau_b$. It compares the order of pairs
of objectives, accounting for ties: one means identical order, zero means no net
agreement, and minus one means complete reversal. Within one backbone, all objectives
share an untrained baseline, so subtracting it preserves their order. Changing the
probe can still change the order by scoring the trained encoders differently.

\paragraph{Agreement on the six main datasets.}
Changing the linear probe gives median $\tau_b=0.93$ across 54 comparisons (100-step Adam,
5{,}000-step Adam and liblinear, each versus the default probe in 18 dataset--backbone
groups). A classifier that predicts from nearby labeled embeddings using cosine similarity
($k$-nearest neighbors, or $k$NN) gives median agreement 0.79
over 18 comparisons. Holding the default probe fixed and changing the backbone gives median
agreement 0.57 over 18 comparisons, one for each backbone pair and dataset.
Agreement ranges from 0.21 to 0.93. Changing normalization statistics alone gives median
agreement 1.00 under the default probe. Thus the evaluation change that substantially shifted
some training gains had little effect on typical within-backbone rankings.

\paragraph{How much disagreement can seed variability produce?}
Changing random seeds can also reorder objectives. We split the ten seeds into disjoint
halves and compare rankings within and across architectures using the same number of
seeds. In 15 of 18 backbone comparisons, even the 95th percentile of across-backbone
agreement falls below the lowest mean within-backbone agreement on that dataset. The
percentile describes the observed partitions. Figure~\ref{fig:seed-halves} gives all
comparisons and the calculation.

\paragraph{Transfer across recipes.}
Selecting an optimizer and learning rate for each objective on validation changes the order
as well (Table~\ref{tab:e3}). Across the six Cora and CiteSeer backbone groups, agreement with the shared recipe
ranges from 0.43 to 0.86. GraphMAE on the Cora Transformer moves from seventh to second.
Under the shared optimizer its best-validation checkpoint on Cora and CiteSeer occurs at
50--75 steps, and test accuracy falls by over ten points by step 300 on each. On the Amazon
and Coauthor graphs, selecting a GCN checkpoint from longer training changes the mean gain
by only 0.03--0.37 points relative to separate short runs (Appendix~\ref{app:e3e5}).
Embedding similarity also depends on the budget. The tendency for embeddings to group by
backbone at 300 steps misses its prespecified persistence criterion at 3{,}000 steps on Cora
(Appendix~\ref{sec:synthesis}). These comparisons change training. We next change only the
prediction problem presented to a set of saved encoders.

\subsection{Changing the task on the same citation graph}
\label{sec:scale}

Each of the 169{,}343 papers on ogbn-arxiv has a subject and a publication year
\citep{hu2020ogb}. We use the official chronological split for the 40 subjects and the first
random split for five quantile-based year classes \citep{lim2021linkx}. The symmetrized graph,
node features and encoder checkpoints are identical. The fraction of edges joining nodes
with the same label changes from 65\% to 22\%, along with the number of classes, the training
labels and the split. Their individual effects cannot be identified from this comparison.
With three training seeds per configuration, results on both tasks are descriptive.

\begin{table}[htb]
  \centering\small
  \caption{Two tasks on ogbn-arxiv with the default regularized probe. Both statistics
  conditions use the same checkpoints and Mac GPU extraction. Untrained and trained
  are test accuracies (\%), the latter averaged over eight objectives. Gain is the paired
  difference in points before rounding. Graph statistics use one unlabeled pass. Each
  configuration has three seeds. Fixed propagation, computed separately on CPU, gives
  69.2\% on subjects and 39.0\% on years.}
  \label{tab:arxiv-tasks}
\begin{tabular}{@{}llrrr@{\hspace{1.5em}}rrr@{}}
\toprule
 & & \multicolumn{3}{c}{Stored statistics} & \multicolumn{3}{c}{Graph statistics} \\
\cmidrule(lr){3-5}\cmidrule(l){6-8}
Task & Backbone & Untrained & Trained & Gain & Untrained & Trained & Gain \\
\midrule
Subject & GCN & 69.2 & 68.2 & $-$1.0 & 68.5 & 68.8 & +0.3 \\
 & GIN & 55.3 & 60.0 & +4.7 & 53.9 & 62.5 & +8.6 \\
 & Transformer & 66.4 & 64.1 & $-$2.4 & 66.0 & 64.1 & $-$1.9 \\
\addlinespace
Year & GCN & 39.8 & 42.1 & +2.4 & 41.8 & 42.2 & +0.3 \\
 & GIN & 40.4 & 39.7 & $-$0.7 & 40.0 & 39.8 & $-$0.2 \\
 & Transformer & 38.5 & 38.4 & $-$0.1 & 38.1 & 38.3 & +0.2 \\
\bottomrule
\end{tabular}

\end{table}

\paragraph{What changing the label changes.}
In the original evaluation (A2 GPU for encoders, CPU for fixed features), the mean pretrained
GCN falls 1.0 point below propagation on subjects and exceeds it by 3.1 points on years.
The untrained GCN already exceeds propagation on years. As on the heterophilous graphs of
Section~\ref{sec:stress}, two-step propagation is a weak reference when few edges join
nodes with the same label. A retrospective extension now selects among the same four
fixed-feature families used above, retaining this task's default probe. The side-by-side
features reach 41.6\%, against 39.0\% for two-step propagation. The original pretrained GCN
mean of 42.1\% is only 0.5 points above this stronger reference; the untrained GCN falls below
it (Appendix~\ref{app:attribution}). The training reference changes the
interpretation again. Under stored statistics in the paired Mac GPU evaluation, the mean GCN
gain is $-1.0$ points on subjects and $+2.4$ points on years. With graph statistics both gains
are $+0.3$ points (Table~\ref{tab:arxiv-tasks}). In particular, recalibration raises untrained
year accuracy by two points with little change to the pretrained mean. This distinction
matters even more for a short probe on subjects, where recalibration raises untrained GCN
accuracy from 46.5\% to 67.9\% in the paired evaluation.

For the GIN, seven of eight objectives improve subject prediction in the original evaluation,
and all eight finish below initialization on year prediction. The mean GIN gain retains this
difference in sign with graph statistics. For subject prediction the GIN and Transformer
orders even have negative agreement ($\tau_b=-0.79$, Table~\ref{tab:arxiv}).

\paragraph{A positive result under the published BGRL recipe.}
On subjects, BGRL under the shared recipe finishes 0.6 points below its untrained GCN. With
the full-graph pretraining recipe of \citet{thakoor2022bgrl}, including a three-layer GCN with
layer normalization and 10{,}000 updates, our implementation reaches 71.3\% versus 69.5\%
at initialization. Both scores use our regularized probe. The original study used a 100-step
AdamW probe and reported 71.6\% and 68.9\%, respectively (Appendix~\ref{app:arxiv}). The
positive gain under that recipe limits what one can infer from the shared-recipe comparison.

\section{Discussion}
\label{sec:recommendations}

The evidence for pretraining depends on the statement being made. Improvement over a matched
initialization establishes a gain for that training procedure and evaluation. It does not
establish an advantage over a competitive alternative without encoder training. The tuned
CiteSeer model separates the two most clearly: it improves by 21.5 points over its replayed
initialization and still finishes 0.62 points below two-step propagation. A selected
pipeline can provide the second comparison: the validation-selected GraphCL configurations
on Cora and PubMed exceed even the expanded fixed-feature family. On Coauthor CS, the same
expansion removes the selected method's advantage over two-step propagation. Reference
strength is therefore part of the empirical question, rather than a fixed judgment about SSL.

The normalization and rescaling controls identify a difficulty within the training-gain
comparison itself. Its starting point can move much more than its endpoint under the same
measurement change. A classifier's attainable linear decision boundaries can even remain
unchanged while its finite-budget score moves substantially. Finally, rankings that are
stable across probes need not transfer to another architecture or task. These controlled
contrasts delimit what a reported gain supports; they do not decompose the information in
an embedding or identify a universal effect of self-supervised learning. Nor do they depend
on the sign of our results: an objective that exceeded every fixed reference would need the
same two comparisons, and its measured gain the same controls.

\subsection{Relation to prior work}
\label{sec:related}

\paragraph{What an encoder can do before pretraining.}
Simplified graph convolution showed that a linear classifier on propagated features can match
a supervised GCN on citation networks \citep{wu2019sgc,nt2019revisiting}. Random encoders also
provide competitive features in some SSL settings. DGI and BGRL report untrained baselines
\citep{velickovic2019dgi,thakoor2022bgrl}. On Coauthor CS, the FastGCL study reports 93.14\% for an untrained GCN and 93.17\% for an untrained SGC, alongside the published BGRL score of 93.31\% \citep[Table~1]{wang2022fastgcl}. In molecular
tasks, an untrained GINE can be competitive with GraphCL at small sample sizes
\citep{khanna2025untrained}. \citet{trivedi2022augmentations} examine the role of graph augmentations, and
\citet{guo2023architecture} show how the architecture and normalization affect the behavior
of contrastive objectives. On PPI, \citet{thakoor2022bgrl} also found that switching from
mean pooling to graph attention helped BGRL and hurt GRACE. GraphMAE later reports different
PPI scores for BGRL and GRACE after fitting their classifiers to convergence rather than
for a fixed number of epochs \citep[Appendix~A.1]{hou2022graphmae}.
We examine the two reference comparisons together and perturb
normalization statistics at fixed weights to locate a discrepancy in their evaluation.

\paragraph{Comparing graph learning methods.}
Prior work on supervised graph learning has documented sensitivity to data splits, baseline
tuning and benchmark construction
\citep{shchur2018pitfalls,errica2020fair,tonshoff2023gap,platonov2023heterophily,luo2024classic}.
Similar concerns apply to unsupervised embeddings on graphs without node features
\citep{gurukar2022illusion}. In graph SSL, PyGCL studies contrastive components with a fixed
encoder \citep{zhu2021pygcl}, and \citet{ma2024gcleval} show how hyperparameters and downstream
tasks affect method rankings. GraphFM evaluates the same representations on node classification, link prediction and clustering
\citep{xu2024graphfm}, and GSTBench studies transfer across graphs \citep{song2025gstbench}.
Molecular pretraining benchmarks also find that gains depend on the evaluation setup
\citep{sun2022pretraining,wang2022molgrapheval}. Our design crosses objectives with backbones,
then changes the probe on each saved model. This lets us compare changes due to retraining
with changes due solely to evaluation.

\paragraph{Probing and sources of variation.}
In vision, rankings of self-supervised pretext tasks can change across architectures
\citep{kolesnikov2019revisiting}. Work on linear evaluation has examined task selection,
regularization, feature scaling and the probe's generalization error
\citep{goyal2019scaling,kornblith2019transfer,lee2023normalization,marks2024probing,dubois2023risk}.
\citet{lee2023normalization} show that normalization before the probe stabilizes visual SSL
scores across fitting hyperparameters. \citet{wu2021batchnorm} study inference statistics
recomputed at fixed model weights and their differences from training-time moving averages.
We do not introduce either mechanism. We use fixed-weight and global-rescaling controls to
measure their unequal effects on pretrained and untrained graph encoders: a training gain
requires two separate probe fits, and changing their difference can change its interpretation. We build on prior analyses of benchmark variance and reproducibility
\citep{bouthillier2021variance,summers2021nondeterminism,pham2020variance} by estimating their
effects on both the gain and the ordering of objectives.

\subsection{What to report in a frozen-encoder study}
\begin{enumerate}
  \item \textbf{State both reference comparisons.} The Cora example motivates reporting
  absolute scores for $X$, an appropriate fixed-feature family, the matched initialization
  and the pretrained encoder, with both $G$ and $D$. The Photo and Coauthor results show
  why retaining raw features alongside propagation is useful beyond heterophily benchmarks.
  \item \textbf{Describe how both endpoints are measured.} The Coauthor CS interventions
  motivate reporting normalization statistics, preprocessing, probe optimizer, penalty and
  fitting budget. Calibration tests dependence on extraction; positive global rescaling
  tests classifier fitting while preserving attainable affine decision boundaries.
  \item \textbf{Separate a selected method from summaries of a grid.} The objective-selection
  analysis motivates reporting the validation-selection rule, held-out evaluation seeds, search
  budget and chosen architecture. Give the objective average and descriptive test maximum
  different labels. Transfer to another architecture or task requires evidence there.
  \item \textbf{Match uncertainty and practical magnitude to the claim.} The near-zero
  differences and split sensitivities motivate reporting effect sizes alongside intervals
  and ranks. Preserve shared initializations in seed resampling; distinguish encoder seeds,
  probe fits and label splits. A small conditional interval does not establish equivalence
  or consistency across labeled-node selections.
\end{enumerate}

The untrained reference also served a diagnostic purpose during development: it exposed
plausible-looking results from incorrect normalization handling and a DGI implementation
that let real and corrupted graphs use separate batch statistics
(Appendix~\ref{sec:details}). This is a practical reason to retain the starting-point
comparison even when the fixed-feature pipeline is stronger.

\subsection{Limitations}
\label{sec:limitations}

Our main evidence concerns frozen transductive encoders on predominantly homophilous graphs,
with one primary test split per dataset. The existing random splits use the default probe
and more training labels on the citation graphs. Under them the mean pretrained GCN exceeds
two-step propagation on Cora and CiteSeer in all ten splits, and trails it on PubMed in all
ten (Table~\ref{tab:split-sensitivity}). They show that the sign of the comparison can depend
on the probe rule and the labeled nodes; they do not replicate the primary two-probe
comparison, and the selected tuned pipelines were evaluated on the original splits only.
Three heterophily benchmarks extend the fixed-reference comparison. Their normalization
checks are descriptive; they do not replicate the full probe and ranking analyses. Two more graphs of the same benchmark,
Tolokers and Questions, were left out for cost before any run. The label-budget follow-up
varies training labels within the standard splits. Its primary rule retains the standard validation labels. A full-graph calibration pass assumes access to that
graph at evaluation. An inductive deployment would have to define which nodes and edges are
available for estimating statistics. The two ogbn-arxiv tasks vary jointly in label meaning,
homophily, class count and split. The conclusions concern frozen node representations, not fine-tuning, inductive transfer
or graph-level prediction.

The shared grid contains three two-layer architectures and a fixed set of eight objectives.
Other reconstruction objectives and decoders are available
\citep{li2023maskgae,tan2023s2gae,hou2023graphmae2}. The joint search tunes the GCN only, with
16 random configurations per objective; the other backbones were tuned for optimizer and
learning rate on Cora and CiteSeer alone. A search keeps the encoder family fixed. For example,
GraphMAE with our GCN reaches 79.8\% on Cora under the default probe, and 79.7\% after the
joint search, while its authors report 84.2\% using GAT and 1{,}500 epochs.
They also report 82.9\% with a GCN, so a gap remains within the same encoder family,
under different training and evaluation recipes \citep[Tables~1, 6 and 7]{hou2022graphmae}.
The BGRL comparison likewise shows how much a published pretraining
recipe can matter.

Seed intervals condition on the observed graph, split, implementations and probe. Three seeds
on ogbn-arxiv, three to five in some budget checks and five in the joint search limit
precision. The joint search ran each graph on one of three devices. Hardware and software
vary across datasets. Matched machine reruns and provenance checks document differences
including changes in validation-selected penalties with thread count
(Appendices~\ref{app:protocol} and~\ref{sec:seeds}). A small seed interval or a failure to
exceed a prespecified threshold supplies no formal test of equivalence. The new method
selection, uniform reference family and scale control are retrospective. The scale control
uses three seeds on one graph; tuned initial states are reconstructed under audited matching
constructors and software, not recovered from archived initial checkpoints. We do not
reconstruct the tuned PubMed initialization across its different software environment.

\paragraph{Conclusion.}
What, then, does graph self-supervision add to frozen node classification in this study?
Relative to its own initialization, the average pretrained GCN improves on every main graph,
by an amount that depends on how the starting point is evaluated. Relative to fixed graph
features, the average objective has a lower mean score on all six, while validation-selected
pipelines exceed even a broader fixed-feature family on Cora and PubMed. Which objective is best transfers across probes more readily than across
architectures. A reported improvement supports a claim about pretraining only when it names
its reference and states how both endpoints were measured.

\paragraph{Code and data.} \codestatement

\bibliography{references}

@inproceedings{zhu2021pygcl,
  author = {Yanqiao Zhu and Yichen Xu and Qiang Liu and Shu Wu},
  title = {An Empirical Study of Graph Contrastive Learning},
  booktitle = {NeurIPS Datasets and Benchmarks Track},
  year = {2021}
}

@article{shchur2018pitfalls,
  author = {Oleksandr Shchur and Maximilian Mumme and Aleksandar Bojchevski and Stephan G{\"u}nnemann},
  title = {Pitfalls of Graph Neural Network Evaluation},
  journal = {arXiv preprint arXiv:1811.05868},
  year = {2018}
}

@inproceedings{errica2020fair,
  author = {Federico Errica and Marco Podda and Davide Bacciu and Alessio Micheli},
  title = {A Fair Comparison of Graph Neural Networks for Graph Classification},
  booktitle = {International Conference on Learning Representations (ICLR)},
  year = {2020}
}

@inproceedings{platonov2023heterophily,
  author = {Oleg Platonov and Denis Kuznedelev and Michael Diskin and Artem Babenko and Liudmila Prokhorenkova},
  title = {A critical look at the evaluation of {GNNs} under heterophily: Are we really making progress?},
  booktitle = {International Conference on Learning Representations (ICLR)},
  year = {2023}
}

@article{tonshoff2023gap,
  author = {Jan T{\"o}nshoff and Martin Ritzert and Eran Rosenbluth and Martin Grohe},
  title = {Where Did the Gap Go? {R}eassessing the Long-Range Graph Benchmark},
  journal = {Transactions on Machine Learning Research},
  year = {2024}
}

@inproceedings{luo2024classic,
  author = {Yuankai Luo and Lei Shi and Xiao-Ming Wu},
  title = {Classic {GNNs} are Strong Baselines: {R}eassessing {GNNs} for Node Classification},
  booktitle = {NeurIPS Datasets and Benchmarks Track},
  year = {2024}
}

@article{ma2024gcleval,
  author = {Qian Ma and Hongliang Chi and Hengrui Zhang and Kay Liu and Zhiwei Zhang and Lu Cheng and Suhang Wang and Philip S. Yu and Yao Ma},
  title = {Overcoming Pitfalls in Graph Contrastive Learning Evaluation: {T}oward Comprehensive Benchmarks},
  journal = {SIGKDD Explorations},
  year = {2025}
}

@article{xu2024graphfm,
  author = {Yuhao Xu and Xinqi Liu and Keyu Duan and Yi Fang and Yu-Neng Chuang and Daochen Zha and Qiaoyu Tan},
  title = {{GraphFM}: A Comprehensive Benchmark for Graph Foundation Model},
  journal = {arXiv preprint arXiv:2406.08310},
  year = {2024}
}

@inproceedings{song2025gstbench,
  author = {Yu Song and Zhigang Hua and Yan Xie and Jingzhe Liu and Bo Long and Hui Liu},
  title = {{GSTBench}: A Benchmark Study on the Transferability of Graph Self-Supervised Learning},
  booktitle = {ACM International Conference on Information and Knowledge Management (CIKM)},
  year = {2025}
}

@inproceedings{wang2022molgrapheval,
  author = {Hanchen Wang and Jean Kaddour and Shengchao Liu and Jian Tang and Joan Lasenby and Qi Liu},
  title = {Evaluating Self-Supervised Learning for Molecular Graph Embeddings},
  booktitle = {NeurIPS Datasets and Benchmarks Track},
  year = {2023}
}

@inproceedings{sun2022pretraining,
  author = {Ruoxi Sun and Hanjun Dai and Adams Wei Yu},
  title = {Does {GNN} Pretraining Help Molecular Representation?},
  booktitle = {Advances in Neural Information Processing Systems (NeurIPS)},
  year = {2022}
}

@inproceedings{trivedi2022augmentations,
  author = {Puja Trivedi and Ekdeep Singh Lubana and Yujun Yan and Yaoqing Yang and Danai Koutra},
  title = {Augmentations in Graph Contrastive Learning: {C}urrent Methodological Flaws \& Towards Better Practices},
  booktitle = {ACM Web Conference (WWW)},
  year = {2022}
}

@article{khanna2025untrained,
  author = {Smayan Khanna and Doruk Efe G{\"o}kmen and Risi Kondor and Vincenzo Vitelli},
  title = {Graph Contrastive Learning versus Untrained Baselines: {T}he Role of Dataset Size},
  journal = {arXiv preprint arXiv:2509.01541},
  year = {2025}
}

@inproceedings{guo2023architecture,
  author = {Xiaojun Guo and Yifei Wang and Zeming Wei and Yisen Wang},
  title = {Architecture Matters: {U}ncovering Implicit Mechanisms in Graph Contrastive Learning},
  booktitle = {Advances in Neural Information Processing Systems (NeurIPS)},
  year = {2023}
}

@inproceedings{kolesnikov2019revisiting,
  author = {Alexander Kolesnikov and Xiaohua Zhai and Lucas Beyer},
  title = {Revisiting Self-Supervised Visual Representation Learning},
  booktitle = {IEEE/CVF Conference on Computer Vision and Pattern Recognition (CVPR)},
  year = {2019}
}

@inproceedings{goyal2019scaling,
  author = {Priya Goyal and Dhruv Mahajan and Abhinav Gupta and Ishan Misra},
  title = {Scaling and Benchmarking Self-Supervised Visual Representation Learning},
  booktitle = {IEEE/CVF International Conference on Computer Vision (ICCV)},
  year = {2019}
}

@inproceedings{kornblith2019transfer,
  author = {Simon Kornblith and Jonathon Shlens and Quoc V. Le},
  title = {Do Better {ImageNet} Models Transfer Better?},
  booktitle = {IEEE/CVF Conference on Computer Vision and Pattern Recognition (CVPR)},
  year = {2019}
}

@article{marks2024probing,
  author = {Markus Marks and Manuel Knott and Neehar Kondapaneni and Elijah Cole and Thijs Defraeye and Fernando Perez-Cruz and Pietro Perona},
  title = {A Closer Look at Benchmarking Self-Supervised Pre-training with Image Classification},
  journal = {International Journal of Computer Vision},
  volume = {133},
  pages = {5013--5025},
  year = {2025}
}

@inproceedings{dubois2023risk,
  author = {Yann Dubois and Tatsunori Hashimoto and Percy Liang},
  title = {Evaluating Self-Supervised Learning via Risk Decomposition},
  booktitle = {International Conference on Machine Learning (ICML)},
  year = {2023}
}

@inproceedings{bouthillier2021variance,
  author = {Xavier Bouthillier and Pierre Delaunay and Mirko Bronzi and Assya Trofimov and Brennan Nichyporuk and Justin Szeto and Naz Sepah and Edward Raff and Kanika Madan and Vikram Voleti and Samira Ebrahimi Kahou and Vincent Michalski and Dmitriy Serdyuk and Tal Arbel and Chris Pal and Ga{\"e}l Varoquaux and Pascal Vincent},
  title = {Accounting for Variance in Machine Learning Benchmarks},
  booktitle = {Conference on Machine Learning and Systems (MLSys)},
  year = {2021}
}

@inproceedings{garrido2023rankme,
  author = {Quentin Garrido and Randall Balestriero and Laurent Najman and Yann LeCun},
  title = {{RankMe}: {A}ssessing the downstream performance of pretrained self-supervised representations by their rank},
  booktitle = {International Conference on Machine Learning (ICML)},
  year = {2023}
}

@article{richemond2020byolbn,
  author = {Pierre H. Richemond and Jean-Bastien Grill and Florent Altch{\'e} and Corentin Tallec and Florian Strub and Andrew Brock and Samuel Smith and Soham De and Razvan Pascanu and Bilal Piot and Michal Valko},
  title = {{BYOL} works even without batch statistics},
  journal = {arXiv preprint arXiv:2010.10241},
  year = {2020}
}

@inproceedings{he2020moco,
  author = {Kaiming He and Haoqi Fan and Yuxin Wu and Saining Xie and Ross Girshick},
  title = {Momentum Contrast for Unsupervised Visual Representation Learning},
  booktitle = {IEEE/CVF Conference on Computer Vision and Pattern Recognition (CVPR)},
  year = {2020}
}

@article{wu2021batchnorm,
  author = {Yuxin Wu and Justin Johnson},
  title = {Rethinking ``{B}atch'' in {BatchNorm}},
  journal = {arXiv preprint arXiv:2105.07576},
  year = {2021}
}

@inproceedings{wu2019sgc,
  author = {Felix Wu and Tianyi Zhang and de Souza, Jr., Amauri Holanda and Christopher Fifty and Tao Yu and Kilian Q. Weinberger},
  title = {Simplifying Graph Convolutional Networks},
  booktitle = {International Conference on Machine Learning (ICML)},
  year = {2019}
}

@article{nt2019revisiting,
  author = {Hoang NT and Takanori Maehara},
  title = {Revisiting Graph Neural Networks: {A}ll We Have is Low-Pass Filters},
  journal = {arXiv preprint arXiv:1905.09550},
  year = {2019}
}

@article{lee2023normalization,
  author = {Jae-Hun Lee and Doyoung Yoon and ByeongMoon Ji and Kyungyul Kim and Sangheum Hwang},
  title = {Rethinking Evaluation Protocols of Visual Representations Learned via Self-supervised Learning},
  journal = {arXiv preprint arXiv:2304.03456},
  year = {2023}
}

@inproceedings{velickovic2019dgi,
  author = {Petar Veli{\v{c}}kovi{\'c} and William Fedus and William L. Hamilton and Pietro Li{\`o} and Yoshua Bengio and R Devon Hjelm},
  title = {Deep Graph {Infomax}},
  booktitle = {International Conference on Learning Representations (ICLR)},
  year = {2019}
}

@inproceedings{you2020graphcl,
  author = {Yuning You and Tianlong Chen and Yongduo Sui and Ting Chen and Zhangyang Wang and Yang Shen},
  title = {Graph Contrastive Learning with Augmentations},
  booktitle = {Advances in Neural Information Processing Systems (NeurIPS)},
  year = {2020}
}

@inproceedings{thakoor2022bgrl,
  author = {Shantanu Thakoor and Corentin Tallec and Mohammad Gheshlaghi Azar and Mehdi Azabou and Eva L. Dyer and R{\'e}mi Munos and Petar Veli{\v{c}}kovi{\'c} and Michal Valko},
  title = {Large-Scale Representation Learning on Graphs via Bootstrapping},
  booktitle = {International Conference on Learning Representations (ICLR)},
  year = {2022}
}

@inproceedings{lee2022afgrl,
  author = {Namkyeong Lee and Junseok Lee and Chanyoung Park},
  title = {Augmentation-Free Self-Supervised Learning on Graphs},
  booktitle = {AAAI Conference on Artificial Intelligence},
  year = {2022}
}

@inproceedings{bardes2022vicreg,
  author = {Adrien Bardes and Jean Ponce and Yann LeCun},
  title = {{VICReg}: {V}ariance-Invariance-Covariance Regularization for Self-Supervised Learning},
  booktitle = {International Conference on Learning Representations (ICLR)},
  year = {2022}
}

@inproceedings{zbontar2021barlow,
  author = {Jure Zbontar and Li Jing and Ishan Misra and Yann LeCun and St{\'e}phane Deny},
  title = {{Barlow Twins}: {S}elf-Supervised Learning via Redundancy Reduction},
  booktitle = {International Conference on Machine Learning (ICML)},
  year = {2021}
}

@inproceedings{caron2021dino,
  author = {Mathilde Caron and Hugo Touvron and Ishan Misra and Herv{\'e} J{\'e}gou and Julien Mairal and Piotr Bojanowski and Armand Joulin},
  title = {Emerging Properties in Self-Supervised Vision Transformers},
  booktitle = {IEEE/CVF International Conference on Computer Vision (ICCV)},
  year = {2021}
}

@inproceedings{kipf2017gcn,
  author = {Thomas N. Kipf and Max Welling},
  title = {Semi-Supervised Classification with Graph Convolutional Networks},
  booktitle = {International Conference on Learning Representations (ICLR)},
  year = {2017}
}

@inproceedings{xu2019gin,
  author = {Keyulu Xu and Weihua Hu and Jure Leskovec and Stefanie Jegelka},
  title = {How Powerful are Graph Neural Networks?},
  booktitle = {International Conference on Learning Representations (ICLR)},
  year = {2019}
}

@inproceedings{hou2022graphmae,
  author = {Zhenyu Hou and Xiao Liu and Yukuo Cen and Yuxiao Dong and Hongxia Yang and Chunjie Wang and Jie Tang},
  title = {{GraphMAE}: {S}elf-Supervised Masked Graph Autoencoders},
  booktitle = {ACM SIGKDD Conference on Knowledge Discovery and Data Mining (KDD)},
  year = {2022}
}

@inproceedings{hou2023graphmae2,
  author = {Zhenyu Hou and Yufei He and Yukuo Cen and Xiao Liu and Yuxiao Dong and Evgeny Kharlamov and Jie Tang},
  title = {{GraphMAE2}: {A} Decoding-Enhanced Masked Self-Supervised Graph Learner},
  booktitle = {ACM Web Conference (WWW)},
  year = {2023}
}

@inproceedings{li2023maskgae,
  author = {Jintang Li and Ruofan Wu and Wangbin Sun and Liang Chen and Sheng Tian and Liang Zhu and Changhua Meng and Zibin Zheng and Weiqiang Wang},
  title = {What's Behind the Mask: {U}nderstanding Masked Graph Modeling for Graph Autoencoders},
  booktitle = {ACM SIGKDD Conference on Knowledge Discovery and Data Mining (KDD)},
  year = {2023}
}

@inproceedings{tan2023s2gae,
  author = {Qiaoyu Tan and Ninghao Liu and Xiao Huang and Soo-Hyun Choi and Li Li and Rui Chen and Xia Hu},
  title = {{S2GAE}: Self-Supervised Graph Autoencoders Are Generalizable Learners with Graph Masking},
  booktitle = {ACM International Conference on Web Search and Data Mining (WSDM)},
  year = {2023}
}

@inproceedings{velickovic2018gat,
  author = {Petar Veli{\v{c}}kovi{\'c} and Guillem Cucurull and Arantxa Casanova and Adriana Romero and Pietro Li{\`o} and Yoshua Bengio},
  title = {Graph Attention Networks},
  booktitle = {International Conference on Learning Representations (ICLR)},
  year = {2018}
}

@inproceedings{fey2019pyg,
  author = {Matthias Fey and Jan Eric Lenssen},
  title = {Fast Graph Representation Learning with {PyTorch Geometric}},
  booktitle = {ICLR Workshop on Representation Learning on Graphs and Manifolds},
  year = {2019}
}

@inproceedings{zhu2020grace,
  author = {Yanqiao Zhu and Yichen Xu and Feng Yu and Qiang Liu and Shu Wu and Liang Wang},
  title = {Deep Graph Contrastive Representation Learning},
  booktitle = {ICML Workshop on Graph Representation Learning and Beyond},
  year = {2020}
}

@inproceedings{hu2020ogb,
  author = {Weihua Hu and Matthias Fey and Marinka Zitnik and Yuxiao Dong and Hongyu Ren and Bowen Liu and Michele Catasta and Jure Leskovec},
  title = {{Open Graph Benchmark}: {D}atasets for Machine Learning on Graphs},
  booktitle = {Advances in Neural Information Processing Systems (NeurIPS)},
  year = {2020}
}

@inproceedings{yang2016planetoid,
  author = {Zhilin Yang and William W. Cohen and Ruslan Salakhutdinov},
  title = {Revisiting Semi-Supervised Learning with Graph Embeddings},
  booktitle = {International Conference on Machine Learning (ICML)},
  year = {2016}
}

@inproceedings{kingma2015adam,
  author = {Diederik P. Kingma and Jimmy Ba},
  title = {{Adam}: {A} Method for Stochastic Optimization},
  booktitle = {International Conference on Learning Representations (ICLR)},
  year = {2015}
}

@inproceedings{loshchilov2019adamw,
  author = {Ilya Loshchilov and Frank Hutter},
  title = {Decoupled Weight Decay Regularization},
  booktitle = {International Conference on Learning Representations (ICLR)},
  year = {2019}
}

@inproceedings{ioffe2015bn,
  author = {Sergey Ioffe and Christian Szegedy},
  title = {Batch Normalization: {A}ccelerating Deep Network Training by Reducing Internal Covariate Shift},
  booktitle = {International Conference on Machine Learning (ICML)},
  year = {2015}
}

@article{fan2008liblinear,
  author = {Rong-En Fan and Kai-Wei Chang and Cho-Jui Hsieh and Xiang-Rui Wang and Chih-Jen Lin},
  title = {{LIBLINEAR}: A Library for Large Linear Classification},
  journal = {Journal of Machine Learning Research},
  volume = {9},
  pages = {1871--1874},
  year = {2008}
}

@article{liu1989lbfgs,
  author = {Dong C. Liu and Jorge Nocedal},
  title = {On the Limited Memory {BFGS} Method for Large Scale Optimization},
  journal = {Mathematical Programming},
  volume = {45},
  pages = {503--528},
  year = {1989}
}

@inproceedings{roy2007erank,
  author = {Olivier Roy and Martin Vetterli},
  title = {The Effective Rank: {A} Measure of Effective Dimensionality},
  booktitle = {European Signal Processing Conference (EUSIPCO)},
  year = {2007}
}

@inproceedings{pham2020variance,
  author = {Hung Viet Pham and Shangshu Qian and Jiannan Wang and Thibaud Lutellier and Jonathan Rosenthal and Lin Tan and Yaoliang Yu and Nachiappan Nagappan},
  title = {Problems and Opportunities in Training Deep Learning Software Systems: {A}n Analysis of Variance},
  booktitle = {IEEE/ACM International Conference on Automated Software Engineering (ASE)},
  year = {2020}
}

@inproceedings{summers2021nondeterminism,
  author = {Cecilia Summers and Michael J. Dinneen},
  title = {Nondeterminism and Instability in Neural Network Optimization},
  booktitle = {International Conference on Machine Learning (ICML)},
  year = {2021}
}

@inproceedings{lim2021linkx,
  author = {Derek Lim and Felix Hohne and Xiuyu Li and Sijia Linda Huang and Vaishnavi Gupta and Omkar Bhalerao and Ser-Nam Lim},
  title = {Large Scale Learning on Non-Homophilous Graphs: {N}ew Benchmarks and Strong Simple Methods},
  booktitle = {Advances in Neural Information Processing Systems (NeurIPS)},
  year = {2021}
}

@inproceedings{shi2021unimp,
  author = {Yunsheng Shi and Zhengjie Huang and Shikun Feng and Hui Zhong and Wenjin Wang and Yu Sun},
  title = {Masked Label Prediction: {U}nified Message Passing Model for Semi-Supervised Classification},
  booktitle = {International Joint Conference on Artificial Intelligence (IJCAI)},
  year = {2021}
}

@inproceedings{kornblith2019cka,
  author = {Simon Kornblith and Mohammad Norouzi and Honglak Lee and Geoffrey Hinton},
  title = {Similarity of Neural Network Representations Revisited},
  booktitle = {International Conference on Machine Learning (ICML)},
  year = {2019}
}

@inproceedings{wang2020uniformity,
  author = {Tongzhou Wang and Phillip Isola},
  title = {Understanding Contrastive Representation Learning through Alignment and Uniformity on the Hypersphere},
  booktitle = {International Conference on Machine Learning (ICML)},
  year = {2020}
}

@article{gurukar2022illusion,
  author = {Saket Gurukar and Priyesh Vijayan and Srinivasan Parthasarathy and Balaraman Ravindran and Aakash Srinivasan and Goonmeet Bajaj and Chen Cai and Moniba Keymanesh and Saravana Kumar and Pranav Maneriker and Anasua Mitra and Vedang Patel},
  title = {Benchmarking and Analyzing Unsupervised Network Representation Learning and the Illusion of Progress},
  journal = {Transactions on Machine Learning Research},
  year = {2022}
}

@article{wang2022fastgcl,
  author = {Yuansheng Wang and Wangbin Sun and Kun Xu and Zulun Zhu and Liang Chen and Zibin Zheng},
  title = {{FastGCL}: Fast Self-Supervised Learning on Graphs via Contrastive Neighborhood Aggregation},
  journal = {arXiv preprint arXiv:2205.00905},
  year = {2022}
}

@article{mernyei2020wikics,
  author = {P{\'e}ter Mernyei and C{\u{a}}t{\u{a}}lina Cangea},
  title = {{Wiki-CS}: A {W}ikipedia-Based Benchmark for Graph Neural Networks},
  journal = {arXiv preprint arXiv:2007.02901},
  year = {2020}
}
\bibliographystyle{tmlr}

\clearpage
\appendix
\section{Protocol and provenance}
\label{app:protocol}

\begin{table}[!htbp]
  \centering
  \small
  \caption{Experiments. SSL training never sees a label, so E2, E7 and M re-evaluate the saved
  models of E1 without retraining. Seeds in parentheses: ogbn-arxiv (E1); GraphCL and AFGRL on
  Coauthor CS, whose 3{,}000-step runs exceeded a cost limit fixed with the design of E5.}
  \label{tab:design}
  \begin{tabularx}{\linewidth}{@{}l l >{\raggedright\arraybackslash}X r@{}}
    \toprule
    & Varies & Levels & Seeds \\
    \midrule
    E1 & objective $\times$ backbone & 8 objectives, untrained encoder, 2 supervised references (not on ogbn-arxiv); GCN, GIN, Transformer; 6 datasets and ogbn-arxiv & 10 (3) \\
    E2 & probe $\times$ split $\times$ labels & 4 linear probes (3 from other codebases, and ours), 3 controls, $k$NN on the standard split; our probe and $k$NN also on 10 random splits; ogbn-arxiv also with year labels & -- \\
    E3 & optimizer $\times$ learning rate & AdamW $\{10^{-4}, 5\cdot10^{-4}, 10^{-3}, 5\cdot10^{-3}, 10^{-2}\}$, SGD with Nesterov momentum $\{10^{-3}, 5\cdot10^{-3}, 10^{-2}, 5\cdot10^{-2}, 10^{-1}\}$; every backbone on Cora and CiteSeer, GCN on Amazon Computers, Amazon Photo and Coauthor CS, there with one more learning rate beyond the grid where a rule fixed in advance requires it & 5 \\
    E4 & training step & 14 steps from 0 to 300, in three optimizer cells; Cora, CiteSeer & 5 \\
    E5 & training budget & 300 and 3{,}000 steps, the longer runs also evaluated at 300, 600, 1{,}000 and 2{,}000; GCN on Cora and the Amazon and Coauthor graphs, GIN and Transformer on Cora & 5 (3) \\
    E6 & reference & logistic regression on the propagated features $\hat A^2X$ (fixed propagation); E1 and E2 on WikiCS, a dataset outside the six, to test it & 10 \\
    E7 & normalization statistics & E1 encoders on the six datasets and ogbn-arxiv evaluated with the statistics stored in their normalization layers, and with statistics recomputed from one pass over the graph & -- \\
    E8 & input features & row-normalized instead of raw features; GCN, Cora, CiteSeer, PubMed & 5 \\
    E9 & training labels & 1, 2, 5, 10 or 20 per class; saved encoders on the six main graphs and WikiCS; standard and reduced validation sets & 10 \\
    E10 & graph family & Roman-empire, Amazon-ratings and Minesweeper; all objectives and backbones; fixed-feature references and graph statistics & 10 \\
    E11 & joint configuration & 16 random configurations plus the shared recipe per objective; GCN on six graphs; validation selection, then five held-out training seeds & 5 \\
    M & -- & label-free metrics of every saved model & -- \\
    \bottomrule
  \end{tabularx}
\end{table}

\begin{table}[!htbp]
  \centering
  \small
  \caption{Shared protocol (E1, and the shared cell of E3 to E5).}
  \label{tab:protocol}
  \begin{tabularx}{\linewidth}{@{}l>{\raggedright\arraybackslash}X@{}}
    \toprule
    Encoders & GCN, GIN or graph Transformer (TransformerConv), 2 layers, 256 units, batch normalization, no dropout \\
    Training & full-batch, 300 steps (E5: also 3{,}000), AdamW, learning rate $5\cdot10^{-4}$, weight decay $10^{-5}$ \\
    Augmentations & edge dropping 0.5 and feature masking 0.2 in each view (GraphCL, BGRL, VICReg,
    Barlow Twins, GraphDINO); node shuffling (DGI); none (AFGRL); its own node masking (GraphMAE) \\
    Momentum encoders & BGRL and AFGRL: cosine schedule from 0.99 to 1; GraphDINO: 0.996 (reference DINO) \\
    GraphDINO head & teacher temperature 0.04, student temperature 0.1, center momentum 0.9 \\
    GraphMAE & mask rate 0.5, replacement rate 0.05, learned mask token, re-masking, bias-free linear map and one-layer GAT decoder, scaled cosine error with exponent 3 (reference configuration for Cora) \\
    Probes & see Section~\ref{sec:setup}; $k$NN with $k=5$ and cosine similarity \\
    Splits & public Planetoid split; random split 0 (10/10/80) as the standard split of Amazon Computers, Amazon Photo and Coauthor CS; 10 random 10/10/80 splits for the split factor; ogbn-arxiv official split; arxiv-year 5 random 50/25/25 splits \\
    Seeds & 10 per configuration (3 on ogbn-arxiv; 5 in E3--E5 and E8, with the E5 exceptions in Table~\ref{tab:design}) \\
    Supervised reference & dropout 0.5, Adam, learning rate 0.01 ($10^{-3}$ with GIN), weight decay $5\cdot10^{-4}$,
    best validation epoch; no normalization layer with GCN \\
    \bottomrule
  \end{tabularx}
\end{table}

Every run is saved as a JSON file with per-seed metrics, the exact model configuration, the git
commit and the package versions. Tables and figures are generated from these files by scripts, and
numerical claims are checked by a separate script that recomputes them from
the per-seed files without using the analysis code (\texttt{benchmarks/verify\_claims.py}). Numbers
of other papers are quoted from those papers.

We added GraphMAE during the study and introduced an optional recalculation of normalization
statistics for E7. The run metadata flags a modified working tree for 6 of the 204 E1 files,
11 of the 96 E5 files and most probe and metric files. On the same machine with the same thread
count, we reproduced two seeds from each of five flagged E1 runs, and all twelve sampled probe
scores, to the recorded precision. The sixth, AFGRL, is the exception: two reruns of the same code on the same
machine already differ from each other, and from the flagged file by up to 0.4 points. The same
reruns with a different number of threads give different numbers, by up to 2.2 points on one
seed of GraphDINO, because the probe selects another penalty on nearly identical embeddings.

\begin{table}[!htbp]
  \centering
  \small
  \caption{Datasets: size, raw edge homophily of the labels and the standard split. Edges are undirected pairs without self-loops. Raw homophily depends on class balance; Minesweeper has adjusted homophily near zero despite raw homophily of 0.68.}
  \label{tab:datasets}
\begin{tabular}{lrrrrrrrr}
\toprule
Dataset & Nodes & Edges & Features & Classes & Homophily & Train & Val. & Test \\
\midrule
Cora & 2{,}708 & 5{,}278 & 1{,}433 & 7 & 0.81 & 140 & 500 & 1{,}000 \\
CiteSeer & 3{,}327 & 4{,}552 & 3{,}703 & 6 & 0.74 & 120 & 500 & 1{,}000 \\
PubMed & 19{,}717 & 44{,}324 & 500 & 3 & 0.80 & 60 & 500 & 1{,}000 \\
Amazon Computers & 13{,}752 & 245{,}861 & 767 & 10 & 0.78 & 1{,}375 & 1{,}375 & 11{,}002 \\
Amazon Photo & 7{,}650 & 119{,}081 & 745 & 8 & 0.83 & 765 & 765 & 6{,}120 \\
Coauthor CS & 18{,}333 & 81{,}894 & 6{,}805 & 15 & 0.81 & 1{,}833 & 1{,}833 & 14{,}667 \\
WikiCS & 11{,}701 & 215{,}603 & 300 & 10 & 0.65 & 580 & 1{,}769 & 5{,}847 \\
Roman-empire & 22{,}662 & 32{,}927 & 300 & 18 & 0.05 & 11{,}331 & 5{,}665 & 5{,}666 \\
Amazon-ratings & 24{,}492 & 93{,}050 & 300 & 5 & 0.38 & 12{,}246 & 6{,}123 & 6{,}123 \\
Minesweeper & 10{,}000 & 39{,}402 & 7 & 2 & 0.68 & 5{,}000 & 2{,}500 & 2{,}500 \\
ogbn-arxiv & 169{,}343 & 1{,}157{,}799 & 128 & 40 & 0.65 & 90{,}941 & 29{,}799 & 48{,}603 \\
arxiv-year & 169{,}343 & 1{,}157{,}799 & 128 & 5 & 0.22 & 84{,}672 & 42{,}336 & 42{,}335 \\
\bottomrule
\end{tabular}

\end{table}

On ogbn-arxiv, the 50-epoch runs use two-hop neighbor sampling with fan-out $10 \times 10$
and 1{,}024 seed nodes per batch. Every node serves as a seed once per epoch. The budget was
chosen from a two-epoch timing pilot covering all objectives and backbones.

\paragraph{Objectives.} We use \emph{objective} to denote a pretraining method together with the heads, augmentations
and other components it requires. The eight include DGI, which detects shuffled node features
\citep{velickovic2019dgi}, and GraphCL, which contrasts two augmented views
\citep{you2020graphcl}. Our node-level GraphCL uses the contrastive loss of GRACE
\citep{zhu2020grace}. BGRL and AFGRL learn to predict a moving target encoder, with different
rules for forming positive pairs \citep{thakoor2022bgrl,lee2022afgrl}. VICReg and Barlow Twins
reduce redundancy between embedding dimensions \citep{bardes2022vicreg,zbontar2021barlow}.
GraphDINO adapts DINO's teacher--student objective to graphs \citep{caron2021dino}, and GraphMAE
reconstructs masked node features with a graph attention decoder \citep{hou2022graphmae,velickovic2018gat}.

\paragraph{Probe implementations.} On the six main datasets we use the following protocols (Table~\ref{tab:audit}).
\begin{itemize}
  \item \emph{100 Adam steps} (DGI, PyG-SSL): a linear layer with Xavier initialization trained
  full-batch with Adam \citep{kingma2015adam} (learning rate 0.01, no weight decay) on the raw
  embeddings; test accuracy after the last step, averaged over 20 initializations.
  \item \emph{5{,}000 Adam steps} (PyGCL; GCA uses the same procedure with 3{,}000): the same
  layer and optimizer for 5{,}000 steps, test accuracy at the best validation accuracy, checked
  every 20 steps, averaged over three probe initializations.
  \item \emph{liblinear} (BGRL): rows L2-normalized, one-vs-rest logistic regression with
  liblinear \citep{fan2008liblinear}, $C \in \{2^{-10}, \dots, 2^{10}\}$ chosen on validation.
  \item \emph{Regularized} (ours): embeddings standardized with the statistics of the training
  rows, multinomial logistic regression with an L2 penalty $\frac{\lambda}{2}\lVert W \rVert^2$,
  fitted with L-BFGS \citep{liu1989lbfgs}. We initialize the first fit at zero and warm-start
  each weaker penalty from the preceding fit. Each fit has a budget of 5{,}000 iterations and
  loss evaluations. Validation accuracy selects $\lambda \in \{10, 1, \dots, 10^{-6}\}$,
  with ties going to the stronger penalty. Unless stated otherwise, results use this probe.
  \item \emph{$k$NN}: cosine similarity, majority vote of the $k=5$ nearest training nodes.
\end{itemize}
Three controls change one ingredient at a time: the two Adam probes on standardized embeddings,
and the regularized probe with a negligible penalty ($\lambda = 10^{-6}$). All of them score the
standard split; the regularized probe and $k$NN also score ten random 10/10/80 splits, and on the
Amazon and Coauthor graphs the standard split is the first of the ten and is counted once.

\begin{table}[!htbp]
  \centering
  \small
  \caption{Linear-evaluation protocols in reference implementations of graph SSL, read from their
  code on 2026-10-24 (linked revisions). The selection column describes the code path for the indicated task and split.}
  \label{tab:audit}
  \begin{tabularx}{\linewidth}{@{}l >{\raggedright\arraybackslash}p{4.1cm} >{\raggedright\arraybackslash}X >{\raggedright\arraybackslash}X@{}}
    \toprule
    Codebase & Classifier and budget & Features and regularization & Model selection \\
    \midrule
    DGI (\coderevision{PetarV-/DGI}{61baf67}{execute.py}) & Adam, 100 steps, 50 runs & raw; no weight decay & last step \\
    PyG-SSL (\coderevision{iDEA-iSAIL-Lab-UIUC/pyg-ssl}{7ab0f64}{src/evaluation/logistic_regression.py}) & Adam, 100 steps, 50 runs by default & raw; no weight decay & last step; without masks, validation = test \\
    DIG (\coderevision{divelab/DIG}{21476b0}{dig/sslgraph/evaluation/eval_node.py}) & Adam, 300 steps & raw; no weight decay & pretraining epoch and learning rate by \textbf{test} accuracy \\
    GraphMAE (\coderevision{THUDM/GraphMAE}{b14f080}{graphmae/evaluation.py}) & Adam, 300 steps (600 on ogbn-arxiv) & raw; weight decay fixed per dataset & last step and best validation, both reported \\
    PyGCL (\coderevision{PyGCL/PyGCL}{aa238b3}{GCL/eval/logistic_regression.py}) & Adam, 5{,}000 steps & raw; no weight decay & best validation, every 20 steps \\
    GCA (\coderevision{CRIPAC-DIG/GCA}{0f17497}{train.py}) & Adam, 3{,}000 steps (800 on WikiCS) & raw; no weight decay & best validation, every 20 steps \\
    GRACE (\coderevision{CRIPAC-DIG/GRACE}{b3b5ac3}{eval.py}) & liblinear, one-vs-rest & L2-normalized; $C$ by 5-fold cross-validation & cross-validation on a random 10\% training split \\
    BGRL (\coderevision{nerdslab/bgrl}{60f9f19}{bgrl/logistic_regression_eval.py}) & liblinear, one-vs-rest & L2-normalized; $C$ on validation (WikiCS) or by cross-validation & validation (WikiCS); otherwise cross-validation on a random 20\% split \\
    \midrule
    Ours & L-BFGS, up to 5{,}000 iterations & standardized; $\lambda$ on validation & validation \\
    \bottomrule
  \end{tabularx}
\end{table}

\paragraph{Validating the implementation.}
With a 256-unit encoder and supervised training adapted from \citet{kipf2017gcn},
our classification head reaches 81.4, 69.5 and
78.4\% on Cora, CiteSeer and PubMed, 0.1 to 0.8 points below the published 81.5, 70.3 and 79.0.
With the BGRL pretraining recipe and our regularized probe on ogbn-arxiv, our BGRL reaches 71.3\% test accuracy over
three seeds against the published 71.6, and the same untrained encoder 69.5 against 68.9
(Appendix~\ref{app:arxiv}). GraphMAE follows its reference code, and the test suite checks each
of its components (masking, re-masking before the decoder, the loss on the masked nodes against
the clean features); each check was verified to fail when its component is removed. Every
objective is trained with every backbone in the test suite of the library, which also checks
that edge features reach the encoder and that saved models reproduce their embeddings.

\paragraph{Hardware.}
For encoder training in E1 we use a laptop CPU for the citation and
Amazon graphs, its GPU for Coauthor CS, and an NVIDIA A2 GPU for ogbn-arxiv and for E3 on Cora and
CiteSeer. E3 on the Amazon and Coauthor graphs runs on the device of their E1 runs, where its
shared cell reproduces every E1 seed to within 0.07 points on the Amazon graphs (laptop CPU) and
0.16 on Coauthor CS (laptop GPU); E5 runs on the laptop GPU (Cora, Amazon Photo) and on the A2
(Amazon Computers, Coauthor CS); E7 and E8 run on the laptop CPU (E7 on ogbn-arxiv on its GPU),
and E7 scores both kinds of statistics in the same runs. The device is therefore confounded with
the dataset. Appendix~\ref{sec:seeds} measures what a change of device does to the same seeds:
the mean absolute accuracy difference ranges from 0.59 to 1.88 points across the four
dataset--backbone pairs, with ranking agreement from 0.71 to 1.00. The runs span several commits, since the study
grew while it ran (above).

\FloatBarrier
\section{Analyses fixed in advance}
\label{app:prereg}

Table~\ref{tab:prereg} lists the analyses reported here whose rules were committed before the results they
apply to, with the commit that fixed them, the time the first of those results was written, and
what had been seen by then. The prespecified rules are reproduced in
\texttt{benchmarks/paper/PREREG.md}. They are the authors' own record, not an external registration:
the commits belong to a development history that also holds private notes and is not part of the
released repository. The rules of E9, E10 and E11 are evaluated in Section~\ref{sec:stress}
and detailed in Appendices~\ref{app:e9}, \ref{app:e10} and~\ref{app:e11}.

{\small
\begin{longtable}{@{}>{\raggedright\arraybackslash}p{2.2cm} >{\raggedright\arraybackslash}p{1.7cm} l >{\raggedright\arraybackslash}p{1.7cm} >{\raggedright\arraybackslash}p{7.9cm}@{}}
  \caption{Analyses fixed in advance. Times in UTC; commits of the development history.}
  \label{tab:prereg}\\
    \toprule
    Analysis & Fixed & Commit & First data & Seen before \\
    \midrule
    \endfirsthead
    \toprule
    Analysis & Fixed & Commit & First data & Seen before \\
    \midrule
    \endhead
    Label-free prediction test & 2026-10-26 04:05 & \texttt{0bd9039} & 2026-10-27 19:37 & development datasets (Cora, CiteSeer); earlier benchmark runs of seven objectives on PubMed (GCN, GIN: accuracy, untrained encoder, effective rank) and one GIN seed on ogbn-arxiv; effective rank of DGI with a GCN on PubMed and Amazon Photo. No CKA distance or smoothing on a test dataset \\
    E4 trajectory definitions & 2026-10-27 05:03 & \texttt{2c83ccc} & 2026-10-27 05:06 & E3 endpoints, one pilot trajectory \\
    E5 decision rule (budget) & 2026-10-28 06:33 & \texttt{63396cc} & 2026-10-28 06:36 & E1 and E4 (300 steps); 3{,}000-step ablations of BGRL, Barlow Twins and GraphDINO on Cora and CiteSeer; nothing beyond 300 steps on the Amazon and Coauthor graphs \\
    E3 decision rule (Amazon, Coauthor) & 2026-10-29 09:03 & \texttt{9263d36} & 2026-10-29 09:12 & E3 on Cora and CiteSeer; E1 on the three graphs \\
    E6 test on WikiCS & 2026-11-13 17:16 & \texttt{6cf6db9} & 2026-11-13 18:03 & everything on the six datasets and ogbn-arxiv, including the comparison with $\hat A^2X$; FastGCL's results on WikiCS; a 3-step run of two objectives and, after the commit, 30-step timing runs of two objectives on WikiCS (76.6 to 79.1\% under our probe). No input-feature row and no 300-step model on WikiCS \\
    E7 rules (normalization statistics) & 2026-11-19 14:46 & \texttt{f0fb47d} & 2026-11-19 15:09 & everything above; the untrained encoders alone under both statistics on Cora, CiteSeer (seeds 0--2), Amazon Computers and Coauthor CS (seeds 0--1), in a diagnostic that motivated E7. No trained model under recomputed statistics, nothing on PubMed, Amazon Photo or ogbn-arxiv; after the commit, a preflight of two checkpoints per dataset \\
    E8 rule (row-normalized features) & 2026-11-19 14:46 & \texttt{f0fb47d} & 2026-11-19 14:54 & no model trained on row-normalized features under the shared protocol; SGC's published accuracies on $\hat A^2X$ \\
    E9 label budgets & 2026-11-26 06:03 & \texttt{8afa4a5} & 2026-11-26 07:13 & E1--E8 and a Cora smoke run with three label draws. Its scaling sensitivity motivated the all-node standardization option. The PCA comparison was planned before the smoke run. \\
    E10 heterophilous graphs & 2026-11-26 06:03 & \texttt{8afa4a5} & 2026-11-26 08:58 & E1--E8; dataset statistics and per-step timings, with no score of any encoder; the input-feature rows of Tolokers in a smoke run; the published results of \citet{platonov2023heterophily} \\
    E11 joint tuning & 2026-11-26 06:03 & \texttt{8afa4a5} & 2026-11-26 06:56 & every result of E1--E8 on the six graphs; per-step timings; a first draw with widths up to 1{,}024, replaced before any real run after it ran out of GPU memory in a 3-step preflight, with no score read \\
    \bottomrule
\end{longtable}
}

\FloatBarrier
\section{E6: the propagated features, and WikiCS}
\label{app:e6}

\begin{table}[htb]
  \centering\small
  \caption{Test accuracy (\%) for the four representations, with the validation rule of
  Figure~\ref{fig:propagation}. The pretrained columns give the mean and range of the eight
  objective means over ten seeds. The maximum describes the observed test scores and would
  require a separate selection rule for deployment. The last column counts objective means
  strictly above propagation before rounding.}
  \label{tab:absolute}
  \begin{tabular}{@{}lrrrrrrr@{}}
\toprule
 & \multicolumn{2}{c}{Fixed features} & & \multicolumn{3}{c}{Pretrained GCN} & \\
\cmidrule(lr){2-3}\cmidrule(lr){5-7}
Dataset & Raw & $\hat A^2 X$ & Untrained GCN & Mean & Lowest & Highest & Above $\hat A^2 X$ \\
\midrule
Cora & 60.2 & 82.0 & 73.2 & 80.1 & 77.4 & 82.1 & 1 of 8 \\
CiteSeer & 62.4 & 72.3 & 59.1 & 66.8 & 61.2 & 69.8 & 0 of 8 \\
PubMed & 72.9 & 79.7 & 73.9 & 79.5 & 76.7 & 81.3 & 3 of 8 \\
Amazon Computers & 79.9 & 89.5 & 87.2 & 88.5 & 87.8 & 89.1 & 0 of 8 \\
Amazon Photo & 86.8 & 92.8 & 91.9 & 92.2 & 91.8 & 92.5 & 0 of 8 \\
Coauthor CS & 92.1 & 93.1 & 91.1 & 92.4 & 91.9 & 93.2 & 1 of 8 \\
\midrule
WikiCS & 72.2 & 80.2 & 78.8 & 79.4 & 78.5 & 80.1 & 0 of 8 \\
\bottomrule
\end{tabular}

\end{table}

\paragraph{Why it was added.} After the results of the six datasets were in, we scored their input
features three more ways: propagated once and twice over the normalized adjacency ($\hat A X$,
$\hat A^2X$) and as the first 256 principal components of $\hat A^2X$, which has the width of the
encoders and uses no labels. Every input-feature row and every model on these six datasets is scored with our probe and
with liblinear, and the comparison uses the one better on validation for each representation
(ties favor ours). On sparse bag-of-words features, solver and preprocessing choices make a
substantial difference. The raw features of Cora, CiteSeer and PubMed reach 52.6, 47.5 and 67.0\% under our
probe and 60.2, 62.4 and 72.9\% under liblinear. Those numbers, and the comparison of
Table~\ref{tab:propagation} on the six datasets, were obtained after the fact. The initial
planning calculation conflated trained and untrained BGRL rows with the same name. The
recorded correction identifies each model by checkpoint. It changes the exploratory values
quoted in the plan and leaves the WikiCS rule intact.

\begin{table}[!htbp]
  \centering
  \caption{Fixed propagation followed by a fitted linear classifier. Test accuracy (\%), standard split, each row scored with
  the regularized probe (ours or liblinear) that is better on validation: logistic regression on
  the raw features, on the features propagated twice over the normalized adjacency ($\hat A^2X$)
  and on their first 256 principal components; the untrained GCN and the eight GCN objectives
  (mean, and range over objectives) over 10 seeds. $D$: the average GCN objective minus $\hat
  A^2X$, with its original 95\% interval (seeds resampled within each objective). Last column: trained models of any backbone above
  $\hat A^2X$. ogbn-arxiv and its year labels (arxiv-year): our probe only, three seeds; the
  dense features of ogbn-arxiv have fewer than 256 dimensions, so no PCA. WikiCS is the prespecified follow-up. The ogbn-arxiv feature rows use CPU and encoder rows use the A2 GPU.}
  \label{tab:propagation}
  \resizebox{\linewidth}{!}{\begin{tabular}{@{}lrrrrrrr@{}}
\toprule
 & \multicolumn{3}{c}{Fixed features} & \multicolumn{3}{c}{GCN} & Any backbone \\
\cmidrule(lr){2-4}\cmidrule(lr){5-7}\cmidrule(l){8-8}
Dataset & Raw & $\hat A^2X$ & PCA-256 & Untrained & Objectives & $D$ [95\% CI] & Above $\hat A^2X$ \\
\midrule
Cora & 60.2 & 82.0 & 81.5 & 73.2 & 80.1 (77.4--82.1) & $-1.9$ $[-2.1, -1.7]$ & 1 of 24 \\
CiteSeer & 62.4 & 72.3 & 72.2 & 59.1 & 66.8 (61.2--69.8) & $-5.5$ $[-5.8, -5.2]$ & 0 of 24 \\
PubMed & 72.9 & 79.7 & 79.7 & 73.9 & 79.5 (76.7--81.3) & $-0.2$ $[-0.4, +0.0]$ & 5 of 24 \\
Amazon Computers & 79.9 & 89.5 & 88.0 & 87.2 & 88.5 (87.8--89.1) & $-1.0$ $[-1.1, -1.0]$ & 0 of 24 \\
Amazon Photo & 86.8 & 92.8 & 92.5 & 91.9 & 92.2 (91.8--92.5) & $-0.6$ $[-0.6, -0.5]$ & 3 of 24 \\
Coauthor CS & 92.1 & 93.1 & 93.1 & 91.1 & 92.4 (91.9--93.2) & $-0.7$ $[-0.7, -0.7]$ & 7 of 24 \\
ogbn-arxiv & 52.5 & 69.2 & -- & 69.2 & 68.2 (63.6--69.5) & $-1.0$ $[-1.1, -1.0]$ & 3 of 24 \\
arxiv-year & 34.9 & 39.0 & -- & 39.7 & 42.1 (41.8--42.4) & $+3.1$ $[+3.1, +3.2]$ & 17 of 24 \\
\midrule
WikiCS & 72.2 & 80.2 & 78.7 & 78.8 & 79.4 (78.5--80.1) & $-0.8$ $[-0.9, -0.8]$ & 0 of 24 \\
\bottomrule
\end{tabular}
}
\end{table}

\begin{figure}[!htbp]
  \centering
  \includegraphics[width=\linewidth]{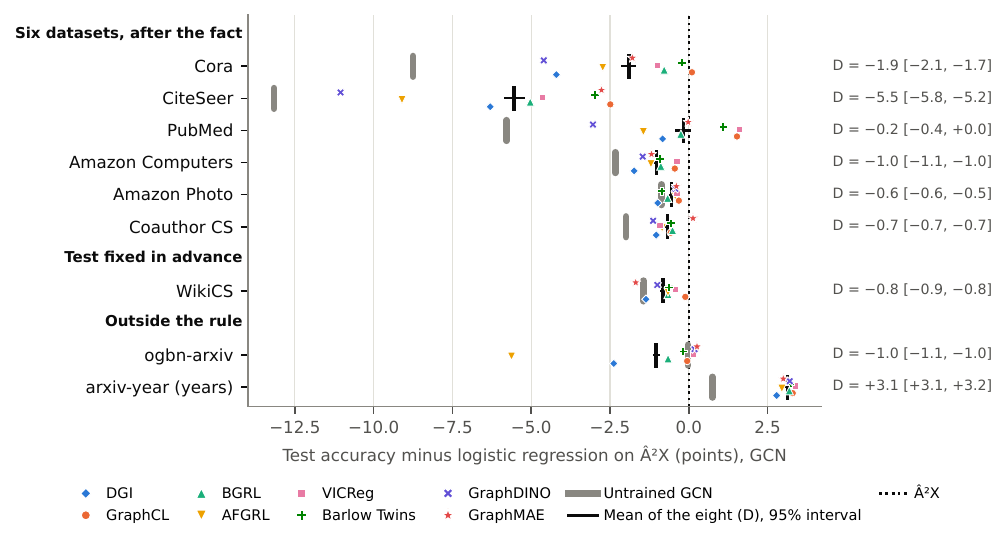}
  \caption{Individual GCN objectives under the probe-selection rule of
  Figure~\ref{fig:propagation} on the six main graphs and WikiCS. Both ogbn-arxiv tasks use
  only the default standardized probe and three training seeds. Each marker is one objective's mean accuracy relative to
  propagated features. The gray bar is the untrained GCN and the black tick is the average
  across objectives with its 95\% seed-bootstrap interval.}
  \label{fig:individual-propagation}
\end{figure}

\begin{table}[t]
  \centering
  \caption{GCN test accuracy minus accuracy on $\hat A^2X$, in percentage points. The first
  seven columns average over eight objectives, or seven objectives when GraphDINO is excluded. Dashes mark settings we did
  not run. The last two columns give the best individual objective under the shared protocol
  and a supervised GCN trained with the standard recipe. E3 selects the optimizer and learning
  rate on validation, and E5 selects a checkpoint from a 3{,}000-step run on validation. Both
  use our default probe for the models and the validation-selected fixed reference. E11 selects
a joint configuration per objective on validation, five seeds, with the validation-selected
probe for every representation (Appendix~\ref{app:e11}). In E1 the
  six dataset means for pretrained GCNs differ by less than half a point between the two
  regularized probes. This empirical bound concerns E1. E7 recomputes normalization statistics. E8 applies
  row normalization to the inputs of both pipelines.}
  \label{tab:checks}
  \footnotesize
  \setlength{\tabcolsep}{3.2pt}
  \resizebox{\linewidth}{!}{%
\begin{tabular}{@{}lrrrrrrrrr@{}}
\toprule
 & \multicolumn{7}{c}{$D$: average GCN objective minus $\hat A^2X$} & \multicolumn{2}{c}{Minus $\hat A^2X$} \\
\cmidrule(lr){2-8}\cmidrule(l){9-10}
Dataset & \begin{tabular}[b]{@{}c@{}}Shared\\protocol\end{tabular} & \begin{tabular}[b]{@{}c@{}}Without\\GraphDINO\end{tabular} & \begin{tabular}[b]{@{}c@{}}Graph statistics\\(E7)\end{tabular} & \begin{tabular}[b]{@{}c@{}}Row-normalized $X$\\(E8)\end{tabular} & \begin{tabular}[b]{@{}c@{}}Tuned\\(E3)\end{tabular} & \begin{tabular}[b]{@{}c@{}}3{,}000 steps\\(E5)\end{tabular} & \begin{tabular}[b]{@{}c@{}}Joint search\\(E11)\end{tabular} & \begin{tabular}[b]{@{}c@{}}Best\\objective\end{tabular} & \begin{tabular}[b]{@{}c@{}}Supervised\\GCN\end{tabular} \\
\midrule
Cora & $-$1.9 & $-$1.5 & $-$1.9 & $-$1.0 & $-$1.0 & $-$2.5 & $-$0.0 & +0.1 & $-$0.9 \\
CiteSeer & $-$5.5 & $-$4.8 & $-$5.6 & $-$5.0 & $-$2.7 & -- & $-$2.4 & $-$2.5 & $-$2.8 \\
PubMed & $-$0.2 & +0.2 & $-$0.1 & $-$0.3 & -- & -- & +1.3 & +1.6 & $-$1.3 \\
Amazon Computers & $-$1.0 & $-$1.0 & $-$1.1 & -- & $-$0.9 & $-$0.6 & $-$0.5 & $-$0.4 & +0.0 \\
Amazon Photo & $-$0.6 & $-$0.6 & $-$0.6 & -- & $-$0.7 & $-$0.6 & $-$0.3 & $-$0.3 & +0.1 \\
Coauthor CS & $-$0.7 & $-$0.6 & $-$0.7 & -- & $-$0.7 & $-$0.8 & $-$0.0 & +0.1 & $-$0.6 \\
\bottomrule
\end{tabular}
}
\end{table}
\begin{table}[!htbp]
  \centering\small
  \caption{Sensitivity of the propagation comparison to the probe. Each protocol scores both
  the features and the frozen GCNs. Selected chooses between the default standardized probe
  and row-normalized liblinear by validation accuracy for each representation. Pretrained
  is the equally weighted mean over the eight objectives.}
  \label{tab:probe-sensitivity}
  \begin{tabular}{@{}lrrr@{\hspace{1.2em}}rr@{\hspace{1.2em}}rr@{}}
\toprule
 & \multicolumn{3}{c}{$D$ (points)} & \multicolumn{2}{c}{$\hat A^2X$ (\%)} & \multicolumn{2}{c}{Pretrained (\%)} \\
\cmidrule(lr){2-4}\cmidrule(lr){5-6}\cmidrule(lr){7-8}
Dataset & Selected & Liblinear & Default & Liblinear & Default & Liblinear & Default \\
\midrule
Cora & $-$1.9 & $-$2.1 & +8.5 & 82.0 & 71.5 & 79.9 & 80.0 \\
CiteSeer & $-$5.5 & $-$5.6 & +19.5 & 72.3 & 46.8 & 66.7 & 66.3 \\
PubMed & $-$0.2 & $-$0.3 & +4.6 & 79.7 & 75.1 & 79.4 & 79.7 \\
Amazon Computers & $-$1.0 & $-$1.1 & $-$0.5 & 89.5 & 88.9 & 88.4 & 88.5 \\
Amazon Photo & $-$0.6 & $-$0.6 & $-$1.0 & 92.8 & 92.9 & 92.2 & 91.9 \\
Coauthor CS & $-$0.7 & $-$0.7 & $-$0.1 & 93.1 & 92.3 & 92.4 & 92.2 \\
WikiCS & $-$0.8 & $-$0.8 & $-$0.6 & 80.2 & 79.4 & 79.4 & 78.8 \\
\bottomrule
\end{tabular}

\end{table}

\paragraph{The test.} Before the full training runs, we specified a test on WikiCS \citep{mernyei2020wikics}
(11{,}701 nodes, dense 300-dimensional features, edge homophily 0.65; its public split 0 as the
standard split), with E1 and E2 there as on the other datasets (eight objectives, the untrained
encoder and the supervised references, three backbones, 10 seeds, all on one laptop CPU). The
quantity is $D$, the average over the eight GCN objectives of the seed-mean test accuracy minus
that of $\hat A^2X$, with a seed bootstrap; $K=2$ was fixed, not chosen. If $D \ge 1$ point with
an interval above zero, the comparison is reported for the six datasets only, with WikiCS as the
counterexample. WikiCS also enters the later label-budget analysis (E9).

\paragraph{Result.} $D = -0.8$ points, with a 95\% interval of $-0.9$ to $-0.8$: the comparison
favors propagation on this additional graph (Table~\ref{tab:e6}). None of the eight GCN
objectives is above $\hat A^2X$; the closest, GraphCL, is 0.1 points below it, and GraphMAE is
also below the untrained GCN. Nor is any of the 24 trained models of the three backbones
(Table~\ref{tab:propagation}): the best with a GIN, GraphCL, reaches 78.7\%, and the best with the
Transformer, AFGRL, 79.2\%. $D$ is $-0.6$ under our probe alone, $-0.8$ under liblinear, and
$-0.5$ on the ten random splits under our probe. The supervised GCN trained with the standard
recipe is 0.2 points below $\hat A^2X$ as well. Training adds 0.6 points to the GCN on average,
as little as on the Amazon and Coauthor graphs. On these dense features the two probes nearly
agree on the raw features (71.9 and 72.2\%, against 47.5 and 62.4\% on CiteSeer), so the probe choice has a much smaller effect on the raw-feature baseline here. The rule
does not cover the 100-step probe: under it $\hat A^2X$ reaches only 74.4\% and every objective is
above it ($D = +4.3$), so this comparison, like the gains of Section~\ref{sec:probe}, depends on the probe protocol.

\begin{table}[!htbp]
  \centering
  \caption{WikiCS, the quantities of the E6 rule. Test accuracy (\%), GCN, mean over 10 seeds
  ($\pm$ standard deviation under the probe chosen on validation, the probe of the rule); the
  input-feature rows have no seed. $-\,\hat A^2X$: difference from $\hat A^2X$ under that probe.
  Last column: our probe on ten random 10/10/80 splits. $\dagger$~Encoders trained on the labels
  of the standard split, not ranked; their random-split accuracy is not shown, since those splits
  test on nodes whose labels the encoder saw.}
  \label{tab:e6}
  \begin{tabular}{@{}lrrrrr@{}}
\toprule
 & \multicolumn{4}{c}{Standard split} & 10 random splits \\
\cmidrule(lr){2-5}\cmidrule(l){6-6}
Representation & Ours & liblinear & Better on val. & $-\,\hat A^2X$ & Ours \\
\midrule
Raw features & 71.9 & 72.2 & 72.2 & $-8.0$ & 74.3 \\
$\hat AX$ & 78.0 & 78.8 & 78.8 & $-1.4$ & 79.6 \\
$\hat A^2X$ & 79.4 & 80.2 & 80.2 & -- & 80.3 \\
PCA-256 of $\hat A^2X$ & 73.2 & 78.7 & 78.7 & $-1.5$ & 76.8 \\
\midrule
Untrained GCN & 78.3 & 78.8 & 78.8 $\pm$ 0.4 & $-1.4$ & 79.3 \\
DGI & 78.2 & 78.9 & 78.9 $\pm$ 0.4 & $-1.4$ & 79.5 \\
GraphCL & 79.4 & 80.1 & 80.1 $\pm$ 0.3 & $-0.1$ & 80.2 \\
BGRL & 78.8 & 79.6 & 79.6 $\pm$ 0.3 & $-0.7$ & 79.8 \\
AFGRL & 78.9 & 79.6 & 79.6 $\pm$ 0.2 & $-0.7$ & 79.7 \\
VICReg & 79.3 & 79.9 & 79.8 $\pm$ 0.3 & $-0.4$ & 80.2 \\
Barlow Twins & 79.0 & 79.6 & 79.6 $\pm$ 0.2 & $-0.6$ & 79.8 \\
GraphDINO & 78.8 & 79.2 & 79.2 $\pm$ 0.4 & $-1.0$ & 79.8 \\
GraphMAE & 78.2 & 78.5 & 78.5 $\pm$ 0.3 & $-1.7$ & 79.3 \\
\midrule
Supervised$^\dagger$ & 74.8 & 75.1 & 75.1 $\pm$ 0.5 & $-5.1$ & -- \\
Supervised, std.\ recipe$^\dagger$ & 79.6 & 80.1 & 80.0 $\pm$ 0.2 & $-0.2$ & -- \\
\bottomrule
\end{tabular}

\end{table}

\paragraph{Planned on the A2, run on a laptop.} The rule placed WikiCS on the A2 GPU. Its preflight
showed every run dominated by the evaluation on the CPU of the shared machine (90 to 118 seconds per
3-step model, against 1.7 on the laptop), so before any real run WikiCS was moved whole to the
laptop CPU; the rule was not changed.

\FloatBarrier
\section{E9: label budgets}
\label{app:e9}

\paragraph{Design.} The seven graphs with node-level models of the main grid, the six main
graphs and WikiCS, are refit without retraining with $K\in\{1,2,5,10,20\}$ training labels per
class, drawn from the nodes outside the standard validation and test sets. Ten draws are
nested across $K$: draw $i$ permutes each class with a fixed seed and takes its first $K$
nodes. The model of seed $s$ is scored on draw $s$, the input-feature rows on every draw, and
the test set is the standard one. For each representation and fit, the standard validation
set chooses among our default probe, the same probe standardized with the statistics of all
nodes, and liblinear. The all-node option was added after a smoke run on Cora showed that our
probe, which standardizes with the training rows, breaks down with one label per class
(Table~\ref{tab:prereg}). Supervised references are excluded, since their encoders saw the
standard training labels.

\paragraph{Rule and result.} For each graph and budget, $D$ is the mean over the eight GCN
objectives of the difference from $\hat A^2X$ on the same draw, with a joint bootstrap over
the ten seed--draw pairs. A budget would have qualified the comparison if at least four of the
seven graphs favored pretraining. No budget does (Table~\ref{tab:budgets}), so the planned
attribution test against PCA-256 of $\hat A^2X$ did not apply. The last row of the table,
reported without a rule, limits the validation set to the same per-class budget.

\begin{table}[!htbp]
  \centering\small
  \caption{Mean pretrained GCN minus twice-propagated features, in accuracy points.
  Bold marks a difference of at least one point with a joint 95\% interval above zero
  (2,000 resamples of ten seed--draw pairs). The last row repeats the count with the
  validation set also limited to the indicated per-class budget.}
  \label{tab:budgets}
\begin{tabular}{@{}lrrrrr@{}}
\toprule
Training labels per class & 1 & 2 & 5 & 10 & 20 \\
\midrule
Cora & \textbf{+8.7} & +1.1 & $-$1.8 & $-$1.5 & $-$1.2 \\
CiteSeer & $-$2.6 & $-$3.2 & $-$7.0 & $-$6.8 & $-$5.9 \\
PubMed & +2.4 & +1.7 & +1.2 & +1.2 & \textbf{+1.2} \\
Amazon Computers & $-$2.8 & +0.2 & $-$4.2 & $-$2.9 & $-$1.4 \\
Amazon Photo & $-$2.8 & $-$2.2 & $-$2.2 & $-$1.7 & $-$1.0 \\
Coauthor CS & \textbf{+2.0} & \textbf{+1.7} & $-$0.2 & $-$0.6 & $-$0.7 \\
WikiCS & $-$2.0 & $-$2.3 & $-$3.6 & $-$3.0 & $-$2.1 \\
\midrule
Graphs meeting rule & 2 & 1 & 0 & 0 & 1 \\
With small validation set & 2 & 2 & 2 & 1 & 1 \\
\bottomrule
\end{tabular}

\end{table}

\FloatBarrier
\section{E10: heterophilous graphs}
\label{app:e10}

\paragraph{Design.} Roman-empire, Amazon-ratings and Minesweeper, as distributed by PyTorch
Geometric \citep{platonov2023heterophily}, with split 0 of their ten fixed splits as the
standard split. The shared protocol is unchanged: eight objectives, three backbones, ten
seeds, and the untrained encoders. No supervised references were trained, since
\citet{platonov2023heterophily} report them. Tolokers and Questions were left out for cost,
from timings and before any score: GraphCL's and AFGRL's quadratic cost on the 49{,}000 nodes
of Questions, and GPU memory on Tolokers. Every encoder and input row is scored with our probe
and liblinear on split 0, selected on validation as in E6, with our probe on the ten fixed
splits, and with our probe after recomputing the normalization statistics on the graph.

\paragraph{Rules and result.} Q1 asks whether the GCN comparison with $\hat A^2X$ extends to
these graphs. It would not if two of the three graphs favored pretraining. Only Roman-empire
does, by 1.4 points. Q2 asks whether pretraining adds beyond the validation-selected fixed-feature reference
$R^*$, the input row among $X$, $\hat A^2X$ and $[X,\hat AX,\hat A^2X]$ with the best
validation score. It would for a backbone whose average objective favored pretraining on two
of the three graphs. No backbone meets the one-point criterion on any graph (Table~\ref{tab:e10}); the
Transformer on Minesweeper is $0.44$ ROC-AUC points ahead, with interval $[0.40,0.47]$. This
is a positive difference, but smaller than the prespecified margin. Its gain over the
untrained Transformer is $-0.47$ points, with interval $[-0.51,-0.43]$. On the ten
fixed splits the differences keep their signs under the default probe. This sensitivity
analysis retains the fixed-feature family selected on split 0; it does not select a new
$R^*$ on each split.

\paragraph{Training gains.} Reported without a rule. With stored statistics the mean gain is
at most two points for every backbone and graph, and negative for the GIN on Roman-empire and
Minesweeper and for the Transformer on both. With statistics recomputed on the graph every GIN
gain is positive, $+10.3$ points on Roman-empire, because the untrained GIN moves and the
trained ones barely do (Section~\ref{sec:probe}).

\begin{table}[!htbp]
  \centering\small
  \caption{E10. Input rows (above each block; bold: $R^*$, chosen on validation) and, per
  backbone, the untrained encoder and the eight objectives (mean of their ten-seed means and
  range), with the probe chosen on validation for each representation. $-\,R^*$: average
  objective minus $R^*$. Gains use our probe, with stored or graph-recomputed normalization
  statistics. Last column: our probe averaged over the ten fixed splits. Minesweeper: ROC-AUC
  points.}
  \label{tab:e10}
  \footnotesize
\begin{tabular}{@{}llrrrrrrr@{}}
\toprule
 & & \multicolumn{4}{c}{Probe chosen on validation} & \multicolumn{2}{c}{Gain, our probe} & 10 splits \\
\cmidrule(lr){3-6}\cmidrule(lr){7-8}\cmidrule(l){9-9}
Graph & Backbone & Untrained & Trained & Range & $-\,R^*$ & Stored & Graph & $-\,R^*$ \\
\midrule
\multicolumn{9}{@{}l}{\emph{Roman-empire}: $X$ 64.5, $\hat AX$ 45.0, $\hat A^2X$ 44.8, \textbf{$[X,\hat AX,\hat A^2X]$ 76.0}} \\
 & GCN & 45.1 & 46.1 & 44.2--47.4 & $-$29.9 & +1.0 & +1.2 & $-$28.7 \\
 & GIN & 69.0 & 55.3 & 45.2--66.3 & $-$20.7 & $-$13.7 & +10.3 & $-$19.8 \\
 & Transformer & 71.0 & 70.5 & 70.0--71.4 & $-$5.6 & $-$0.1 & +1.5 & $-$4.4 \\
\multicolumn{9}{@{}l}{\emph{Amazon-ratings}: $X$ 39.1, $\hat AX$ 41.0, \textbf{$\hat A^2X$ 42.9}, $[X,\hat AX,\hat A^2X]$ 41.8} \\
 & GCN & 41.5 & 42.1 & 41.7--42.8 & $-$0.8 & +0.3 & +0.9 & $-$0.8 \\
 & GIN & 39.7 & 41.4 & 39.8--42.9 & $-$1.5 & +1.7 & +1.0 & $-$1.3 \\
 & Transformer & 39.5 & 39.9 & 39.3--40.2 & $-$3.0 & +0.4 & +0.2 & $-$2.7 \\
\multicolumn{9}{@{}l}{\emph{Minesweeper} (ROC-AUC): $X$ 52.6, $\hat AX$ 81.2, $\hat A^2X$ 71.1, \textbf{$[X,\hat AX,\hat A^2X]$ 89.4}} \\
 & GCN & 71.0 & 71.1 & 71.0--71.2 & $-$18.2 & +0.1 & +0.1 & $-$17.5 \\
 & GIN & 89.1 & 87.2 & 85.8--88.3 & $-$2.2 & $-$1.9 & +1.6 & $-$2.4 \\
 & Transformer & 90.3 & 89.8 & 89.4--90.1 & +0.4 & $-$0.4 & $-$0.3 & +0.4 \\
\bottomrule
\end{tabular}

\end{table}

\paragraph{Execution.} Before any E10 run, scoring was moved from the A2's machine to a laptop
CPU, after the preflight measured about a day of CPU time per graph there. Minesweeper's
training was moved to the laptop as well before anything had run on it. Seven Transformer
training runs on the A2 ran out of GPU memory beside other runs and were repeated.

\FloatBarrier
\section{E11: joint tuning}
\label{app:e11}

\paragraph{Design.} GCN, the six main graphs and the eight objectives. One random draw with a
fixed seed gave 16 configurations per objective from a joint space: width 128, 256 or 512, the
range of the published configurations of GRACE, BGRL and DGI; depth 1, 2 or 3; batch, layer or
no normalization; dropout 0 or 0.2; an AdamW learning rate log-uniform in $[10^{-4},10^{-2}]$
and weight decay log-uniform in $[10^{-6},10^{-3}]$; 100, 300 or 1{,}000 updates; edge dropping
from $U(0,0.7)$ and feature masking from $U(0,0.5)$ for the augmentation-based objectives; and
each objective's own hyperparameters. The shared recipe is added as configuration c00. The
same 17 configurations serve every graph, and the draw was committed with the rule. Views stay
symmetric and the encoder stays a GCN.

\paragraph{Selection.} Stage 1 trains every configuration with one seed and scores it with our
probe on the standard validation set. Stage 2 adds two seeds for the three best. The
configuration with the best mean validation accuracy over the three seeds is retrained with
seeds 0--4, which the selection never saw, and these models are scored with our probe and
liblinear, selected on validation per model as in E6. Search runs also record test scores, but configuration selection reads only validation
scores. The final evaluation uses fresh training runs with seeds held out from that selection,
not a new validation or test split.

\paragraph{Rule and result.} $D_\text{tuned}$ is the mean over the eight tuned objectives of
their five-seed mean minus $\hat A^2X$. With four or more graphs favoring the tuned objectives
the comparison would be withdrawn as a general statement; with one to three it is stated for
the shared protocol and the exceptions are named. One graph, PubMed, favors them
(Table~\ref{tab:e11}). Choosing the propagation depth among one, two and three steps on
validation, instead of two (no rule), changes no verdict: PubMed then favors pretraining by
2.8 points, and no other graph does.

\begin{table}[!htbp]
  \centering\small
  \caption{E11. Test accuracy (\%), GCN, probe chosen on validation for each representation.
  Shared: the ten-seed mean of the main grid, on another device for PubMed, Amazon Photo and
  Coauthor CS. Tuned: each objective's selected configuration, seeds 0--4. $D_\text{tuned}$:
  tuned mean minus $\hat A^2X$, with a joint seed bootstrap (2{,}000 draws); bold meets the
  rule. Above: tuned objectives above $\hat A^2X$. Last column: tuned minus the shared
  recipe trained in stage 1 on the same device, one seed, our probe, averaged over objectives.}
  \label{tab:e11}
  \footnotesize
  \setlength{\tabcolsep}{4pt}
  \resizebox{\linewidth}{!}{%
\begin{tabular}{@{}lrrrrrlr@{}}
\toprule
 & & \multicolumn{2}{c}{Mean of eight objectives} & & & & Lift over \\
\cmidrule(lr){3-4}
Dataset & $\hat A^2X$ & Shared & Tuned & $D_\text{tuned}$ [95\% interval] & Above & Best tuned objective & shared, one seed \\
\midrule
Cora & 82.0 & 80.1 & 82.0 & $-$0.01 [$-$0.27, +0.25] & 4 of 8 & GraphCL (+1.7) & +2.0 \\
CiteSeer & 72.3 & 66.8 & 69.9 & $-$2.35 [$-$2.52, $-$2.22] & 0 of 8 & GraphMAE ($-$0.6) & +2.2 \\
PubMed & 79.7 & 79.5 & 81.0 & \textbf{+1.35 [+1.26, +1.44]} & 7 of 8 & VICReg (+3.0) & +0.6 \\
Amazon Computers & 89.5 & 88.5 & 88.9 & $-$0.55 [$-$0.64, $-$0.43] & 1 of 8 & GraphMAE (+0.0) & +0.5 \\
Amazon Photo & 92.8 & 92.2 & 92.5 & $-$0.34 [$-$0.41, $-$0.26] & 1 of 8 & VICReg (+0.2) & +0.5 \\
Coauthor CS & 93.1 & 92.4 & 93.1 & $-$0.01 [$-$0.07, +0.02] & 5 of 8 & GraphMAE (+0.5) & +0.6 \\
\bottomrule
\end{tabular}
}
\end{table}

\paragraph{What the search selected.} Reported without a rule (Table~\ref{tab:e11-configs}).
Batch normalization, used by every model of the shared recipe, is kept in 17 of the 48 selected
configurations. Five of the six GraphDINO selections are the same configuration, with layer
normalization. The search tends to lift the weakest objectives of the shared recipe most, and
on the three citation graphs the spread between the best and the worst objective narrows.

\begin{table}[!htbp]
  \centering\small
  \caption{Configurations selected on validation: width / depth / normalization (BN batch, LN
  layer, -- none) / updates.}
  \label{tab:e11-configs}
  \footnotesize
\begin{tabular}{@{}lcccccc@{}}
\toprule
Objective & Cora & CiteSeer & PubMed & Computers & Photo & CS \\
\midrule
DGI & 512/2/--/100 & 512/3/LN/300 & 512/2/--/100 & 512/2/--/100 & 256/2/LN/300 & 256/2/LN/1000 \\
GraphCL & 512/2/--/100 & 256/1/--/100 & 128/2/LN/1000 & 512/2/BN/300 & 128/2/LN/1000 & 128/2/LN/1000 \\
BGRL & 256/3/BN/300 & 256/1/BN/300 & 128/2/BN/1000 & 128/2/BN/1000 & 512/1/LN/100 & 256/1/BN/300 \\
AFGRL & 256/1/--/300 & 256/1/--/300 & 512/1/BN/1000 & 512/1/BN/1000 & 256/1/BN/1000 & 512/1/BN/300 \\
VICReg & 256/3/BN/100 & 256/2/--/100 & 256/3/BN/100 & 512/3/LN/1000 & 512/3/LN/1000 & 512/1/--/300 \\
Barlow Twins & 512/3/--/300 & 128/1/BN/100 & 512/3/BN/1000 & 512/3/BN/100 & 256/3/LN/100 & 128/1/--/100 \\
GraphDINO & 512/2/LN/300 & 512/2/LN/300 & 128/2/--/300 & 512/2/LN/300 & 512/2/LN/300 & 512/2/LN/300 \\
GraphMAE & 128/1/BN/300 & 128/3/--/300 & 512/2/LN/100 & 128/2/LN/1000 & 128/3/LN/1000 & 128/1/BN/300 \\
\bottomrule
\end{tabular}

\end{table}

\paragraph{Execution.} The rule placed PubMed, Coauthor CS and Amazon Computers on the A2 GPU.
With five concurrent runs the A2 gave each a third to a half of the speed of a laptop CPU run,
so Coauthor CS and Amazon Computers were moved to the laptop CPU before any of their scores
existed. The three stage-1 runs already made for Coauthor CS on the A2 were set aside unread.
Each graph ran on a single device: Cora, CiteSeer, Coauthor CS and Amazon Computers on the
laptop CPU, Amazon Photo on the laptop GPU and PubMed on the A2. A coding error ran the first
final steps with seeds 0--9 instead of 0--4; it was corrected before any E11 score was read,
and the analysis uses seeds 0--4 throughout. Seven PubMed steps that failed beside other runs,
six of them out of GPU memory, were repeated, as the rule allows.

\FloatBarrier
\section{E8: row-normalized features}
\label{app:e8}

\paragraph{The test.} The encoders of the study read the raw node features. DGI, GRACE and the
PyGCL examples divide each Planetoid node's features by their sum. E8 trains the GCN with the eight
objectives on Cora, CiteSeer and PubMed with row-normalized features (5 seeds, the shared protocol
otherwise) and compares them with $\hat A^2X$ computed from the same row-normalized features, under
the probe better on validation, with a rule committed before the runs: if the average objective
exceeds $\hat A^2X$ by at least one point with an interval above zero on any of the three
datasets, the comparison of Section~\ref{sec:graph} is stated for raw features only.

\paragraph{Result.} The estimated differences stayed below the prespecified threshold under both preprocessing choices. With
row-normalized features the average GCN objective is 1.0 (95\% interval $-1.3$ to $-0.7$), 5.0
($-5.4$ to $-4.6$) and 0.3 ($-0.5$ to $0.0$) points below $\hat A^2X$ of the same features on Cora,
CiteSeer and PubMed, with 3, 0 and 3 of the eight objectives above it; on the same five seeds with
raw features it is 1.9, 5.6 and 0.2 points below. Row normalization lowers $\hat A^2X$ on Cora
(82.0 to 80.4\%) and CiteSeer (72.3 to 71.6\%) and leaves it level on PubMed (79.7 to 79.9\%);
these values are close to SGC's published 81.0, 71.9 and 78.9\%, obtained with the same
preprocessing and a trained classifier \citep{wu2019sgc}.

\FloatBarrier
\section{E7: normalization statistics at evaluation}
\label{app:e7}

\paragraph{Why it was added.} A review of the draft found that the untrained encoder is evaluated
with the initial running statistics of its batch normalization layers (mean 0, variance 1 and initial affine parameters, so the
layers are approximately identity maps in evaluation mode), while every trained model holds statistics
accumulated over its training passes. The reference then differs from a trained model in its
weights and in its normalization statistics. E7 re-scores every model of E1 on the six datasets,
the eight objectives and the untrained encoder with the three backbones, with \emph{graph
statistics}: the running statistics of every batch normalization layer are reset and recomputed
from one forward pass over the full graph, with only those layers in training mode, no gradient and
no label, before the usual extraction in evaluation mode. Both kinds of statistics are scored by the
same script on the same machine and thread count; the scores with the statistics the models hold
repeat those of E2 to the digit, for all 1{,}620 checkpoints and every probe. ogbn-arxiv and arxiv-year are
scored in the same way on the laptop GPU (the A2 was unreachable), with the 100-step probe, ours
and $k$NN; they are descriptive. On these checkpoints the laptop GPU repeats the A2's scores of E2
within 0.6 points under our probe (0.03 on average) and 0.04 under $k$NN; the 100-step probe differs
by 0.2 points on average, up to 4.4 on one checkpoint, since its random initializations are drawn
on the device.

\paragraph{The rules.} Three rules were committed before the runs (Table~\ref{tab:prereg}). R1
(the small GCN gains): if on Amazon Computers, Amazon Photo or Coauthor CS the mean gain of the GCN
objectives under our probe, with graph statistics for every model, exceeds the gain with the
statistics they hold by at least one point with an interval above zero, those gains are reported
as dependent on the statistics. R2 (the comparison with $\hat A^2X$): if $D$, with the objectives
under graph statistics, is at least one point with an interval above zero on any of the six
datasets, the comparison is stated with the models' own statistics only. R3 compares the additional gain reported by the
100-step probe relative to our probe. Let $I_{\rm stored}$ and $I_{\rm graph}$ be this
difference averaged over the nine many-label combinations before and after recomputing statistics.
If $I_{\rm graph}<I_{\rm stored}/2$, the rule attributes most of the excess to the untrained
encoder's initial statistics. For the nine few-label combinations the rule requires that the
100-step probe still give a smaller gain in at least six after recomputation.

\paragraph{Results.} For R1, none of the three increases reached one point. With graph
statistics the mean gain changes by $-0.2$ [$-0.5$, $+0.1$] points on Amazon Computers, $+0.3$
[$+0.1$, $+0.5$] on Amazon Photo and $0.0$ [$-0.2$, $+0.2$] on Coauthor CS. For R2, all six mean advantages remain negative. $D$ changes by at most 0.1 points on any of the six datasets
(Table~\ref{tab:checks}). R3: the mean inflation of the 100-step probe over the nine cells with
many labels falls from 6.4 to 0.6 points, below half, so it is attributed to the untrained encoder's
initial statistics; recomputing the statistics of the untrained encoder alone gives 0.4. With 20
labels per class the 100-step probe still shrinks the gain in 5 of the 9 cells (6 of 9 with the
statistics the models hold), below the rule's threshold of six, so the overfitting bias is
reported as dependent on the statistics. Outside the rules: the rankings of the objectives under
the two kinds of statistics agree with a median Kendall's $\tau$ of 1.00 under our probe and 0.89
under $k$NN; with graph statistics the embeddings of the untrained GCN have the scale of the
trained ones, within 7\%.

\begin{figure}[!htbp]
  \centering
  \includegraphics[width=\linewidth]{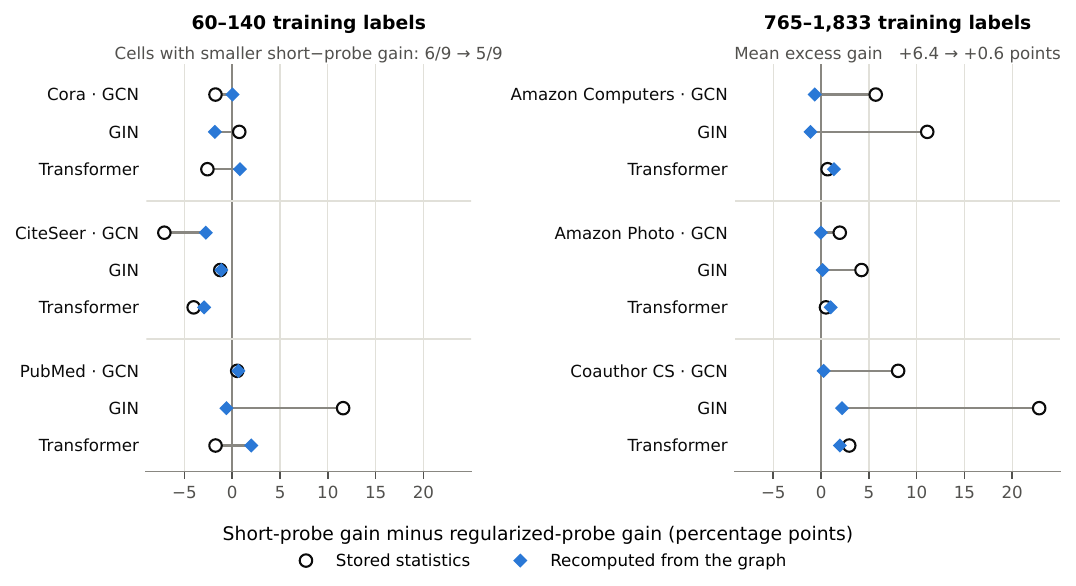}
  \caption{Short-probe gain minus regularized-probe gain. Each row pairs the same checkpoints
  before and after recomputing normalization statistics. Positive values mean the short
  probe gives the larger training gain. Both panels use the same scale.}
  \label{fig:calibration-grid}
\end{figure}

\begin{table}[!htbp]
  \centering
  \caption{E7: the untrained encoder (test accuracy, \%) and the mean gain of the eight objectives
  over it, with the normalization statistics each model holds (run.) and with statistics
  recomputed from one pass over the graph for every model (graph); our probe, the 100-step probe
  and $k$NN; standard split, 10 seeds.}
  \label{tab:e7}
  \resizebox{\linewidth}{!}{%
\begin{tabular}{@{}llrrrrrrrrrr@{}}
\toprule
 & & \multicolumn{4}{c}{Untrained encoder} & \multicolumn{6}{c}{Mean gain of the eight objectives} \\
\cmidrule(lr){3-6}\cmidrule(l){7-12}
 & & \multicolumn{2}{c}{Ours} & \multicolumn{2}{c}{100 steps} & \multicolumn{2}{c}{Ours} & \multicolumn{2}{c}{100 steps} & \multicolumn{2}{c}{$k$NN} \\
\cmidrule(lr){3-4}\cmidrule(lr){5-6}\cmidrule(lr){7-8}\cmidrule(lr){9-10}\cmidrule(l){11-12}
Dataset & Backbone & run. & graph & run. & graph & run. & graph & run. & graph & run. & graph \\
\midrule
Cora & GCN & 73.1 & 72.6 & 72.3 & 69.8 & +6.9 & +7.4 & +5.1 & +7.5 & +13.1 & +7.8 \\
 & GIN & 41.7 & 50.3 & 40.3 & 51.6 & +30.1 & +21.6 & +30.9 & +19.8 & +39.5 & +19.0 \\
 & Transformer & 49.7 & 55.4 & 47.7 & 49.9 & +20.6 & +14.7 & +18.0 & +15.5 & +32.4 & +15.8 \\
\addlinespace
CiteSeer & GCN & 57.7 & 57.7 & 56.8 & 52.5 & +8.6 & +8.5 & +1.5 & +5.8 & +11.4 & +10.1 \\
 & GIN & 36.1 & 37.8 & 35.6 & 37.3 & +20.0 & +18.1 & +18.7 & +16.9 & +24.4 & +18.4 \\
 & Transformer & 42.7 & 42.9 & 39.5 & 38.7 & +16.0 & +15.8 & +12.0 & +12.9 & +19.3 & +14.7 \\
\addlinespace
PubMed & GCN & 74.1 & 74.0 & 72.1 & 71.8 & +5.6 & +5.7 & +6.2 & +6.3 & +4.7 & +4.8 \\
 & GIN & 56.6 & 58.5 & 45.0 & 59.2 & +17.1 & +15.5 & +28.7 & +14.9 & +39.2 & +16.5 \\
 & Transformer & 66.5 & 63.6 & 66.0 & 59.3 & +9.9 & +12.7 & +8.1 & +14.7 & +13.1 & +11.0 \\
\addlinespace
Amazon Computers & GCN & 87.1 & 87.2 & 80.6 & 87.2 & +1.4 & +1.2 & +7.1 & +0.6 & $-$0.5 & +1.5 \\
 & GIN & 80.2 & 78.1 & 65.3 & 76.5 & +1.4 & +3.9 & +12.5 & +2.8 & +0.1 & $-$0.1 \\
 & Transformer & 76.0 & 79.2 & 74.5 & 77.0 & +8.7 & +5.4 & +9.4 & +6.8 & +26.8 & +14.5 \\
\addlinespace
Amazon Photo & GCN & 91.7 & 91.5 & 89.6 & 91.3 & +0.2 & +0.5 & +2.2 & +0.5 & $-$0.3 & +1.4 \\
 & GIN & 85.7 & 84.9 & 80.4 & 83.8 & +2.1 & +3.3 & +6.4 & +3.5 & +1.2 & +1.4 \\
 & Transformer & 83.0 & 85.8 & 82.3 & 84.5 & +8.3 & +5.5 & +8.8 & +6.5 & +34.1 & +15.2 \\
\addlinespace
Coauthor CS & GCN & 90.7 & 90.8 & 82.3 & 90.1 & +1.5 & +1.5 & +9.5 & +1.8 & +2.2 & +1.8 \\
 & GIN & 80.2 & 79.5 & 56.7 & 76.8 & +10.7 & +11.5 & +33.5 & +13.7 & +22.4 & +12.1 \\
 & Transformer & 86.5 & 86.8 & 83.0 & 84.2 & +5.7 & +5.5 & +8.6 & +7.5 & +25.5 & +7.2 \\
\bottomrule
\end{tabular}
}
\end{table}

\FloatBarrier
\section{Optimizer, stopping step and training budget}
\label{app:e3e5}

\paragraph{Which selected checkpoints can be rescored?}
E3 retained evaluation records. E5 retained the endpoints of the short and long runs;
its selection rule can choose an intermediate step of the long run. The separate short
run cannot supply those weights. For every GCN objective whose selected step coincides
with the saved long-run endpoint, we extracted the embeddings again on CPU and fitted both
probes (Table~\ref{tab:e5-endpoints}). The coverage differs by objective, so the original
eight-objective comparisons in Table~\ref{tab:checks} retain their original probe labels.

\begin{table}[!htbp]
  \centering\small
  \caption{Test accuracy (\%) on the available validation-selected E5 GCN checkpoints.
  Default and liblinear score the same frozen embeddings. Selected chooses the probe by
  validation accuracy for each checkpoint before averaging.}
  \label{tab:e5-endpoints}
\begin{tabular}{@{}llrrrr@{}}
\toprule
Dataset & Objective & Seeds & Default & Liblinear & Selected \\
\midrule
Amazon Computers & GraphCL & 5 & 89.7 & 89.6 & 89.7 \\
Amazon Computers & BGRL & 5 & 89.4 & 89.4 & 89.4 \\
Amazon Computers & AFGRL & 5 & 88.5 & 88.7 & 88.7 \\
Amazon Computers & VICReg & 5 & 89.2 & 89.5 & 89.5 \\
Amazon Photo & GraphCL & 5 & 92.5 & 93.0 & 93.0 \\
Amazon Photo & VICReg & 5 & 92.7 & 93.0 & 92.9 \\
Coauthor CS & DGI & 5 & 91.6 & 92.2 & 92.2 \\
\bottomrule
\end{tabular}

\end{table}

\paragraph{E3.} Table~\ref{tab:e3} gives every lift. With GIN and the Transformer on CiteSeer the
spread between the best and the worst objective shrinks by 4.5 and 6.0 points, because the weakest
methods gain the most. Some cells diverge and are never chosen: DGI and VICReg with GIN under SGD
at $5\cdot10^{-2}$ and $10^{-1}$ in every seed; with a GCN, DGI at SGD $10^{-1}$ in two to five
seeds of five, and on the Amazon and Coauthor graphs also DGI at $5\cdot10^{-2}$ (one to five
seeds) and VICReg at $10^{-1}$ (three or four). On the Amazon and Coauthor graphs three choices lay at the edge of the
grid: DGI at AdamW $10^{-4}$ on both Amazon graphs and GraphCL at SGD $10^{-1}$ on Coauthor CS.
With one more cell each, as the rule requires, DGI keeps $10^{-4}$, and GraphCL moves to SGD
$5\cdot10^{-1}$ and gains 0.3 points; the verdict does not change.

\paragraph{Stopping step (E4).}
In the shared optimizer setting, 15 of 48 trajectories decline after the best-validation
checkpoint and 33 rise and hold under the prespecified definitions. Selecting that checkpoint
instead of the final step gives ranking agreement of $\tau=0.79$ to 1.00 for GCN and GIN, and
0.50 and 0.64 for the Transformer on Cora and CiteSeer. At an AdamW learning rate of $10^{-2}$,
all 16 GIN trajectories reach effective rank below 10 or mean pair cosine above 0.95 at some
step. BGRL with a GIN on Cora, for example, reaches 58.1\% after two steps and ends at 33.8\%.

\paragraph{E5.} On Amazon Computers longer training helps a little: seven of the eight objectives
reach their best validation accuracy at 1{,}000 steps or later, and GraphDINO gains the most
(0.72 points). On Coauthor CS six stop at 300 steps, and at the 3{,}000-step endpoint GraphCL, AFGRL
and GraphDINO are below the untrained encoder. GraphCL and AFGRL on Coauthor CS ran three seeds
instead of five, under a cost limit fixed with the design of E5 (more than 45 minutes per
3{,}000-step seed). On Cora the mean change in gain with the longer budget is $-0.27$ points
for the GCN, $-0.13$ for the Transformer and $+1.43$ for the GIN. BGRL accounts for most of
the GIN improvement ($+8.9$ points). We use separate short and long runs because teacher
momentum in BGRL, AFGRL and GraphDINO depends on the scheduled length of the run.

\begin{table}[!htbp]
  \centering
  \footnotesize
  \setlength{\tabcolsep}{3.4pt}
  \caption{E3: each objective at the shared protocol and at its own best optimizer and learning
  rate, chosen on validation among ten cells (eleven for DGI on both Amazon graphs and GraphCL on
  Coauthor CS, where the rule fixed in advance added a cell beyond the edge of the grid); 5 seeds,
  regularized probe. Rows: Kendall's $\tau$ between the shared and the tuned ranking, with a 95\%
  seed-bootstrap interval that repeats the choice of the cell in every replicate; the spread
  between the best and the worst objective; the mean lift over the objectives, tuned minus
  shared, with its interval (the quantity of the decision rule); the lift of each objective.
  Transf.: Transformer.}
  \label{tab:e3}
\begin{tabular}{lrrrrrrrrr}
\toprule
 & \multicolumn{3}{c}{Cora} & \multicolumn{3}{c}{CiteSeer} & \multicolumn{1}{c}{\begin{tabular}[b]{@{}c@{}}Amazon\\Computers\end{tabular}} & \multicolumn{1}{c}{\begin{tabular}[b]{@{}c@{}}Amazon\\Photo\end{tabular}} & \multicolumn{1}{c}{\begin{tabular}[b]{@{}c@{}}Coauthor\\CS\end{tabular}} \\
\cmidrule(lr){2-4} \cmidrule(lr){5-7} \cmidrule(lr){8-8} \cmidrule(lr){9-9} \cmidrule(lr){10-10}
 & GCN & GIN & Transf. & GCN & GIN & Transf. & GCN & GCN & GCN \\
\midrule
$\tau$, shared vs tuned & 0.71 & 0.86 & 0.50 & 0.86 & 0.43 & 0.57 & 1.00 & 0.50 & 0.79 \\
 & {\tiny [0.57, 0.93]} & {\tiny [0.79, 0.93]} & {\tiny [0.29, 0.71]} & {\tiny [0.64, 1.00]} & {\tiny [0.36, 0.86]} & {\tiny [0.29, 0.86]} & {\tiny [0.71, 1.00]} & {\tiny [0.14, 0.86]} & {\tiny [0.64, 0.93]} \\
Spread, shared & 6.0 & 18.2 & 20.3 & 9.2 & 21.7 & 26.8 & 1.7 & 0.8 & 1.6 \\
Spread, tuned & 5.8 & 19.7 & 19.1 & 9.2 & 17.2 & 20.8 & 1.8 & 1.1 & 1.6 \\
Mean lift (points) & +1.0 & +1.5 & +3.3 & +3.3 & +3.6 & +7.6 & +0.1 & +0.1 & +0.2 \\
 & {\tiny [+0.4, +1.1]} & {\tiny [+1.0, +2.2]} & {\tiny [+2.6, +3.7]} & {\tiny [+2.8, +3.7]} & {\tiny [+2.4, +4.5]} & {\tiny [+6.6, +8.0]} & {\tiny [$-$0.0, +0.1]} & {\tiny [$-$0.0, +0.2]} & {\tiny [+0.1, +0.2]} \\
\midrule
\multicolumn{10}{l}{\textit{Lift of each objective (points)}} \\
DGI & +2.1 & +2.3 & +2.2 & +3.4 & +3.4 & +6.6 & $-$0.0 & $-$0.2 & +0.0 \\
GraphCL & +0.4 & +1.8 & +1.1 & +3.3 & +2.6 & +3.7 & $-$0.0 & +0.4 & +0.3 \\
BGRL & +0.0 & $-$1.8 & +2.8 & +3.0 & +4.9 & +13.3 & +0.1 & +0.2 & +0.2 \\
AFGRL & +0.0 & +0.7 & +2.7 & +5.4 & +6.0 & +6.3 & +0.1 & $-$0.3 & +0.1 \\
VICReg & +1.4 & +2.6 & +3.5 & +3.3 & +3.2 & +10.1 & +0.0 & +0.1 & +0.3 \\
Barlow Twins & +1.7 & +0.0 & +1.5 & +3.2 & $-$0.3 & +3.1 & +0.2 & +0.4 & $-$0.1 \\
GraphDINO & +0.8 & +2.5 & +3.9 & +3.0 & +8.0 & +9.7 & +0.2 & +0.1 & +0.5 \\
GraphMAE & +1.5 & +4.1 & +8.9 & +1.7 & +0.8 & +8.4 & +0.2 & +0.0 & +0.0 \\
\bottomrule
\end{tabular}

\end{table}

\FloatBarrier
\section{BGRL with the published pretraining recipe}
\label{app:arxiv}

To check the implementation independently of our shared protocol, we trained BGRL on ogbn-arxiv
following the pretraining settings of \citet{thakoor2022bgrl}, with a three-layer GCN with layer normalization and weight
standardization, 10{,}000 full-graph steps, edge dropping 0.6, and a learning rate of $10^{-2}$
with warm-up and cosine decay. For evaluation, we fit our standardized L-BFGS probe. The
original study used L2-normalized embeddings and a 100-step AdamW probe, and reported scores over 20
model initializations. Over three seeds our implementation reaches 72.31 $\pm$ 0.14\% validation and
71.32 $\pm$ 0.10\% test accuracy, against 72.53 $\pm$ 0.09 and 71.64 $\pm$ 0.12 in the paper; the
untrained encoder gives 70.23 $\pm$ 0.18 and 69.53 $\pm$ 0.03, against 69.90 and 68.94. Pretraining therefore improves test accuracy by 1.79 points in our run
and by 2.70 points in the published comparison.

\FloatBarrier
\section{Retrospective analyses of the interpretation of training gains}
\label{app:attribution}

These analyses were added after the primary results were observed. Their rules and executed
code are documented separately from the earlier decision records. They reuse saved search
records or frozen checkpoints; no encoder was retrained. Table~\ref{tab:prereg} and all
original thresholds remain unchanged.

\subsection{Selecting an objective without consulting final test scores}

For each dataset and objective, we reconstruct the original 17-configuration search from its
per-seed records: seed 100 ranks the configurations; the best three also use seeds 101 and
102; their mean validation accuracy chooses a configuration. We then select the objective
with the highest such mean validation accuracy, breaking an exact tie lexicographically by
objective name. Only after freezing these choices do we read the final-seed evaluations.
The final evaluation uses seeds 0--4, retaining the original validation-based choice between
two probes. The search uses the default probe. Its recorded test scores are never inputs to
either selection. Final seeds are independent of the search seeds, but use the same graph
and labeled-node split.

\begin{table}[htb]
  \centering\footnotesize
  \caption{Retrospective objective selection from E11. Search validation averages seeds
  100--102; final test averages seeds 0--4. $D$ compares the selected pipeline with two-step
  propagation. Oracle $D$ is the largest observed final test mean among objectives and is
  not used for selection. Configuration IDs refer to the archived E11 search space.}
  \label{tab:selected-pipeline}
  \begin{tabular}{@{}lllrrrr@{}}
\toprule
Dataset & Selected objective & Config & Search val. & Final test & $D$ [95\% interval] & Oracle $D$ \\
\midrule
Cora & GraphCL & c09 & 80.87 & 83.72 & $+1.72$ [$+0.98, +2.46$] & $+1.72$ \\
CiteSeer & GraphMAE & c10 & 72.40 & 71.68 & $-0.62$ [$-1.26, +0.18$] & $-0.62$ \\
PubMed & GraphCL & c04 & 84.80 & 81.38 & $+1.68$ [$+1.10, +2.22$] & $+3.00$ \\
Computers & VICReg & c07 & 90.81 & 89.22 & $-0.27$ [$-0.44, -0.18$] & $+0.02$ \\
Photo & GraphMAE & c01 & 92.90 & 92.49 & $-0.31$ [$-0.46, -0.14$] & $+0.15$ \\
Coauthor CS & GraphMAE & c11 & 93.27 & 93.53 & $+0.46$ [$+0.38, +0.53$] & $+0.46$ \\
\bottomrule
\end{tabular}

\end{table}

Intervals use 10{,}000 bootstrap draws of the five final seed indices, with random generator
seed 20261129. They condition on the chosen objective and configuration, the fixed-feature
fit, the graph and the split. They do not repeat configuration or objective selection and
do not measure variability across labeled-node sets. The objective average, selected
pipeline and oracle maximum are different summaries of the same search. In particular,
PubMed's selected advantage is 1.68 points while its oracle advantage is 3.00 points.

\subsection{Replaying the selected configuration's initialization}

The tuned checkpoints contain their configuration and final state, but no initial state or
initial-state hash. A fresh encoder with the same numerical seed would not by itself establish
matching. We audit the runner: it calls \texttt{torch.manual\_seed} before constructing a
\texttt{DataModule}, which consumes no random numbers, and then constructs the full model on
CPU. The full constructor preserves the order of encoder and auxiliary-head initialization;
we do not substitute an isolated encoder constructor. GraphDINO's teacher starts from the
student encoder state, while BGRL evaluation uses the online encoder.

On Cora, CiteSeer, Computers, Photo and Coauthor CS, the construction code and PyTorch/PyG
versions match the archived run environment. For every selected objective and each of seeds
0--4, replaying its shared-recipe constructor reproduces every tensor of the corresponding
archived untrained encoder exactly. We then apply the same reconstruction to its chosen
tuned configuration and score the reconstructed initialization and saved trained checkpoint
together on CPU, with three threads and the two-probe validation rule. We also rescore the
fixed reference. This is an audited deterministic replay, not independent evidence from
an archived tuned initial state. PubMed used different PyTorch and PyG versions; we leave
its tuned gain unreported rather than assume seed equivalence across those environments.

\begin{table}[htb]
  \centering\small
  \caption{Both comparisons for the validation-selected tuned configurations, with their
  starting points reconstructed under the audited replay. Each row averages five paired
  seeds. Both encoder endpoints and the fixed reference are rescored together on CPU.
  The gain interval resamples paired trained/untrained seeds.}
  \label{tab:tuned-replay}
  \begin{tabular}{@{}lrrrrr@{}}
\toprule
Dataset & Replayed $U$ & Pretrained $T$ & Fixed $P$ & $G$ [95\% interval] & $D$ \\
\midrule
Computers & 88.41 & 89.22 & 89.49 & $+0.81$ [$+0.63, +1.00$] & $-0.27$ \\
Photo & 91.01 & 92.49 & 92.79 & $+1.47$ [$+1.35, +1.60$] & $-0.31$ \\
CiteSeer & 50.14 & 71.68 & 72.30 & $+21.54$ [$+18.72, +24.18$] & $-0.62$ \\
Coauthor CS & 88.76 & 93.53 & 93.07 & $+4.78$ [$+4.50, +5.01$] & $+0.46$ \\
Cora & 76.72 & 83.72 & 82.00 & $+7.00$ [$+6.00, +8.08$] & $+1.72$ \\
\bottomrule
\end{tabular}

\end{table}

The gain is positive on all five graphs, even when the selected model remains below the
fixed reference. These comparisons do not attribute the improvement over the shared recipe
to any one architectural or optimization change. They measure improvement from the selected
configuration's own reconstructed starting point.

\subsection{A positive-scaling control}

For Coauthor CS, we extract the saved GCN embeddings for seeds 0--2, all eight objectives
and the shared untrained encoder, on CPU with three threads. Let $n(R)$ be a representation's
mean node norm. For each encoder seed, the norm-matching target is the mean of $n(T)$ over
its eight pretrained objectives. Each representation is multiplied by its target-to-norm
ratio, computed without labels. We also apply fixed multipliers 0.1, 1 and 10. None is
selected using validation or test accuracy. Positive global scaling changes neither the
family of attainable affine linear boundaries nor normalized angles; it can change a
finite-budget fit with a fixed initialization and optimizer.

Each short-probe comparison uses the same five classifier seeds, Xavier-initialized weights,
zero bias, 100 full-batch Adam updates, learning rate 0.01 and no weight decay. We repeat
these fits after training-node coordinate standardization. At the original and norm-matched
scales, we also fit the default standardized, validation-regularized probe. Probe repeats
are averaged within an encoder seed before objectives and encoder seeds are averaged.

\begin{table}[htb]
  \centering\small
  \caption{Global-scaling control on Coauthor CS GCNs. Scores and gains are in percentage
  points; $U$ and $T$ are accuracies. ``Standardized'' uses the same short probe after
  coordinate standardization. The final column uses the default regularized probe.
  Dashes mark fits not run. These are descriptive three-seed means.}
  \label{tab:scale-control}
  \begin{tabular}{@{}lrrrrr@{}}
\toprule
Scaling & Short $U$ & Short $T$ & Short $G$ & Standardized $G$ & Regularized $G$ \\
\midrule
Original & 82.25 & 91.85 & +9.60 & +2.26 & 1.36 \\
Norm matched & 90.60 & 91.84 & +1.24 & +2.26 & 1.35 \\
$\alpha=0.1$ & 25.95 & 92.02 & +66.07 & +2.26 & -- \\
$\alpha=10$ & 91.17 & 89.48 & -1.69 & +2.26 & -- \\
\bottomrule
\end{tabular}

\end{table}

The fixed multipliers also illustrate why selecting a favorable scale on test accuracy would
be inappropriate: the short probe can make the same pretrained features appear to gain
66.1 points or lose 1.7 points. Norm matching supplies a label-free, interpretable comparison;
it reduces the original 9.6-point gain to 1.2 points. The standardized short-probe gain remains
2.26 points across all four conditions, and the regularized gain changes by less than 0.01
points. This control isolates global scale in one setting, not the entire effect of BatchNorm.
The raw records include every multiplier, node norm, coordinate standard deviation,
classifier training loss, validation/test score, checkpoint hash and executed-source hash.

\subsection{Auditing the normalization intervention}

Recalibration works on a deep copy, leaving the caller's parameters and buffers unchanged.
It resets each tracked BatchNorm layer's running means, variances and batch counter, sets
its momentum to \texttt{None}, and enables training mode only for BatchNorm. This requests
cumulative averaging: with one full-graph batch, the stored mean and unbiased variance
replace their reset values, rather than blend with them using the usual momentum.
All other layers, including dropout, remain in evaluation mode, and gradients are disabled.
After the pass, momentum settings are restored and the entire copy enters evaluation mode.
No trainable parameter is updated. A regression test checks these properties, including
exact parameter equality, the observed batch statistics and a batch counter of one.

The graphs in these controls are processed in full; there is no neighbor sampling during
recalibration. With several normalization layers, later layers observe the activations
produced during the calibration pass. PyTorch uses a biased variance for normalization
within that pass but stores the unbiased estimate for later evaluation. We therefore do not
claim exact equality between calibration-pass and evaluation-pass activations. The operation
uses unlabeled graph data and changes the encoder function despite leaving its weights fixed.

\subsection{A common fixed-feature family}

We select among $X$, $\hat AX$, $\hat A^2X$ and their side-by-side concatenation. Within each
representation, the probe and penalty are selected on validation, then the representation
is selected on validation; exact ties retain the listed order. The six main graphs and
WikiCS are rescored together on CPU with three threads and both probes. E10 already contains
all four families under both probes, so its records are reused. The year-prediction task
retains its default probe; this extends the feature family there without introducing the
second probe. We do not extend the subject-prediction task in this sensitivity analysis.

\begin{table}[htb]
  \centering\footnotesize
  \caption{Retrospective fixed-feature family. Every numeric column uses that representation's
  validation-selected probe, except arxiv-year, which uses the default probe throughout.
  The selected family is chosen by validation, not by the largest number in this test table.
  Minesweeper uses ROC-AUC; the others use accuracy (\%).}
  \label{tab:fixed-family}
  \begin{tabular}{@{}llrrrrr@{}}
\toprule
Dataset & Selected family & $X$ & $\hat AX$ & $\hat A^2X$ & Concatenation & Selected test \\
\midrule
Cora & $\hat A^2X$ & 60.20 & 78.40 & 82.00 & 75.80 & 82.00 \\
CiteSeer & $\hat A^2X$ & 62.40 & 71.60 & 72.30 & 72.30 & 72.30 \\
PubMed & $\hat A^2X$ & 72.90 & 76.10 & 79.70 & 77.60 & 79.70 \\
Computers & $\hat A^2X$ & 79.89 & 89.37 & 89.49 & 88.47 & 89.49 \\
Photo & $[X,\hat AX,\hat A^2X]$ & 86.76 & 92.92 & 92.79 & 94.13 & 94.13 \\
Coauthor CS & $[X,\hat AX,\hat A^2X]$ & 92.13 & 93.41 & 93.07 & 94.42 & 94.42 \\
WikiCS & $\hat A^2X$ & 72.19 & 78.81 & 80.23 & 78.95 & 80.23 \\
arxiv-year & $[X,\hat AX,\hat A^2X]$ & 34.89 & 39.85 & 38.99 & 41.57 & 41.57 \\
Roman-empire & $[X,\hat AX,\hat A^2X]$ & 64.49 & 44.95 & 44.78 & 76.01 & 76.01 \\
Amazon-ratings & $\hat A^2X$ & 39.10 & 41.01 & 42.90 & 41.78 & 42.90 \\
Minesweeper & $[X,\hat AX,\hat A^2X]$ & 52.63 & 81.20 & 71.15 & 89.37 & 89.37 \\
\bottomrule
\end{tabular}

\end{table}

The family retains two-step propagation on the three citation graphs, Computers and WikiCS,
and retains E10's original choices. Concatenation improves the practical reference on Photo,
Coauthor CS and year prediction. The selected Cora and PubMed objectives remain ahead;
Coauthor CS changes from $+0.46$ against two-step propagation to $-0.89$ against the expanded
family. This is a comparison against an alternative pipeline with a different feature width,
not an architecture-matched control or an exhaustive search over graph filters.

\subsection{What the existing alternative splits support}

The archived random-split evaluations use the default standardized probe, not the primary
two-probe selection rule. Their 10/10/80 splits also change the label counts on the citation
graphs; arxiv-year uses five 50/25/25 splits. For each split we first average model seeds
within an objective and then the eight objectives. Table~\ref{tab:split-sensitivity} gives
the observed distribution over splits without treating model seeds as additional independent
label draws. These ranges are sensitivity summaries, not confidence intervals.

\begin{table}[htb]
  \centering\small
  \caption{Existing random-split sensitivity of the shared-recipe GCN mean, relative to
  two-step propagation, under the default probe. Neither the label resources nor the probe
  rule are held equal to the main validation-selected comparison on every dataset.}
  \label{tab:split-sensitivity}
  \begin{tabular}{@{}lrrrr@{}}
\toprule
Dataset & Splits & Mean $D$ & Range of $D$ & Positive splits \\
\midrule
Cora & 10 & +4.23 & [+2.95, +5.72] & 10 \\
CiteSeer & 10 & +7.12 & [+4.71, +9.50] & 10 \\
PubMed & 10 & -0.79 & [-1.60, -0.28] & 0 \\
Computers & 10 & -0.58 & [-1.05, -0.10] & 0 \\
Photo & 10 & -0.70 & [-1.14, -0.22] & 0 \\
Coauthor CS & 10 & +0.03 & [-0.13, +0.23] & 4 \\
WikiCS & 10 & -0.50 & [-1.20, +0.69] & 2 \\
arxiv-year & 5 & +3.20 & [+3.11, +3.41] & 5 \\
\bottomrule
\end{tabular}

\end{table}

The positive citation-graph differences under this probe therefore do not replicate the
primary selected-probe result. Near-zero differences also change sign across splits on
Coauthor CS and WikiCS. On E10's ten fixed splits, retaining the fixed-feature family chosen
on split 0 and using the default probe, the Minesweeper Transformer's advantage remains
positive in all ten splits, ranging from 0.28 to 0.67 ROC-AUC points. The Roman-empire
Transformer remains 3.96--4.81 points below its reference. The tuned objective-selection
analysis has no alternative-split evaluation; its positive results remain conditional on
the original labeled-node sets.

\paragraph{Reproduction.}
\texttt{analyze\_attribution.py} reconstructs search-only selection and split sensitivity;
\texttt{attribution\_controls.py} runs the scaling control;
\texttt{replay\_tuned\_initialization.py} audits and replays initial states;
\texttt{fixed\_family\_sensitivity.py} scores the expanded feature family; and
\texttt{summarize\_attribution.py} generates their tables. New records are separate from the
original results and include configuration, software and source/checkpoint provenance.
\FloatBarrier

\FloatBarrier
\section{Variation across the grid and embedding geometry}
\label{sec:synthesis}

The accuracy differences and rankings in Section~\ref{sec:robust} describe predictive
performance. Here we summarize dispersion across the experimental grid and examine a separate
question about embedding similarity. Neither analysis identifies a cause of rank changes.

\begin{figure}[t]
  \centering
  \includegraphics[width=\linewidth]{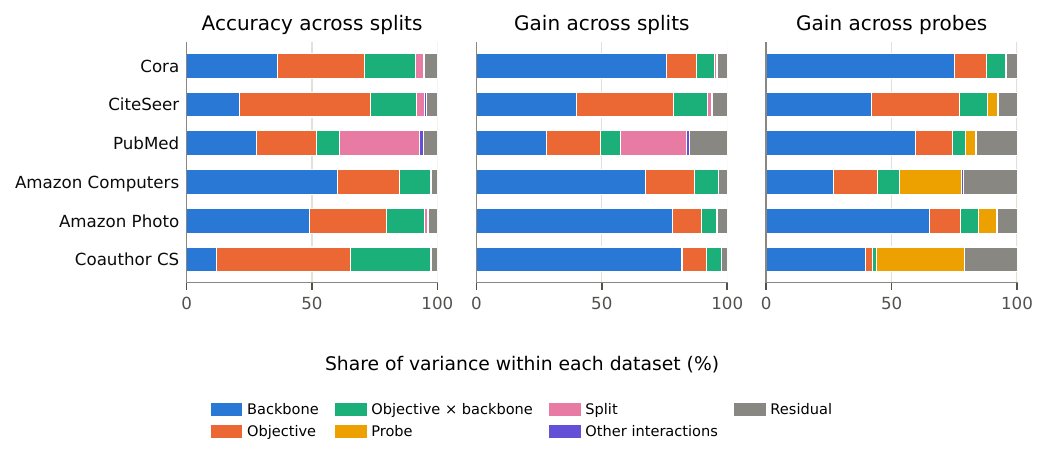}
  \caption{Shares of the total variation within each dataset. Accuracy and gain use the default
  probe across standard and random splits. The four-probe comparison uses the standard split.
  The residual includes seed variation and interactions outside those shown. The shares depend
  on the objectives, backbones and evaluation settings included in each grid.}
  \label{fig:shares}
\end{figure}

\paragraph{Accuracy and gain have different sources of variation.}
For the eight objectives and three backbones under the shared recipe, the objective accounts
for 24--54\% of the variance in test accuracy within a dataset. The backbone accounts for
12--60\%, and the objective--backbone interaction accounts for a further 9--32\%
(Figure~\ref{fig:shares}). This interaction measures how much an objective's relative
performance changes with the backbone. The split accounts for at most 3\% on five datasets.
On PubMed it accounts for 32\%, where the public split provides 60 training labels and a
random split provides about 2{,}000. These shares describe the sampled grid. They do not
estimate the fraction of any individual model's predictions due to pretraining.

Subtracting the untrained encoder's accuracy changes the picture. On Coauthor CS the three
untrained backbones span 10.6 points and their trained means span only 1.3 points. This difference
in starting points helps explain why the backbone accounts for 82\% of the variance in gain there.
On Amazon Computers, under the standard split, recomputing batch-normalization statistics
reduces the backbone's share of gain variance from 68\% to 35\%. Holding the split fixed and
varying the four linear probes instead, the probe accounts for up to 35\% of the variance in gain
and at most 12\% of the variance in accuracy.

\begin{figure}[t]
  \centering
  \includegraphics[width=\linewidth]{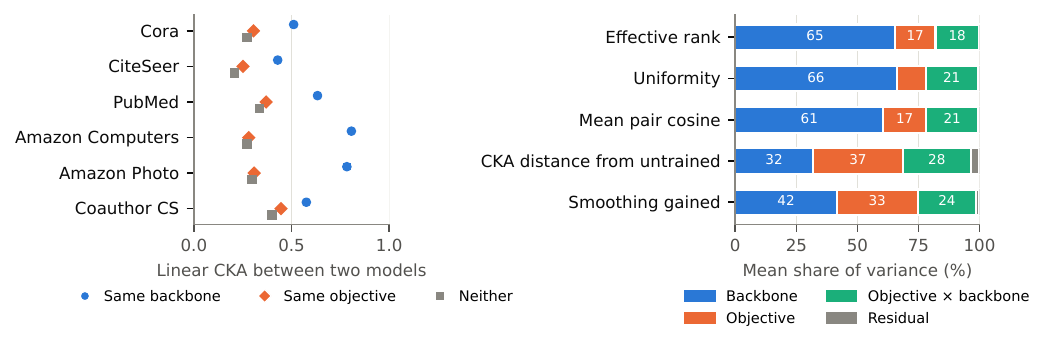}
  \caption{Similarity of frozen embeddings after 300 training steps. The left panel shows mean
  CKA and 95\% seed-bootstrap intervals for pairs trained with the same seed that share a backbone,
  an objective or neither. Higher
  linear CKA means more similar representations. The right panel shows variance shares for five
  properties of the embeddings, averaged over the six datasets.}
  \label{fig:cka}
\end{figure}

\paragraph{Embeddings group by backbone at 300 steps.}
Linear centered kernel alignment (CKA) compares how two encoders represent the same set of
nodes \citep{kornblith2019cka}. On each of the six datasets, mean CKA is higher for encoders
sharing a backbone (0.43--0.81 across datasets) than for encoders sharing an objective
(0.25--0.45) at the shared budget (Figure~\ref{fig:cka}). For example, GraphCL with a GCN on Cora
is closer to DGI with a GCN (0.73) than to GraphCL with a GIN (0.60). Sharing the initial weights
might explain this pattern. Across different seeds, models sharing a backbone still reach
0.38--0.80 mean CKA. The backbone also accounts for 52--76\% of the variance in effective rank,
a measure of how many embedding dimensions carry substantial variation \citep{roy2007erank}.

The training budget limits this result. On Cora we specified that the grouping would count as
persistent if pairs sharing a backbone exceeded pairs sharing an objective by at least 0.10 mean
CKA across seeds at 3{,}000 steps. The margin falls from 0.15 at 300 steps to 0.04 at 3{,}000
(0.27 versus 0.23), below the specified threshold. A claim about the geometry learned by these
architectures therefore needs a training budget as well as a backbone.

\paragraph{Can embedding geometry select an objective without labels?}
We tested effective rank, CKA distance from the untrained encoder, the change in smoothing across
edges, and negated uniformity \citep{garrido2023rankme,wang2020uniformity}. The prespecified test
covers the seven objectives available before GraphMAE was added. We used Cora and CiteSeer for
development and evaluated 15 held-out dataset--backbone groups. To pass, a metric had to correlate
positively with training gain (a seed-bootstrap interval above zero for mean Spearman's $\rho$)
and choose an objective with lower average regret than random choice. Regret is the accuracy
lost relative to the best objective in the same group (Table~\ref{tab:predict}).

Only CKA distance passed. Its mean $\rho$ was 0.26 [0.24, 0.35]. Choosing the objective whose
embeddings moved furthest from initialization lost 2.90 points on average, against 2.99 for a
random choice. Omitting either ogbn-arxiv or Amazon Photo reverses that small advantage. Effective
rank has a different use. Across objectives its test-set correlation with gain was negative
($\rho=-0.16$). Within an objective, all 13 optimizer comparisons with a collapsed endpoint
(effective rank below 10) gave that endpoint the lower accuracy. This makes low rank a useful
diagnostic in those runs, while the selection test offers little support for choosing an
objective from geometry alone.

\begin{table}[!htbp]
  \centering
  \caption{Label-free prediction of the gain over the untrained encoder, with the rules fixed in
  advance (Section~\ref{sec:synthesis}). Mean Spearman $\rho$ between the metric and the gain over
  the dataset--backbone groups. Regret: points lost by choosing the objective the metric ranks
  first instead of the best one; random: the expected loss of a uniformly random choice. $k$NN:
  a sensitivity analysis, not part of the rule.}
  \label{tab:predict}
  \resizebox{\linewidth}{!}{%
\begin{tabular}{llrrrrl}
\toprule
Gain by & Candidate & \multicolumn{2}{c}{Mean $\rho$ [95\% interval]} & Regret & Random & Verdict (test) \\
\cmidrule(lr){3-4}
 & & development & test & test & test & \\
\midrule
Regularized probe & Effective rank & $-$0.51 & $-$0.16 [$-$0.19, $-$0.10] & 3.06 & 2.99 & does not predict \\
 & CKA distance from untrained & +0.39 & +0.26 [+0.24, +0.35] & 2.90 & 2.99 & predicts \\
 & Smoothing gained & +0.57 & +0.27 [+0.22, +0.34] & 3.27 & 2.99 & does not predict \\
 & Uniformity (negated) & $-$0.09 & $-$0.10 [$-$0.15, $-$0.05] & 3.42 & 2.99 & does not predict \\
\midrule
$k$NN (sensitivity) & Effective rank & $-$0.40 & $-$0.19 [$-$0.22, $-$0.12] & 5.11 & 4.60 & does not predict \\
 & CKA distance from untrained & +0.27 & +0.27 [+0.22, +0.34] & 3.88 & 4.60 & predicts \\
 & Smoothing gained & +0.51 & +0.26 [+0.21, +0.31] & 3.58 & 4.60 & predicts \\
 & Uniformity (negated) & $-$0.04 & $-$0.11 [$-$0.15, $-$0.05] & 6.18 & 4.60 & does not predict \\
\bottomrule
\end{tabular}
}
\end{table}

\FloatBarrier
\section{What the untrained baseline exposed during development}
\label{sec:details}

\begin{table}[!htbp]
  \centering
  \small
  \caption{Development diagnostics. Entries give test accuracy (\%) for two implementations
  and an untrained reference. The one-seed DGI stress runs span code revisions. The BGRL
  comparison changes the predictor's activation and normalization together.}
  \label{tab:details}
  \footnotesize
  \setlength{\tabcolsep}{3.4pt}
  \begin{tabular}{@{}>{\raggedright\arraybackslash}p{3.6cm}llrr@{}}
    \toprule
    Setting & Change & Result & Untrained & Seeds \\
    \midrule
    GCN BatchNorm momentum & $0.01 \to 0.1$ & supervised head, PubMed: $49.7 \to 73.8$ & -- & 10 \\
     & & DGI $k$NN, Cora: $64.9 \to 73.5$ & 60.3 & 10 \\
    DGI, real and corrupted graph & two passes $\to$ one & $k$NN, ogbn-arxiv: $33.7 \to 52.9$ & 53.3 & 1 \\
    BGRL predictor, layer-norm encoder & BN/ReLU $\to$ PReLU & ogbn-arxiv: $67.9 \to 71.3$ & 69.5 & 3 \\
    \bottomrule
  \end{tabular}
\end{table}

Two defects of our own implementation produced plausible numbers without any error, and the
comparison with the untrained encoder is what exposed them (Table~\ref{tab:details}); they are
the reason for the first recommendation of Section~\ref{sec:recommendations}. With a BatchNorm momentum of 0.01, the running statistics used at evaluation
retain a fraction $0.99^{300}\simeq 0.05$ of their initial values after 300 updates. When the
activation variance is small relative to the initial running variance of one, evaluation-mode
embeddings can consequently have a different scale from those seen during training \citep[cf.][]{wu2021batchnorm}.
Linear-probe accuracy, refitted on whatever embeddings it is given, barely moved; $k$NN and the
supervised head did. When DGI encodes real and corrupted graphs separately, each pass uses its own batch
statistics. On mini-batches of ogbn-arxiv the mean discriminator loss fell below 0.001 at epoch two and the embeddings ended below those of the untrained encoder. Encoding both graphs in one
pass makes the two groups share statistics. MoCo also controls the batch information available to its encoders \citep{he2020moco}. We
found both defects only by comparing every result with the untrained encoder and by running on a
larger graph.

In the BGRL comparison, both the normalization and the activation in the predictor change.
Our default predictor uses BatchNorm followed by ReLU. The reference predictor uses PReLU
without normalization. With the layer-normalized encoder on ogbn-arxiv, the default predictor
finishes below the untrained encoder in each seed (1.6 points below on average). The reference
predictor finishes 1.8 points above it. This comparison measures the effect of changing the
predictor as a whole. Related work on normalization in bootstrapped representations also reports
sensitivity to architectural choices \citep{richemond2020byolbn}.

\FloatBarrier
\section{Same seeds, another machine}
\label{sec:seeds}

\begin{table}[!htbp]
  \centering
  \small
  \caption{The seven objectives of the first study (GraphMAE was added later) trained twice with the
  same seeds and configuration: on an NVIDIA
  A2 with PyTorch 2.11 and on a laptop CPU with PyTorch 2.14. Linear-probe test accuracy,
  absolute differences in points. ``Two independent seeds'' is a Gaussian reference for the mean absolute difference, $2\sigma/\sqrt{\pi}$,
  using the root mean within-objective variance for $\sigma$.}
  \label{tab:reruns}
  \footnotesize
\begin{tabular}{llrrrrrr}
\toprule
Data & Backbone & \begin{tabular}[b]{@{}c@{}}Mean per-seed\\difference\end{tabular} & \begin{tabular}[b]{@{}c@{}}Max per-seed\\difference\end{tabular} & \begin{tabular}[b]{@{}c@{}}Two independent\\seeds (expected)\end{tabular} & \begin{tabular}[b]{@{}c@{}}Same-seed\\correlation\end{tabular} & \begin{tabular}[b]{@{}c@{}}Max difference\\of 10-seed means\end{tabular} & \begin{tabular}[b]{@{}c@{}}Kendall $\tau$\\of means\end{tabular} \\
\midrule
CiteSeer & GCN & 0.89 & 3.8 & 1.58 & 0.76 & 0.94 & 0.90 \\
CiteSeer & GIN & 1.88 & 7.5 & 2.78 & 0.47 & 1.73 & 0.71 \\
Cora & GCN & 0.59 & 2.9 & 1.30 & 0.76 & 0.49 & 0.90 \\
Cora & GIN & 1.63 & 6.5 & 2.08 & 0.44 & 1.23 & 1.00 \\
\bottomrule
\end{tabular}

\end{table}

The GCN and GIN models of Cora and CiteSeer were trained twice with the same code,
configurations and seeds: on an NVIDIA A2 GPU with PyTorch 2.11, and on a laptop CPU with PyTorch
2.14 for the study. Both runs initialize models on CPU with the same numerical seeds, but
this comparison does not verify equality of the initial tensors across library versions.
The augmentations are drawn from the random generator of the device that holds the graph,
so their draws can differ as well. Table~\ref{tab:reruns} pairs runs by seed value. With a
GCN, a seed differs from its rerun by 0.59 to 0.89 points on average, about half of what two
independent seeds would differ by, and the two runs are correlated at 0.76 across seeds. With a
GIN the difference is 1.63 to 1.88 points, three quarters of a new seed, and the correlation
falls to 0.44--0.47; one seed moved by 7.5 points. The means over ten seeds move by at most 0.94
points with a GCN and 1.73 with a GIN, and the rankings agree with $\tau$ between 0.71 and 1.00.

These reruns change hardware and software together. Matching numerical seeds leaves appreciable
variation in individual accuracies, especially with the GIN; it does not isolate a hardware effect. Reporting the means and their
variation across seeds helps readers interpret differences between implementations \citep{summers2021nondeterminism,pham2020variance}. AFGRL does not repeat even on one machine
(Appendix~\ref{app:protocol}).
\FloatBarrier

\FloatBarrier
\section{Objective scores, probe controls and rankings}
\label{app:tables}

\begin{figure}[!htbp]
  \centering
  \includegraphics[width=0.86\linewidth]{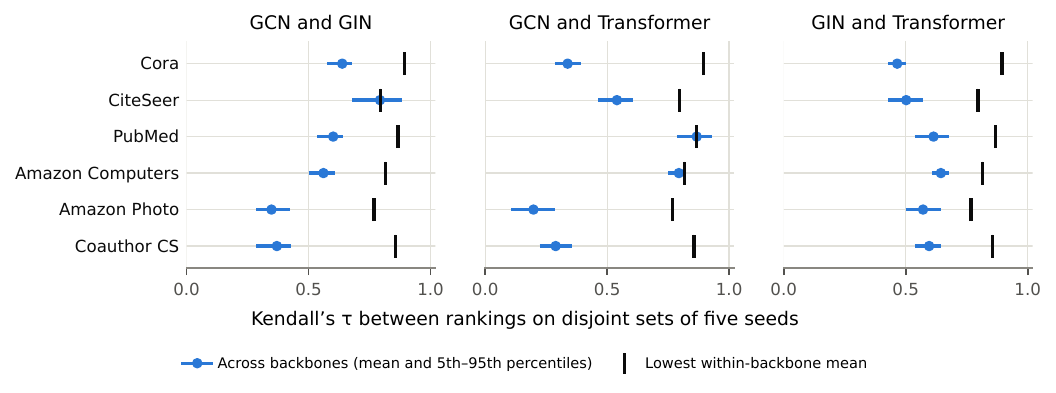}
  \caption{How much ranking disagreement can seed variation produce? For each of the 126
  ways to divide ten seeds into disjoint halves of five, rank the eight objectives on each
  half. Across two backbones we swap the assignments of halves and average. Each blue point
  is the mean agreement over partitions and the line spans their 5th to 95th percentiles.
  Each black tick gives the lowest of the three within-backbone means for that dataset.
  The overlapping partitions describe variability within the observed seeds.}
  \label{fig:seed-halves}
\end{figure}

Table~\ref{tab:main} gives absolute accuracies for every objective and backbone.
Figure~\ref{fig:seed-halves} compares ranking disagreement with seed variability.
Table~\ref{tab:arxiv} gives scores for both larger-graph tasks.
\paragraph{Which objectives rank well across settings?}
GraphCL finishes first in 13 of the 18 main comparisons and in the top three in 17. Its lead over
the runner-up has a seed-bootstrap interval above zero in eight of the 13 first places. All four
of its GCN first places have intervals that include zero. These intervals compare the observed
top two objectives after ranking them on the same test scores and are descriptive. After optimizer and learning-rate tuning (E3), GraphCL places in the top
three in all nine groups and first in three. On ogbn-arxiv it places first on none of the
backbones. For this objective, frequent placement near the top transfers across the six main
datasets more reliably than an outright win.

\begin{table}[!htbp]
  \centering
  \caption{Test accuracy (\%) on the standard split, averaged over 10 seeds with the standardized
  L-BFGS probe. Small numbers give the gain in percentage points over the untrained encoder
  with its initial statistics. The indented row gives the untrained encoder with statistics
  recomputed from the graph. Bold marks the best SSL objective in each column.
  $\dagger$ denotes supervised encoders evaluated with the same probe.}
  \label{tab:main}
  \resizebox{\linewidth}{!}{%
\begin{tabular}{lrrrrrrrrr}
\addlinespace[8pt]\toprule
 & \multicolumn{3}{c}{Cora} & \multicolumn{3}{c}{CiteSeer} & \multicolumn{3}{c}{PubMed} \\
\cmidrule(lr){2-4} \cmidrule(lr){5-7} \cmidrule(lr){8-10}
 & GCN & GIN & Transformer & GCN & GIN & Transformer & GCN & GIN & Transformer \\
\midrule
\textit{Raw features} & \multicolumn{3}{c}{52.6} & \multicolumn{3}{c}{47.5} & \multicolumn{3}{c}{67.0} \\
\textit{Untrained encoder} & 73.1 & 41.7 & 49.7 & 57.7 & 36.1 & 42.7 & 74.1 & 56.6 & 66.5 \\
\textit{\quad statistics from the graph (E7)} & 72.6 & 50.3 & 55.4 & 57.7 & 37.8 & 42.9 & 74.0 & 58.5 & 63.6 \\
\midrule
DGI & 77.7\,{\tiny\color{gray}+4.6} & 66.7\,{\tiny\color{gray}+25.0} & 72.2\,{\tiny\color{gray}+22.5} & 65.2\,{\tiny\color{gray}+7.5} & 47.8\,{\tiny\color{gray}+11.7} & 61.4\,{\tiny\color{gray}+18.7} & 78.8\,{\tiny\color{gray}+4.7} & 66.5\,{\tiny\color{gray}+9.9} & 76.4\,{\tiny\color{gray}+9.9} \\
GraphCL & \textbf{82.2}\,{\tiny\color{gray}+9.0} & \textbf{81.0}\,{\tiny\color{gray}+39.3} & 75.6\,{\tiny\color{gray}+25.9} & \textbf{69.4}\,{\tiny\color{gray}+11.7} & \textbf{65.2}\,{\tiny\color{gray}+29.2} & \textbf{67.7}\,{\tiny\color{gray}+25.0} & \textbf{81.5}\,{\tiny\color{gray}+7.5} & \textbf{80.7}\,{\tiny\color{gray}+24.1} & \textbf{79.7}\,{\tiny\color{gray}+13.3} \\
BGRL & 81.3\,{\tiny\color{gray}+8.2} & 62.8\,{\tiny\color{gray}+21.1} & 69.6\,{\tiny\color{gray}+19.9} & 66.7\,{\tiny\color{gray}+9.0} & 49.4\,{\tiny\color{gray}+13.3} & 55.1\,{\tiny\color{gray}+12.4} & 79.8\,{\tiny\color{gray}+5.7} & 67.9\,{\tiny\color{gray}+11.3} & 75.9\,{\tiny\color{gray}+9.4} \\
AFGRL & 79.3\,{\tiny\color{gray}+6.1} & 70.4\,{\tiny\color{gray}+28.7} & \textbf{76.8}\,{\tiny\color{gray}+27.2} & 62.6\,{\tiny\color{gray}+4.9} & 49.2\,{\tiny\color{gray}+13.1} & 59.5\,{\tiny\color{gray}+16.8} & 78.4\,{\tiny\color{gray}+4.3} & 74.3\,{\tiny\color{gray}+17.7} & 72.4\,{\tiny\color{gray}+5.9} \\
VICReg & 80.8\,{\tiny\color{gray}+7.7} & 77.5\,{\tiny\color{gray}+35.8} & 71.2\,{\tiny\color{gray}+21.6} & 67.7\,{\tiny\color{gray}+10.0} & 63.8\,{\tiny\color{gray}+27.8} & 58.8\,{\tiny\color{gray}+16.1} & 81.5\,{\tiny\color{gray}+7.4} & 79.2\,{\tiny\color{gray}+22.6} & 79.3\,{\tiny\color{gray}+12.8} \\
Barlow Twins & 81.9\,{\tiny\color{gray}+8.8} & 79.2\,{\tiny\color{gray}+37.4} & 72.8\,{\tiny\color{gray}+23.1} & 69.2\,{\tiny\color{gray}+11.5} & 64.8\,{\tiny\color{gray}+28.7} & 66.4\,{\tiny\color{gray}+23.7} & 80.8\,{\tiny\color{gray}+6.7} & 76.4\,{\tiny\color{gray}+19.8} & 78.7\,{\tiny\color{gray}+12.3} \\
GraphDINO & 77.3\,{\tiny\color{gray}+4.1} & 62.8\,{\tiny\color{gray}+21.0} & 56.5\,{\tiny\color{gray}+6.8} & 60.7\,{\tiny\color{gray}+3.0} & 43.0\,{\tiny\color{gray}+6.9} & 41.3\,{\tiny\color{gray}$-$1.4} & 77.1\,{\tiny\color{gray}+3.0} & 70.6\,{\tiny\color{gray}+14.0} & 69.6\,{\tiny\color{gray}+3.2} \\
GraphMAE & 79.8\,{\tiny\color{gray}+6.7} & 74.3\,{\tiny\color{gray}+32.6} & 67.3\,{\tiny\color{gray}+17.7} & 69.2\,{\tiny\color{gray}+11.5} & 65.1\,{\tiny\color{gray}+29.0} & 59.9\,{\tiny\color{gray}+17.1} & 79.9\,{\tiny\color{gray}+5.8} & 74.1\,{\tiny\color{gray}+17.5} & 78.5\,{\tiny\color{gray}+12.0} \\
\midrule
\textit{Supervised, shared protocol} $\dagger$ & 72.7\,{\tiny\color{gray}$-$0.5} & 71.7\,{\tiny\color{gray}+30.0} & 71.3\,{\tiny\color{gray}+21.6} & 56.4\,{\tiny\color{gray}$-$1.3} & 51.1\,{\tiny\color{gray}+15.0} & 60.7\,{\tiny\color{gray}+18.0} & 74.1\,{\tiny\color{gray}+0.0} & 71.6\,{\tiny\color{gray}+15.0} & 70.9\,{\tiny\color{gray}+4.4} \\
\textit{Supervised (std. recipe)} $\dagger$ & 80.9\,{\tiny\color{gray}+7.8} & 77.0\,{\tiny\color{gray}+35.3} & 68.4\,{\tiny\color{gray}+18.7} & 68.0\,{\tiny\color{gray}+10.3} & 61.3\,{\tiny\color{gray}+25.3} & 60.0\,{\tiny\color{gray}+17.3} & 78.7\,{\tiny\color{gray}+4.6} & 75.2\,{\tiny\color{gray}+18.6} & 74.6\,{\tiny\color{gray}+8.1} \\
\bottomrule
\addlinespace[8pt]\toprule
 & \multicolumn{3}{c}{Amazon Computers} & \multicolumn{3}{c}{Amazon Photo} & \multicolumn{3}{c}{Coauthor CS} \\
\cmidrule(lr){2-4} \cmidrule(lr){5-7} \cmidrule(lr){8-10}
 & GCN & GIN & Transformer & GCN & GIN & Transformer & GCN & GIN & Transformer \\
\midrule
\textit{Raw features} & \multicolumn{3}{c}{78.2} & \multicolumn{3}{c}{86.8} & \multicolumn{3}{c}{91.3} \\
\textit{Untrained encoder} & 87.1 & 80.2 & 76.0 & 91.7 & 85.7 & 83.0 & 90.7 & 80.2 & 86.5 \\
\textit{\quad statistics from the graph (E7)} & 87.2 & 78.1 & 79.2 & 91.5 & 84.9 & 85.8 & 90.8 & 79.5 & 86.8 \\
\midrule
DGI & 87.4\,{\tiny\color{gray}+0.3} & 78.0\,{\tiny\color{gray}$-$2.3} & 81.2\,{\tiny\color{gray}+5.1} & 91.5\,{\tiny\color{gray}$-$0.2} & 85.1\,{\tiny\color{gray}$-$0.7} & 88.3\,{\tiny\color{gray}+5.2} & 91.5\,{\tiny\color{gray}+0.8} & 88.4\,{\tiny\color{gray}+8.3} & 90.5\,{\tiny\color{gray}+4.0} \\
GraphCL & \textbf{89.1}\,{\tiny\color{gray}+2.0} & \textbf{86.9}\,{\tiny\color{gray}+6.6} & 87.1\,{\tiny\color{gray}+11.1} & 92.0\,{\tiny\color{gray}+0.3} & \textbf{91.7}\,{\tiny\color{gray}+6.0} & 92.8\,{\tiny\color{gray}+9.7} & 92.4\,{\tiny\color{gray}+1.7} & \textbf{94.6}\,{\tiny\color{gray}+14.4} & \textbf{93.8}\,{\tiny\color{gray}+7.3} \\
BGRL & 88.6\,{\tiny\color{gray}+1.6} & 79.4\,{\tiny\color{gray}$-$0.8} & 84.9\,{\tiny\color{gray}+8.9} & 91.8\,{\tiny\color{gray}+0.1} & 87.1\,{\tiny\color{gray}+1.4} & 91.7\,{\tiny\color{gray}+8.7} & 92.5\,{\tiny\color{gray}+1.7} & 90.1\,{\tiny\color{gray}+10.0} & 93.6\,{\tiny\color{gray}+7.1} \\
AFGRL & 88.3\,{\tiny\color{gray}+1.2} & 77.9\,{\tiny\color{gray}$-$2.3} & 82.2\,{\tiny\color{gray}+6.1} & 92.2\,{\tiny\color{gray}+0.5} & 85.5\,{\tiny\color{gray}$-$0.2} & 90.3\,{\tiny\color{gray}+7.2} & 92.2\,{\tiny\color{gray}+1.4} & 88.2\,{\tiny\color{gray}+8.1} & 91.9\,{\tiny\color{gray}+5.4} \\
VICReg & 89.1\,{\tiny\color{gray}+2.0} & 83.4\,{\tiny\color{gray}+3.1} & \textbf{88.0}\,{\tiny\color{gray}+12.0} & 92.1\,{\tiny\color{gray}+0.4} & 89.6\,{\tiny\color{gray}+3.9} & \textbf{93.6}\,{\tiny\color{gray}+10.5} & 92.1\,{\tiny\color{gray}+1.4} & 94.0\,{\tiny\color{gray}+13.8} & 93.7\,{\tiny\color{gray}+7.2} \\
Barlow Twins & 88.6\,{\tiny\color{gray}+1.5} & 81.7\,{\tiny\color{gray}+1.4} & 86.5\,{\tiny\color{gray}+10.5} & 91.6\,{\tiny\color{gray}$-$0.1} & 87.1\,{\tiny\color{gray}+1.4} & 92.8\,{\tiny\color{gray}+9.8} & 92.4\,{\tiny\color{gray}+1.7} & 91.7\,{\tiny\color{gray}+11.5} & 93.4\,{\tiny\color{gray}+6.9} \\
GraphDINO & 88.0\,{\tiny\color{gray}+1.0} & 81.4\,{\tiny\color{gray}+1.1} & 82.3\,{\tiny\color{gray}+6.3} & 91.9\,{\tiny\color{gray}+0.2} & 87.4\,{\tiny\color{gray}+1.7} & 88.9\,{\tiny\color{gray}+5.9} & 91.6\,{\tiny\color{gray}+0.9} & 88.3\,{\tiny\color{gray}+8.1} & 89.2\,{\tiny\color{gray}+2.6} \\
GraphMAE & 88.6\,{\tiny\color{gray}+1.5} & 84.4\,{\tiny\color{gray}+4.1} & 85.6\,{\tiny\color{gray}+9.5} & \textbf{92.4}\,{\tiny\color{gray}+0.7} & 89.4\,{\tiny\color{gray}+3.7} & 92.4\,{\tiny\color{gray}+9.4} & \textbf{93.1}\,{\tiny\color{gray}+2.3} & 91.8\,{\tiny\color{gray}+11.6} & 91.5\,{\tiny\color{gray}+5.0} \\
\midrule
\textit{Supervised, shared protocol} $\dagger$ & 88.0\,{\tiny\color{gray}+0.9} & 87.3\,{\tiny\color{gray}+7.0} & 86.6\,{\tiny\color{gray}+10.5} & 92.0\,{\tiny\color{gray}+0.3} & 90.4\,{\tiny\color{gray}+4.7} & 92.6\,{\tiny\color{gray}+9.6} & 89.5\,{\tiny\color{gray}$-$1.3} & 90.7\,{\tiny\color{gray}+10.5} & 91.3\,{\tiny\color{gray}+4.8} \\
\textit{Supervised (std. recipe)} $\dagger$ & 89.6\,{\tiny\color{gray}+2.5} & 88.3\,{\tiny\color{gray}+8.1} & 85.1\,{\tiny\color{gray}+9.0} & 92.6\,{\tiny\color{gray}+0.9} & 92.0\,{\tiny\color{gray}+6.3} & 93.4\,{\tiny\color{gray}+10.4} & 92.5\,{\tiny\color{gray}+1.7} & 90.5\,{\tiny\color{gray}+10.3} & 93.3\,{\tiny\color{gray}+6.8} \\
\bottomrule
\end{tabular}
}
\end{table}

\paragraph{Probe controls.} Table~\ref{tab:probe-ablation} gives the scaling, solver and budget controls for the six main graphs.

\begin{table}[!htbp]
  \centering
  \caption{Mean gain of the eight objectives over the untrained encoder under each probe
  (standard split, 10 seeds), and the untrained encoder's own accuracy. The 100-step probe is
  that of DGI and PyG-SSL, the 5{,}000-step probe that of PyGCL, liblinear that of BGRL;
  ``std.'' rows standardize the embeddings first; ``$\lambda=10^{-6}$'' is the regularized probe
  without its penalty selection.}
  \label{tab:probe-ablation}
  \resizebox{\linewidth}{!}{%
\begin{tabular}{lrrrrrrrrr}
\addlinespace[8pt]\toprule
 & \multicolumn{3}{c}{Cora} & \multicolumn{3}{c}{CiteSeer} & \multicolumn{3}{c}{PubMed} \\
\cmidrule(lr){2-4} \cmidrule(lr){5-7} \cmidrule(lr){8-10}
 & GCN & GIN & Transformer & GCN & GIN & Transformer & GCN & GIN & Transformer \\
\midrule
\multicolumn{10}{l}{\textit{Mean gain of the objectives}} \\
100 steps & +5.1 & +30.9 & +18.0 & +1.5 & +18.7 & +12.0 & +6.2 & +28.7 & +8.1 \\
100 steps, std. & +7.5 & +31.0 & +20.9 & +5.9 & +19.6 & +14.1 & +6.7 & +17.4 & +13.1 \\
5,000 steps & +5.8 & +30.6 & +18.0 & +3.7 & +20.4 & +12.1 & +5.0 & +16.8 & +8.3 \\
5,000 steps, std. & +7.5 & +31.8 & +21.0 & +7.4 & +21.4 & +14.9 & +6.6 & +18.1 & +13.4 \\
liblinear & +6.7 & +31.0 & +20.8 & +7.3 & +22.0 & +16.8 & +6.4 & +17.7 & +9.5 \\
Reg., $\lambda=10^{-6}$ & +6.9 & +28.0 & +19.7 & +5.0 & +16.0 & +13.0 & +6.4 & +16.4 & +11.9 \\
Regularized & +6.9 & +30.1 & +20.6 & +8.6 & +20.0 & +16.0 & +5.6 & +17.1 & +9.9 \\
kNN & +13.1 & +39.5 & +32.4 & +11.4 & +24.4 & +19.3 & +4.7 & +39.2 & +13.1 \\
\multicolumn{10}{l}{\textit{Accuracy of the untrained encoder}} \\
100 steps & 72.3 & 40.3 & 47.7 & 56.8 & 35.6 & 39.5 & 72.1 & 45.0 & 66.0 \\
100 steps, std. & 69.4 & 38.8 & 45.5 & 51.8 & 32.3 & 37.5 & 71.7 & 55.7 & 61.6 \\
Regularized & 73.1 & 41.7 & 49.7 & 57.7 & 36.1 & 42.7 & 74.1 & 56.6 & 66.5 \\
\bottomrule
\addlinespace[8pt]\toprule
 & \multicolumn{3}{c}{Amazon Computers} & \multicolumn{3}{c}{Amazon Photo} & \multicolumn{3}{c}{Coauthor CS} \\
\cmidrule(lr){2-4} \cmidrule(lr){5-7} \cmidrule(lr){8-10}
 & GCN & GIN & Transformer & GCN & GIN & Transformer & GCN & GIN & Transformer \\
\midrule
\multicolumn{10}{l}{\textit{Mean gain of the objectives}} \\
100 steps & +7.1 & +12.5 & +9.4 & +2.2 & +6.4 & +8.8 & +9.5 & +33.5 & +8.6 \\
100 steps, std. & +1.2 & +8.0 & +10.0 & +0.2 & +5.9 & +9.5 & +2.3 & +10.6 & +7.9 \\
5,000 steps & +1.3 & +1.4 & +9.6 & $-$0.1 & +1.6 & +8.4 & +1.0 & +10.1 & +5.8 \\
5,000 steps, std. & +1.2 & +1.2 & +10.8 & +0.3 & +2.2 & +9.5 & +1.8 & +11.4 & +7.6 \\
liblinear & +1.2 & +2.9 & +7.8 & +0.3 & +2.1 & +7.5 & +1.3 & +9.1 & +5.6 \\
Reg., $\lambda=10^{-6}$ & +2.2 & +0.3 & +11.0 & +1.2 & +2.6 & +8.7 & +2.2 & +11.8 & +7.1 \\
Regularized & +1.4 & +1.4 & +8.7 & +0.2 & +2.1 & +8.3 & +1.5 & +10.7 & +5.7 \\
kNN & $-$0.5 & +0.1 & +26.8 & $-$0.3 & +1.2 & +34.1 & +2.2 & +22.4 & +25.5 \\
\multicolumn{10}{l}{\textit{Accuracy of the untrained encoder}} \\
100 steps & 80.6 & 65.3 & 74.5 & 89.6 & 80.4 & 82.3 & 82.3 & 56.7 & 83.0 \\
100 steps, std. & 86.4 & 67.5 & 73.9 & 91.4 & 79.3 & 81.6 & 89.2 & 79.4 & 83.7 \\
Regularized & 87.1 & 80.2 & 76.0 & 91.7 & 85.7 & 83.0 & 90.7 & 80.2 & 86.5 \\
\bottomrule
\end{tabular}
}
\end{table}

\FloatBarrier

\begin{table}[!htbp]
  \centering
  \caption{Original evaluation on the A2 GPU. Test accuracy (\%) on ogbn-arxiv, averaged over three seeds after 50 epochs with
  neighbor sampling. Each model is scored on both subject and year labels. Small numbers give
  percentage-point gains over the corresponding untrained encoder with its initial statistics.
  For years the table shows the first of five random 50/25/25 splits.}
  \label{tab:arxiv}
  \resizebox{\linewidth}{!}{%
\begin{tabular}{llrrrrrr}
\toprule
 &  & \multicolumn{3}{c}{Subject labels (ogbn-arxiv)} & \multicolumn{3}{c}{Year labels (arxiv-year)} \\
\cmidrule(lr){3-5} \cmidrule(lr){6-8}
Backbone & Model & 100 steps & Regularized & kNN & 100 steps & Regularized & kNN \\
\midrule
-- & \textit{Raw features} & 46.8 & 52.5 & 47.1 & 34.6 & 34.9 & 31.4 \\
\midrule
GCN & \textit{Untrained encoder} & 46.6 & 69.2 & 66.8 & 37.1 & 39.7 & 39.7 \\
 & DGI & 63.1\,{\tiny\color{gray}+16.6} & 66.8\,{\tiny\color{gray}$-$2.4} & 63.1\,{\tiny\color{gray}$-$3.7} & 41.5\,{\tiny\color{gray}+4.4} & 41.8\,{\tiny\color{gray}+2.0} & 41.5\,{\tiny\color{gray}+1.8} \\
 & GraphCL & 66.9\,{\tiny\color{gray}+20.4} & 69.1\,{\tiny\color{gray}$-$0.0} & 65.2\,{\tiny\color{gray}$-$1.6} & 41.9\,{\tiny\color{gray}+4.8} & 42.3\,{\tiny\color{gray}+2.5} & 40.7\,{\tiny\color{gray}+1.0} \\
 & BGRL & 66.7\,{\tiny\color{gray}+20.2} & 68.5\,{\tiny\color{gray}$-$0.6} & 64.9\,{\tiny\color{gray}$-$1.8} & 41.9\,{\tiny\color{gray}+4.8} & 42.2\,{\tiny\color{gray}+2.4} & 40.6\,{\tiny\color{gray}+0.9} \\
 & AFGRL & 60.5\,{\tiny\color{gray}+14.0} & 63.6\,{\tiny\color{gray}$-$5.6} & 60.0\,{\tiny\color{gray}$-$6.7} & 40.8\,{\tiny\color{gray}+3.7} & 41.9\,{\tiny\color{gray}+2.2} & 38.8\,{\tiny\color{gray}$-$0.9} \\
 & VICReg & 68.0\,{\tiny\color{gray}+21.4} & 69.4\,{\tiny\color{gray}+0.2} & 66.0\,{\tiny\color{gray}$-$0.8} & 42.0\,{\tiny\color{gray}+4.9} & 42.4\,{\tiny\color{gray}+2.6} & 41.8\,{\tiny\color{gray}+2.1} \\
 & Barlow Twins & 67.4\,{\tiny\color{gray}+20.8} & 69.0\,{\tiny\color{gray}$-$0.2} & 65.6\,{\tiny\color{gray}$-$1.2} & 41.9\,{\tiny\color{gray}+4.8} & 42.2\,{\tiny\color{gray}+2.5} & 41.6\,{\tiny\color{gray}+1.9} \\
 & GraphDINO & 67.6\,{\tiny\color{gray}+21.0} & 69.4\,{\tiny\color{gray}+0.2} & 65.4\,{\tiny\color{gray}$-$1.4} & 41.8\,{\tiny\color{gray}+4.7} & 42.2\,{\tiny\color{gray}+2.5} & 41.6\,{\tiny\color{gray}+1.9} \\
 & GraphMAE & 68.6\,{\tiny\color{gray}+22.1} & 69.5\,{\tiny\color{gray}+0.3} & 66.6\,{\tiny\color{gray}$-$0.2} & 41.7\,{\tiny\color{gray}+4.6} & 42.0\,{\tiny\color{gray}+2.3} & 42.3\,{\tiny\color{gray}+2.6} \\
\midrule
GIN & \textit{Untrained encoder} & 36.2 & 55.3 & 53.4 & 31.6 & 40.5 & 37.9 \\
 & DGI & 26.8\,{\tiny\color{gray}$-$9.4} & 47.6\,{\tiny\color{gray}$-$7.7} & 55.0\,{\tiny\color{gray}+1.6} & 27.6\,{\tiny\color{gray}$-$4.0} & 37.5\,{\tiny\color{gray}$-$3.0} & 37.5\,{\tiny\color{gray}$-$0.4} \\
 & GraphCL & 60.7\,{\tiny\color{gray}+24.4} & 63.7\,{\tiny\color{gray}+8.4} & 54.9\,{\tiny\color{gray}+1.5} & 38.7\,{\tiny\color{gray}+7.1} & 40.1\,{\tiny\color{gray}$-$0.4} & 33.7\,{\tiny\color{gray}$-$4.2} \\
 & BGRL & 62.5\,{\tiny\color{gray}+26.3} & 63.7\,{\tiny\color{gray}+8.4} & 60.3\,{\tiny\color{gray}+6.9} & 37.9\,{\tiny\color{gray}+6.4} & 39.7\,{\tiny\color{gray}$-$0.8} & 36.3\,{\tiny\color{gray}$-$1.7} \\
 & AFGRL & 51.5\,{\tiny\color{gray}+15.2} & 57.9\,{\tiny\color{gray}+2.6} & 58.8\,{\tiny\color{gray}+5.4} & 28.9\,{\tiny\color{gray}$-$2.7} & 39.8\,{\tiny\color{gray}$-$0.8} & 37.3\,{\tiny\color{gray}$-$0.7} \\
 & VICReg & 59.9\,{\tiny\color{gray}+23.6} & 63.0\,{\tiny\color{gray}+7.7} & 55.5\,{\tiny\color{gray}+2.2} & 38.5\,{\tiny\color{gray}+6.9} & 40.1\,{\tiny\color{gray}$-$0.4} & 34.5\,{\tiny\color{gray}$-$3.4} \\
 & Barlow Twins & 59.0\,{\tiny\color{gray}+22.8} & 62.5\,{\tiny\color{gray}+7.2} & 55.3\,{\tiny\color{gray}+2.0} & 38.5\,{\tiny\color{gray}+6.9} & 40.3\,{\tiny\color{gray}$-$0.2} & 34.6\,{\tiny\color{gray}$-$3.4} \\
 & GraphDINO & 60.0\,{\tiny\color{gray}+23.8} & 62.8\,{\tiny\color{gray}+7.5} & 59.3\,{\tiny\color{gray}+6.0} & 33.7\,{\tiny\color{gray}+2.2} & 39.9\,{\tiny\color{gray}$-$0.6} & 36.2\,{\tiny\color{gray}$-$1.7} \\
 & GraphMAE & 54.4\,{\tiny\color{gray}+18.2} & 58.6\,{\tiny\color{gray}+3.4} & 61.4\,{\tiny\color{gray}+8.0} & 30.9\,{\tiny\color{gray}$-$0.7} & 39.9\,{\tiny\color{gray}$-$0.7} & 39.6\,{\tiny\color{gray}+1.7} \\
\midrule
Transformer & \textit{Untrained encoder} & 54.5 & 66.4 & 52.5 & 36.2 & 38.5 & 33.2 \\
 & DGI & 34.8\,{\tiny\color{gray}$-$19.7} & 65.1\,{\tiny\color{gray}$-$1.3} & 54.2\,{\tiny\color{gray}+1.8} & 31.2\,{\tiny\color{gray}$-$5.1} & 39.0\,{\tiny\color{gray}+0.5} & 35.0\,{\tiny\color{gray}+1.9} \\
 & GraphCL & 57.5\,{\tiny\color{gray}+3.0} & 62.9\,{\tiny\color{gray}$-$3.6} & 47.0\,{\tiny\color{gray}$-$5.4} & 37.1\,{\tiny\color{gray}+0.8} & 37.8\,{\tiny\color{gray}$-$0.7} & 29.8\,{\tiny\color{gray}$-$3.4} \\
 & BGRL & 59.4\,{\tiny\color{gray}+4.9} & 61.8\,{\tiny\color{gray}$-$4.6} & 47.0\,{\tiny\color{gray}$-$5.4} & 36.7\,{\tiny\color{gray}+0.4} & 37.6\,{\tiny\color{gray}$-$0.9} & 29.8\,{\tiny\color{gray}$-$3.4} \\
 & AFGRL & 67.6\,{\tiny\color{gray}+13.1} & 69.0\,{\tiny\color{gray}+2.6} & 66.5\,{\tiny\color{gray}+14.1} & 37.3\,{\tiny\color{gray}+1.0} & 40.2\,{\tiny\color{gray}+1.8} & 37.8\,{\tiny\color{gray}+4.7} \\
 & VICReg & 57.2\,{\tiny\color{gray}+2.7} & 62.5\,{\tiny\color{gray}$-$3.9} & 46.7\,{\tiny\color{gray}$-$5.8} & 37.1\,{\tiny\color{gray}+0.9} & 37.8\,{\tiny\color{gray}$-$0.7} & 29.8\,{\tiny\color{gray}$-$3.4} \\
 & Barlow Twins & 57.5\,{\tiny\color{gray}+3.0} & 63.1\,{\tiny\color{gray}$-$3.4} & 46.9\,{\tiny\color{gray}$-$5.6} & 37.0\,{\tiny\color{gray}+0.8} & 37.8\,{\tiny\color{gray}$-$0.7} & 29.8\,{\tiny\color{gray}$-$3.3} \\
 & GraphDINO & 61.1\,{\tiny\color{gray}+6.6} & 63.4\,{\tiny\color{gray}$-$3.1} & 49.2\,{\tiny\color{gray}$-$3.2} & 37.5\,{\tiny\color{gray}+1.3} & 38.0\,{\tiny\color{gray}$-$0.5} & 30.5\,{\tiny\color{gray}$-$2.6} \\
 & GraphMAE & 60.2\,{\tiny\color{gray}+5.7} & 64.7\,{\tiny\color{gray}$-$1.7} & 57.7\,{\tiny\color{gray}+5.2} & 35.8\,{\tiny\color{gray}$-$0.5} & 38.7\,{\tiny\color{gray}+0.2} & 35.4\,{\tiny\color{gray}+2.3} \\
\bottomrule
\end{tabular}
}
\end{table}

\end{document}